\documentclass{article}

\usepackage{microtype}
\usepackage{graphicx}
\usepackage{subfigure}
\usepackage{booktabs} % for professional tables

\usepackage{hyperref}

\usepackage[accepted]{icml2024}

\usepackage{amsmath}
\usepackage{amssymb}
\usepackage{mathtools}
\usepackage{amsthm}

\usepackage[capitalize,noabbrev]{cleveref}

\theoremstyle{plain}
\newtheorem{theorem}{Theorem}[section]

\newtheorem{lemma}[theorem]{Lemma}
\newtheorem{corollary}[theorem]{Corollary}
\theoremstyle{definition}
\newtheorem{definition}[theorem]{Definition}
\newtheorem{assumption}{Assumption}

\theoremstyle{remark}
\newtheorem{remark}{Remark}

\usepackage[textsize=tiny]{todonotes}

\def\safedef#1{% 
   \ifx#1\undefined
      \expandafter\def\expandafter#1%
   \else
      \errmessage{The \string#1 is defined already}% 
      \expandafter\def\expandafter\tmp
   \fi
}

\usepackage{amsthm,amsmath,bbm,amsfonts,amssymb}
\usepackage{dsfont} % require for using \mathds{1}
\usepackage{mathtools}
\usepackage{commath}
\usepackage{eqnarray}
\mathtoolsset{showonlyrefs=false}
\allowdisplaybreaks % This allows breaking multi-line equations

\usepackage{xspace}
\usepackage[T1]{fontenc}
\usepackage[utf8]{inputenc}
\usepackage[usenames,dvipsnames]{xcolor} % for color names

\usepackage{hyperref,url}
\usepackage{wrapfig}
\usepackage[export]{adjustbox}
\safedef\imagetop#1{\vtop{\null\hbox{#1}}} % adjusting location of box in table
\usepackage{graphicx,tabularx}
\usepackage{multirow,hhline}% http://ctan.org/pkg/multirow
\usepackage{booktabs}% to use \midrule for drawing a horizontal line in align environment
\usepackage{pbox} % new line in a table cell -- http://tex.stackexchange.com/questions/2441/how-to-add-a-forced-line-break-inside-a-table-cell
\usepackage{algorithm,algorithmic}

\usepackage[export]{adjustbox}

\usepackage{enumitem}
\newcommand{\kj}[1]{{\color{RedOrange}[KJ: #1]}}
\newcommand{\ks}[1]{{\color{OliveGreen}Kyoungseok: #1}}
\newcommand{\chicheng}[1]{{\color{blue}Chicheng: #1}}

\newcommand{\gray}[1]{{ \color[rgb]{.6,.6,.6} #1 }}

\usepackage{framed}
\usepackage[most]{tcolorbox}
\definecolor{kjgray}{rgb}{.7,.7,.7}

\newtheoremstyle{kjstyle}
{1ex} % Space above
{\topsep} % Space below
{\itshape} % Body font
{} % Indent amount
{\bfseries} % Theorem head font
{.} % Punctuation after theorem head
{.5em} % Space after theorem head
{} % Theorem head spec (can be left empty, meaning `normal')

\newtheoremstyle{kjstyle2}
{.0em} % Space above
{.0em} % Space below
{\itshape} % Body font
{} % Indent amount
{\bfseries} % Theorem head font
{.} % Punctuation after theorem head
{.5em} % Space after theorem head
{} % Theorem head spec (can be left empty, meaning `normal')

\newtheoremstyle{kjstylenoitalic}
{1ex} % Space above
{\topsep} % Space below
{} % Body font
{} % Indent amount
{\bfseries} % Theorem head font
{.} % Punctuation after theorem head
{.5em} % Space after theorem head
{} % Theorem head spec (can be left empty, meaning `normal')

\tcbset{kjboxstyle/.style={title={},breakable,colback=white,enhanced jigsaw,boxrule=1.3pt,sharp corners,colframe=kjgray,boxsep=0pt,coltitle={black},attach title to upper={},left=.8ex,bottom=.4em}}    

\tcbset{kjboxstylec/.style={title={},breakable,colback=white,enhanced jigsaw,boxrule=1.3pt,sharp corners,colframe=kjgray,boxsep=0pt,coltitle={black},attach title to upper={},before skip=1.5ex,left=.8ex,bottom=.4em,top=0.4em,enlarge top by=-0.3em,enlarge bottom by=-0.3em}}    

\usepackage{mdframed}
\usepackage{lipsum}
\definecolor{kjgray}{rgb}{.7,.7,.7}
\makeatletter



\makeatletter
\renewcommand{\paragraph}{%
  \@startsection{paragraph}{4}%
  {\z@}{0.50ex \@plus 1ex \@minus .2ex}{-1em}%
  {\normalfont\normalsize\bfseries}%
}
\makeatother

\usepackage[customcolors,shade]{hf-tikz} % after {}, (right, below) and then (left, above) offset... I don't know how it works; done by trial-and-error.

\safedef\horizontalline{\noindent\rule{\textwidth}{1pt} }

\usepackage{tabularx} 
\newcolumntype{P}[1]{>{\centering\arraybackslash}p{#1}}
\newcolumntype{M}[1]{>{\centering\arraybackslash}m{#1}}

\def\ddefloop#1{\ifx\ddefloop#1\else\ddef{#1}\expandafter\ddefloop\fi}

\def\ddef#1{\expandafter\def\csname #1#1\endcsname{\ensuremath{\mathbb{#1}}}}
\ddefloop ABCDFGHIJKLMNORSTUWXYZ\ddefloop % except for E P V Q

\def\ddef#1{\expandafter\def\csname c#1\endcsname{\ensuremath{\mathcal{#1}}}}
\ddefloop ABCDEFGHIJKLMNOPQRSTUVWXYZ\ddefloop

\def\ddef#1{\expandafter\def\csname b#1\endcsname{\ensuremath{{\mathbf{#1}}}}}
\ddefloop ABCDEFGHIJKLMNOPQRSTUVWXYZ\ddefloop  
\def\ddef#1{\expandafter\def\csname b#1\endcsname{\ensuremath{{\boldsymbol{#1}}}}}
\ddefloop abcdeghijklmnopqrtsuvwxyz\ddefloop  % except for bf, which often causes an issue

\def\ddef#1{\expandafter\def\csname h#1\endcsname{\ensuremath{\hat{#1}}}}
\ddefloop ABCDEFGHIJKLMNOPQRSTUVWXYZabcdefghijklmnopqrsuvwxyz\ddefloop % except for 't'
\def\ddef#1{\expandafter\def\csname hc#1\endcsname{\ensuremath{\hat{\mathcal{#1}}}}}
\ddefloop ABCDEFGHIJKLMNOPQRSTUVWXYZ\ddefloop
\def\ddef#1{\expandafter\def\csname hb#1\endcsname{\ensuremath{\hat{\mathbf{#1}}}}}
\ddefloop ABCDEFGHIJKLMNOPQRSTUVWXYZ\ddefloop % 
\def\ddef#1{\expandafter\def\csname hb#1\endcsname{\ensuremath{\hat{\boldsymbol{#1}}}}}
\ddefloop abcdefghijklmnopqrstuvwxyz\ddefloop % 

\def\ddef#1{\expandafter\def\csname t#1\endcsname{\ensuremath{\tilde{#1}}}}
\ddefloop ABCDEFGHIJKLMNOPQRSTUVWXYZabcdefgijklmnpqtsuvwxyz\ddefloop % except for o and h and r
\def\ddef#1{\expandafter\def\csname tc#1\endcsname{\ensuremath{\tilde{\mathcal{#1}}}}}
\ddefloop ABCDEFGHIJKLMNOPQRSTUVWXYZ\ddefloop
\def\ddef#1{\expandafter\def\csname tb#1\endcsname{\ensuremath{\tilde{\mathbf{#1}}}}}
\ddefloop ABCDEFGHIJKLMNOPQRSTUVWXYZ\ddefloop
\def\ddef#1{\expandafter\def\csname tb#1\endcsname{\ensuremath{\tilde{\boldsymbol{#1}}}}}
\ddefloop abcdefghijklmnopqrstuvwxyz\ddefloop % 

\def\ddef#1{\expandafter\def\csname bar#1\endcsname{\ensuremath{\bar{#1}}}}
\ddefloop ABCDEFGHIJKLMNOPQRSTUVWXYZabcdefghijklmnopqrtsuvwxyz\ddefloop
\def\ddef#1{\expandafter\def\csname barc#1\endcsname{\ensuremath{\bar{\mathcal{#1}}}}}
\ddefloop ABCDEFGHIJKLMNOPQRSTUVWXYZ\ddefloop
\def\ddef#1{\expandafter\def\csname barb#1\endcsname{\ensuremath{\bar{\mathbf{#1}}}}}
\ddefloop ABCDEFGHIJKLMNOPQRSTUVWXYZ\ddefloop
\def\ddef#1{\expandafter\def\csname barb#1\endcsname{\ensuremath{\bar{\boldsymbol{#1}}}}}
\ddefloop abcdefghijklmnopqrstuvwxyz\ddefloop % 

\def\ddef#1{\expandafter\def\csname war#1\endcsname{\ensuremath{\overline{#1}}}}
\ddefloop ABCDEFGHIJKLMNOPQRSTUVWXYZabcdefghijklmnopqrtsuvwxyz\ddefloop
\def\ddef#1{\expandafter\def\csname warc#1\endcsname{\ensuremath{\overline{\mathcal{#1}}}}}
\ddefloop ABCDEFGHIJKLMNOPQRSTUVWXYZ\ddefloop
\def\ddef#1{\expandafter\def\csname warb#1\endcsname{\ensuremath{\overline{\mathbf{#1}}}}}
\ddefloop ABCDEFGHIJKLMNOPQRSTUVWXYZ\ddefloop
\def\ddef#1{\expandafter\def\csname warb#1\endcsname{\ensuremath{\overline{\boldsymbol{#1}}}}}
\ddefloop abcdefghijklmnopqrstuvwxyz\ddefloop % 

\safedef\tilr{\tilde r}
\safedef\bff{{\boldsymbol f}}
\safedef\hbff{{\hat{\boldsymbol f}}}
\safedef\hatt{\hat{t}}
\safedef\tilo{{\tilde{o}}}
\safedef\tilh{{\tilde{h}}}
\safedef\bell{{{\boldsymbol\ell}}}
\safedef\tell{\ensuremath{\tilde{\ell}}} 
\safedef\btell{\ensuremath{\widetilde{\boldsymbol{\ell}}}} 
\safedef\hell{{{\hat\ell}}}

\safedef\rialpha{\ensuremath{{\mathring{\alpha}}}} 
\safedef\riz{\ensuremath{\mathring{z}}} 
\safedef\ribeta{\ensuremath{\mathring{\beta}}} 

\newcommand{\fr}[2]{ { \frac{#1}{#2} }}

\newcommand{\wbar}[1]{{\ensuremath{\overline{#1}}}}

\newcommand{\T}{\top}
\safedef\wed{\wedge}
\safedef\tsty{\textstyle}
\safedef\bec{\because}

\safedef\cd{\cdot}
\safedef\cc{{\circ}}
\safedef\la{\langle}
\safedef\ra{\rangle}
\safedef\dsum{\ensuremath{\displaystyle\sum}} 
\safedef\der{\ensuremath{\partial}\xspace}
\safedef\llfl{\left\lfloor} 
\safedef\rrfl{\right\rfloor}  
\safedef\llcl{\left\lceil}  
\safedef\rrcl{\right\rceil}  
\safedef\lfl{\lfloor} 
\safedef\rfl{\rfloor}  
\safedef\lcl{\lceil}  
\safedef\rcl{\rceil}  
\safedef\larrow{\ensuremath{\leftarrow}\xspace} 
\safedef\rarrow{\ensuremath{\rightarrow}\xspace} 
\safedef\sm{{\ensuremath{\setminus}\xspace} }
\safedef\grad{\ensuremath{\mathbf{\nabla}}\xspace}  
\safedef\lt{\left}
\safedef\rt{\right}

\definecolor{mygrn}{rgb}{0,.8,0}
\definecolor{myred}{rgb}{.8,0,0}

\safedef\sig{\sigma}
\safedef\om{\omega}
\safedef\dt{\delta}
\safedef\gam{\gamma}
\safedef\lam{\lambda}
\safedef\kap{\kappa}
\safedef\eps{\varepsilon}
\def\epsilon{\varepsilon}

\safedef\Lam{\Lambda}
\safedef\Dt{\Delta}
\safedef\Gam{\Gamma}
\safedef\Sig{\Sigma}
\safedef\Th{\Theta} 
\safedef\Om{\Omega}

\usepackage{pgffor}
\safedef\greeksymbols{alpha,beta,gamma,gam,delta,dt,eps,epsilon,zeta,eta,theta,th,iota,kappa,kap,lambda,lam,mu,nu,xi,pi,rho,sigma,sig,tau,phi,chi,psi,omega,om,Gamma,Gam,Delta,Dt,Theta,Th,Lambda,Lam,Pi,Sigma,Sig,Phi,Psi,Omega,Om}
\safedef\greeksymbolsnoeta{alpha,beta,gamma,gam,delta,dt,eps,epsilon,zeta,theta,th,iota,kappa,kap,lambda,lam,mu,nu,xi,pi,rho,sigma,sig,tau,phi,chi,psi,omega,om,Gamma,Gam,Delta,Dt,Theta,Th,Lambda,Lam,Pi,Sigma,Sig,Phi,Psi,Omega,Om} % except for eta

\foreach \x in \greeksymbolsnoeta{\expandafter\xdef\csname b\x\endcsname{\noexpand\ensuremath{\noexpand\boldsymbol{\csname \x\endcsname}}}}
\safedef\bfeta{{\boldsymbol \eta}}

\foreach \x in \greeksymbols{\expandafter\xdef\csname h\x\endcsname{\noexpand\ensuremath{\noexpand\hat{\csname \x\endcsname}}}}
\foreach \x in \greeksymbolsnoeta{\expandafter\xdef\csname hb\x\endcsname{\noexpand\ensuremath{\noexpand\hat{\noexpand\boldsymbol{ \csname \x\endcsname}}}}}
\safedef\hbfeta{{\hat{\boldsymbol \eta}}}

\foreach \x in \greeksymbols{\expandafter\xdef\csname bar\x\endcsname{\noexpand\ensuremath{\noexpand\bar{\csname \x\endcsname}}}}
\foreach \x in \greeksymbolsnoeta{%
\expandafter\xdef\csname barb\x\endcsname{\noexpand\ensuremath{\noexpand\bar{\noexpand\boldsymbol{ \csname \x\endcsname}}}}
}
\safedef\barbfeta{{\bar{\boldsymbol \eta}}}

\foreach \x in \greeksymbols{\expandafter\xdef\csname t\x\endcsname{\noexpand\ensuremath{\noexpand\tilde{\csname \x\endcsname}}}}
\foreach \x in \greeksymbolsnoeta{\expandafter\xdef\csname tb\x\endcsname{\noexpand\ensuremath{\noexpand\tilde{\noexpand\boldsymbol{ \csname \x\endcsname}}}}}
\safedef\tbfeta{{\tilde{\boldsymbol \eta}}}

\safedef\dmu{{\dot\mu}}
\safedef\ddmu{{\ddot\mu}}

\DeclareMathOperator{\EE}{\mathbb{E}} % ensures the left space is proper (e.g., 2\EE[X])

\DeclareMathOperator{\PP}{\mathbb{P}}

\DeclareMathOperator{\tr}{\text{\normalfont tr}}

\safedef\clip#1{\wbar{\del{#1}}}
\DeclareMathOperator{\one}{\mathds{1}\hspace{-.1em}}
\safedef\onec#1{\one\cbr{#1}}

\DeclarePairedDelimiterX{\inp}[2]{\langle}{\rangle}{#1, #2}

\newcommand\declareop[3]{%
  \newcommand#1{%
    \mskip\muexpr\medmuskip*#2\relax
    {#3}%
    \mskip\muexpr\medmuskip*#2\relax
}}
\declareop\capprox{1}{{\sr{\const}{\approx}}} % I think 1 means the amount of space.
\declareop\logapprox{1}{{\sr{\mathsf{log}}{\approx}}} % I think 1 means the amount of space.

\safedef\Bin{\mathsf{Bin}}
\safedef\Uniform{{\mathsf{Uniform}}}
\safedef\Bernoulli{{\ensuremath{\mathsf{Bernoulli}}}}

\safedef\kt{{\mathsf{kt}}}
\safedef\mle{{\mathsf{mle}}}

\safedef\Approx{{\mathsf{Approx}} }
\safedef\denom{{\mathsf{denom}}}
\safedef\eff{{\mathsf{eff}}}
\safedef\Seff{{S_{\mathsf{eff}}}}
\safedef\opt{{\mathsf{opt}}}
\safedef\pes{{\mathsf{pes}}}
\safedef\faury{{\mathsf{faury}}}
\safedef\nice{{\textsf{nice} } } 
\safedef\ErrPrb{{\mathsf{ErrPrb}}}
\safedef\Seg{{\mathsf{Seg}}}
\safedef\COM{\mathsf{COM}}
\safedef\const{\mathsf{const}}
\safedef\wo{{\ensuremath{\mathsf{wo}}}}
\safedef\Top{\mathsf{Top}}
\safedef\Bot{\mathsf{Bot}}
\safedef\Sim{\mathsf{Sim}}
\safedef\TV{\mathsf{TV}}
\safedef\tmin{{\min}}
\safedef\tmax{{\max}}
\DeclareMathOperator{\rank}{{\text{rank}}}

\safedef\kl{{\mathsf{kl}}}
\safedef\err{\mathsf{err}} 
\safedef\logloss{{\mathsf{logloss}}}
\safedef\Ber{{\mathsf{Ber}}}
\safedef\erf{{\text{erf}}}
\safedef\erfc{\mathsf{erfc}}
\safedef\rcF{\ensuremath{\mathring{\cF}}} 
\safedef\Cf{{\ensuremath{{\normalfont{\text{Cf}}}}}}
\safedef\barCf{{\ensuremath{\wbar{\text{Cf}}}}}

\safedef\SR{{\ensuremath{\text{SR}}}\xspace}
\safedef\lin{{\ensuremath{\mathsf{lin}}}}

\safedef\IC{{\ensuremath{\normalfont{\text{IC}}}}}

\safedef\Reward{\ensuremath{\text{Reward}}}
\safedef\poly{\operatorname{poly}}
\safedef\Misid{\operatorname{Misid}}
\safedef\Corral{\ensuremath{\normalfont{\textsc{Corral}}}\xspace}
\safedef\AUL{{\ensuremath{\normalfont{\text{AUL}}}}} 
\safedef\Rel{{\ensuremath{\normalfont{\text{Rel}}}}} 
\safedef\Mis{{\ensuremath{\normalfont{\text{Mis}}}}} 
\safedef\Rad{\ensuremath{\text{\normalfont{Rad}}}} 
\DeclareMathOperator*{\argmax}{arg~max}
\DeclareMathOperator*{\argmin}{arg~min}
\DeclareMathOperator{\diag}{{\text{diag}}}

\DeclareMathOperator{\KL}{{\mathsf{KL}}}

\safedef\Reg{{\mathsf{Reg}}}
\safedef\Regret{\ensuremath{\normalfont{\text{Regret}}}}
\safedef\Wealth{\ensuremath{\normalfont{\text{Wealth}}}}
\safedef\Active{\ensuremath{\text{Active}}}
\safedef\decomp{\ensuremath{\mbox{decomp}}\xspace}
\safedef\sym{{\ensuremath{\text{Sym}}\xspace}} 
\safedef\suchthat{\ensuremath{\text{ s.t. }}}

\usepackage{pifont}% http://ctan.org/pkg/pifont
\newcommand{\sr}{\stackrel}

\safedef\bigmid{\,\middle|\,\xspace}

\makeatletter
\newcommand{\vast}{\bBigg@{3}}
\newcommand{\Vast}{\bBigg@{4}}
\makeatother

\safedef\rhoX{{\rho_{\mathcal{X}}}}
\safedef\lammin{{\lambda_{\min}}}
\safedef\elllog{{\ell^{\mathsf{log}}}}
\safedef\lampp{{\lam_\pp}}

\safedef\resh{\text{resh}}
\safedef\SVD{\text{SVD}}
\safedef\op{{\text{op}}}
\safedef\sT{{*\T}}

\newcommand{\edit}[2]{{\xspace\textcolor{blue}{\sout{#1}}}{ \textcolor{red}{#2}}}
\safedef\Vol{{\text{Vol}}}
\safedef\pp{\perp}

\let\vec\undefined % place it before MnSymbol to prevent an error
\DeclareMathOperator{\vec}{\text{\normalfont vec}}

\safedef\rell{{\mathring{\ell}}}
\safedef\lamI{{\lam I}}

\usepackage{enumitem}
\setlist[itemize]{topsep=.5pt,itemsep=0pt,parsep=2pt}
\setlist[enumerate]{topsep=.5pt,itemsep=0pt,parsep=2pt}
\usepackage[normalem]{ulem}

\newcommand\mycommfont[1]{{\footnotesize\ttfamily\textcolor{blue}{#1}}}

\usepackage{color}
\usepackage{makecell}
\usepackage{tablefootnote}

\usepackage{minitoc}

\usepackage[toc,page,header]{appendix}
\usepackage{silence}
\usepackage{bibunits}

\newcommand\wlpopart{\textsf{Warm-LowPopArt}\xspace}
\newcommand\lpopart{\textsf{LowPopArt}\xspace}
\renewcommand\det{\mathrm{det}}
\newcommand\frob{\mathrm{Frob}}
\newcommand\ldyad{B_{\min}(\cA)}
\newcommand\threshold{\lambda_{\text{th}}}
\renewcommand\blambda{B(Q)}
\newcommand{\rmax}{R_{\max}}
\newcommand\inner[2]{{\langle#1,#2\rangle}}
\def\op{{\normalfont\text{op}}}

\def\cC{\mathcal{C}}
\def\cA{\mathcal{A}}
\def\RR{\mathbb{R}}

\def\ve{{\text{vec}}}
\def\rank{{\text{rank}}}

\def\kl{{\text{kl}}}

\newcommand\reshape{\text{reshape}}

\usepackage{amsthm}
\usepackage{amssymb}

\newtheorem{claim}{Claim}

\newif\ifFINAL
\FINALtrue
\ifFINAL

    \def\gray#1{}
    \def\kj#1{}
    
    \renewcommand{\kj}[1]{}
    \renewcommand{\ks}[1]{}
    \renewcommand{\chicheng}[1]{}
    
    \renewcommand{\edit}[2]{#2}
    
\else
    
    \usepackage[inline]{showlabels}

\fi

\icmltitlerunning{Efficient Low-Rank Matrix Estimation, Experimental Design, and Arm-Set-Dependent Low-Rank Bandits}

\begin{document}
%--- adjust the sapce after equation
\setlength{\abovedisplayskip}{4pt}
\setlength{\belowdisplayskip}{4pt}
\setlength{\abovedisplayshortskip}{4pt}
\setlength{\belowdisplayshortskip}{4pt}
%--- adjust the space after algorithm environment
\textfloatsep=.5em
  
\doparttoc % Tell to minitoc to generate a toc for the parts
\faketableofcontents % Run a fake tableofcontents command for the partocs
%\part{} % Start the document part
%\parttoc % Insert the document TOC
% \begin{bibunit}[abbrvnat_lastname_first_local]

\twocolumn[
\icmltitle{Efficient Low-Rank Matrix Estimation, Experimental Design, and Arm-Set-Dependent Low-Rank Bandits}



\icmlsetsymbol{equal}{*}

\begin{icmlauthorlist}
\icmlauthor{Kyoungseok Jang}{unimi}
\icmlauthor{Chicheng Zhang}{uofa}
\icmlauthor{Kwang-Sung Jun}{uofa}
\end{icmlauthorlist}

\icmlaffiliation{unimi}{Dipartimento di Informatica, 
Università degli Studi di Milano, Milan, MI, Italy}
\icmlaffiliation{uofa}{Department of Computer Science, University of Arizona, Tucson, AZ, United States}

\icmlcorrespondingauthor{Chicheng Zhang}{chichengz@cs.arizona.edu}

\icmlkeywords{Machine Learning, ICML}

\vskip 0.3in
]



\printAffiliationsAndNotice{}  % leave blank if no need to mention equal contribution

\begin{abstract}

We study low-rank matrix trace regression and the related problem of low-rank matrix bandits.
Assuming access to the distribution of the covariates, we propose a novel low-rank matrix estimation method called \lpopart and provide its recovery guarantee that depends on a novel quantity denoted by $B(Q)$ that characterizes the hardness of the problem, where $Q$ is the covariance matrix of the measurement distribution.
%low-rank matrix recovery
%geometric characteristic of the measurement distribution $\nu$ denoted by  where $Q$ is the covariance matrix of the measurement distribution. where. %$\ldyad$. % hardness of exploration parameter $B_{\min}(Q)$ 
% In this paper, we propose a novel low-rank matrix recovery method called \lpopart and provide its recovery guarantee that depends on a geometric characteristic of the measurement distribution denoted by $\ldyad$. 
%traditional approaches such as 
We show that our method can provide tighter recovery guarantees than classical nuclear norm penalized least squares \citep{tsybakov2011nuclear} in several problems.
%In order 
To perform efficient estimation with a limited number of measurements from an arbitrarily given measurement set $\cA$, we also propose a novel experimental design criterion that minimizes $B(Q)$ with computational efficiency.
We leverage our novel estimator and design of experiments to derive two low-rank linear bandit algorithms for general arm sets that enjoy improved regret upper bounds. 
This improves over previous works on low-rank bandits, which make somewhat restrictive assumptions that the arm set is the unit ball or that an efficient exploration distribution is given.
%In order 
%-neither of these are practical
To our knowledge, our experimental design criterion is the first one tailored to low-rank matrix estimation beyond the naive reduction to linear regression, which can be of independent interest.

%nontrivial 
%experimental design criterion

% over previous approaches
\end{abstract}
\section{Introduction and related work}\label{sec:intro}
In many real-world applications, data exhibit low-rank structure. For example, in the Netflix problem~\citep{bennett2007netflix}, the user-movie rating matrix can be well-approximated by a low-rank matrix; in demographic surveys~\citep{udell2016generalized}, the respondents' answers to the survey questions are also oftentimes modeled as a low-rank matrix. 
Motivated by these applications, estimation with low-rank structure is {one of the central themes} in high-dimensional statistics~\citep[][Chapter 10]{wainwright_2019}. 
%widespread 

%Low-rank bandit has garnered significant attention from researchers recently. With the increasing complexity of real-world environments, there is a growing interest in high-dimensional reinforcement learning and bandits, where many features need to be taken into account to make optimal decisions. However, as suggested by \citet{wainwright_2019}, without specific structures to exploit, it is practically impossible to learn effective policies. In response, researchers have proposed various structural assumptions such as sparsity or low-rankness to make the learning problem more tractable. This low-rank bandit model can be applicable to various practical scenarios, such as traveling websites, online advertisements, drug discovery, and clothing websites \citep{lu2021low, jun19bilinear}. 

%based on the past arms and rewards
%low-rank

%\chicheng{The current introduction does not formally introduce the low-rank trace regression problem. I think we need it, so I changed the text below a bit.
%}

We study the low-rank trace regression problem~\cite{tsybakov2011nuclear,rohde2011estimation,hamidi20worst} and the related problem of low-rank linear bandits~\cite{jun19bilinear,lu2021low}. In the low-rank linear bandit problem, a learner sequentially learns to choose arms from a given arm set to maximize reward. For each time step $t\in \{1, \cdots, n\}$, the learner chooses an arm $A_t$ from an arm set $\cA\subset \RR^{d_1 \times d_2}$, and receives a noisy reward $y_t = \inner{\Theta^*}{A_t} + \eta_t$, where $\Theta^*$ is a rank-$r$ matrix and $\eta_t$ is $\sigma$-subgaussian noise. The learner's objective is to maximize its cumulative reward, $\sum_{t=1}^n y_t$.
This low-rank bandit model is applicable to various practical scenarios~\citep{natarajan2014inductive,luo17anetwork, jun19bilinear}.

 To name a few examples, in drug discovery~\cite{luo17anetwork}, each $A_t$ represent the outer product $u_t v_t^\T$ of the feature representations of a pair of (drug $u_t$, protein $v_t$), and $\Theta^*$ encodes the interaction between them; in online advertising~\cite{jain2013provable}, each $A_t$ represent the outer product of the feature representation of a pair of (user $u_t$, product $v_t$), and $\Theta^*$ models their interactions.
The bandit problem setup naturally induces an exploration-exploitation tradeoff: as the learner does not know the reward predictor matrix $\Theta^*$, she may need to choose arms that are informative in learning $\Theta^*$; on the other hand, since the learner's objective is maximizing the expected reward, it may also be a good idea to choose arms that the learner believes to yield high reward, based on the past observations.

% should 
%some time steps to figure out the reward
%which is crucial in estimating the reward
%should stop spending too much time steps for exploring $\Theta^*$ 
%such as traveling websites, online advertisements, drug discovery, and clothing websites . 

%with the fact that $\tr{A^\top B} = \inner{\vec(A)}{\vec(B)}$ 
%shown that applying known matrix results
%specialized procedures yield
%too much on the geometry

Early studies on low-rank bandits  \citep{jun19bilinear,lu2021low, jang21improved} have designed bandit algorithms with lower regret than naive approaches that view this problem as a $d_1d_2$-dimensional linear bandit problem~\citep{ay11improved,abe99associative,auer02using, dani08stochastic}. 
However, previous studies lack understandings on the relationship between the geometry of the arm set and regret bounds. 
Usually they assume that a ``nice'' exploration distribution over the arm set is given  \citep{jun19bilinear,lu2021low,kang2022efficient, li2022simple}, or assume that the arm set has some curvature property (e.g., the unit Frobenius norm ball)~\citep{lattimore21bandit, huang2021optimal}. 
Also, some of them rely on subprocedures that are either computationally intractable~\citep[][Algorithm 1]{lu2021low}, or nonconvex optimization steps without computational efficiency guarantees~\citep{lattimore21bandit,jang21improved}; see Appendix \ref{sec:relwork} for more related works.
%\footnote{\ks{Maybe instead of this footnote, add their work in our table?} We exclude \cite{kang2022efficient} from our comparison since their regret bounds involve quantities that have hidden dependence on dimensionality; see Appendix \ref{Appendix:counter examples} for details.} 
To bridge this gap, we ask the following first question: 
\begin{center}
\textit{Can we develop computationally efficient low-rank bandit algorithms that allow generic arm sets and provide guarantees that adapts to the geometry of the arm set?}
\end{center}

It is natural to apply efficient low-rank trace regression results for answering this question, since smaller estimation error leads to fewer samples for exploration thus smaller cumulative regret in bandit problems. In the low-rank trace regression problem, where a learner is given a set of measurements $(X_i, y_i)$ that satisfy that $y_i = \inner{\Theta^*}{X_i} + \eta_i$, where $\Theta^*$ is an unknown matrix with rank at most $r\ll \min(d_1, d_2)$, and $\eta_i$ is a zero-mean $\sigma$-subgaussian noise. The goal is to recover $\Theta^*$ with low error. Throughout, we will use $X_i$ for the supervised learning setting and $A_i$ for the bandit setting.

The low-rank trace regression problem is one of the extensively studied areas within the field of low-rank matrix recovery problems. \citet{keshavan10matrix} provides recovery guarantees for projection based rank-$r$ matrix optimization for matrix completion, and \citet{rohde2011estimation,tsybakov2011nuclear} provide analysis of nuclear norm regularized estimation method for general trace regression, with~\citet{rohde2011estimation} providing further analysis on the (computationally inefficient) Schatten-$p$-norm penalized least squares method. 
Among these approaches, researchers regarded the nuclear norm penalized least square \cite{rohde2011estimation,tsybakov2011nuclear} as the classic approach and applied this method directly \cite{lu2021low} to achieve state-of-the-art algorithm for the low-rank bandit with a general arm set. Since better estimation can lead to better bandit algorithms, we are interested in investigating the following second question:
\begin{center}
    \textit{For low-rank trace regression, can we design estimation algorithms that can outperform the classical nuclear norm penalized least squares?}
\end{center}

\begin{table*}
\small
\begin{tabular}{c|c|c|c|c}
   \toprule
   & Regret bound & \makecell{Regret when\\ $\cA = \cB_{\op}(1)$}& \makecell{Regret when\\ $\cA = \cA_{\text{hard}}$}&Limitation
   \\
   \midrule
\makecell{OFUL \\\!\!\citep{ay11improved}}\!\! & $\tilde{O}(d^2\sqrt{T})$ & $\tilde{O}(d^2\sqrt{T})$& $\tilde{O}(d^2\sqrt{T})$& 
   \\
   \makecell{ESTR \\\citep{jun19bilinear}}& $\tilde{O}(\sqrt{\frac{rd T}{\lambda_{\min}(Q(\pi))}}\del{\frac{\lambda_{1}}{\lambda_{r}}}^3)$& -& -&Bilinear
   % \chicheng{For the rows whose regret bound has $\pi$, we may need to explain $\pi$}
   \\
   
   \makecell{$\epsilon$-FALB \\\citep{jang21improved}}& $\tilde{O}(\sqrt{d^3 T})$& - & - &\makecell{Bilinear \& \\Comp. intractable}\\

   \makecell{rO-UCB \\\citep{jang21improved}}& $\tilde{O}(\sqrt{r d^3 T})$& - & - &\makecell{Bilinear \&\\ Requires oracle}\\
   \makecell{LowLOC \\\citep{lu2021low}}& $\tilde{O}(\sqrt{rd^3 T})$&
   $\tilde{O}(\sqrt{rd^3 T})$ & $\tilde{O}(\sqrt{rd^3 T})$& Comp. intractable
   \\   \makecell{LowESTR\footnotemark{} \\\citep{lu2021low}}& $\tilde{O}( d^{1/4}\sqrt{r\frac{1}{\lambda_{\min}(Q(\pi))^2} T}\del{\frac{S_*}{\lambda_{r}}})$&  $\tilde{O}(\sqrt{rd^{5/2} T})$& $\tilde{O}(\sqrt{rd^{13/2}T})$& \\\makecell{G-ESTT \\
   \citep{kang2022efficient}}& $\tilde{O}( d^{1/4}\sqrt{rdM T}\del{\frac{S_*}{\lambda_{r}}})$&  $\tilde{O}(\sqrt{rd^{5/2} T})$& -\footnotemark{}& \\
   %\citet{lattimore21bandit} &$\tilde{O}( d\sqrt{T})$& \xmark  (symmetric bilinear, $\cA=\SS^{d_1-1}$)& $r=1$\\
   %\citet{huang2021optimal}& $\min(\tilde{O}(d\sqrt{r T}/\lambda_{\min}, (rdT)^{2/3})$&\xmark ($\cA=\{ A: \|A\|_F \leq 1\}$)& \\
   \makecell{Lower bound \\\citep{lu2021low}} & $\Omega(rd\sqrt{T})$ & & \\
   \midrule
   LPA-ETC (Algorithm \ref{alg:etc-lowrank-dr}) & $\tilde{O}((R_{\max} r^2 \ldyad T^2 )^{1/3})$& $\tilde{O}(r^{2/3}d^{2/3} T^{2/3})$& $\tilde{O}(r^{2/3}d T^{2/3})$&\\
   
   LPA-ESTR (Algorithm \ref{alg:estr-lowrank-dr}) & $\tilde{O}( d^{1/4}\sqrt{\ldyad T}\del[1]{\frac{S_*}{\lambda_{r}}}))$& $\tilde{O}(\sqrt{d^{5/2}T})$& $\tilde{O}(\sqrt{d^{7/2}T})$&\\
%   Lower bound (Ours) & $\Omega(r^{1/3} T^{2/3} \cC_{\min}^{-1/3})$ &\\
   \bottomrule
\end{tabular}
\vspace{-.5em}
\caption{
A comparison with existing results on low-rank bandits with fixed arm sets and 1-subgaussian noise. 
Here, $\lambda_{r}$ is abbreviation of $\lambda_{r}(\Theta^*)$, $Q(\pi)$ is the covariance matrix defined in Eq. \eqref{eq: covariance matrix}, $\cB_{\op}(1)$ is the unit operator norm ball, $\cA_{\text{hard}}$ is a special arm set (See Lemma~\ref{lem: comparison between ldyad and cmin}), and 
$\ldyad$ is an arm set dependent constant defined in Eq. \eqref{eq: def of blambda}.
When $\cA \subseteq \cB_{\op}(1)$,  we have $\ldyad = \Omega(d^2)$ and $\lambda_{\min}(Q(\pi)) = O(\frac{1}{d})$, $\forall \pi \in \cP (\cA)$. $M$ is another arm set dependent constant in \cite{kang2022efficient}, see Appendix \ref{appsubsec: kang et al} for more details.
For the third and fourth columns, we set $\pi$ to be the most favorable sampling distribution for prior results as they did not specify the sampling distribution $\pi$ but assumed favorable conditions to hold.
% Since $\ldyad \leq \frac{d}{\lambda_{\min}(Q(\pi))}$ holds for any sampling distribution $\pi \in \cP (\cA)$, \last{\textbf{our LPA-ESTR always outperforms previously known low-rank bandit algorithms, especially LowESTR}.
% Further, compared with the best $\lam_{\min}(Q(\pi))$-free guarantee, which attains $\tO(rd^3 T)$, LPA-ETC attains an improved regret when the dimension is relatively large compared to $T$, i.e., $T< O\del{\frac{d^9}{r\ldyad^2}}$. % compared to other low-rank algorithms. 
% used favorable assumptions and regarded existence of such sampling distribution $\pi$ as their assumption.
% }
%Note that most of the prior works assume Frobenius norm bound of $\Theta^*$, so in the setups of the prior works, $\lam_\tmin \leq 1/\sqrt{r}$ but for us we assumed nuclear norm bound of $\Theta^*$ so it is $\lambda_{\min} \leq 1/r$.
%\chicheng{(1) Can we include the lower bound here? (2) For Kang et al, leave a footnote that ``see Section~\ref{sec:relwork} for detailed explanations'' (3) For Huang et al, they have another result on arm set $\cA = \cbr{ u v^\top: u \in \SS^{d_1-1}, v \in \SS^{d_2-1} }$ - they only present this implicitly in their Remark 3.2. Perhaps also include that result in this table?
%}
$S_*$ and $\rmax$ are upper bounds for $\|\Theta^*\|_*$ and $\max_{A \in \cA} |\inner{A}{\Theta^*}|$ respectively, see Assumption \ref{assumption:nuc-bound} and \ref{assumption:reward-bound}.
\vspace{-1.3em}
}

\label{table:comparison}
%\ks{1) Erase all the bilinear results from the third column. 2) For LowESTR and Ours, try to use $S_*$ for demonstration. 3) Change $Sig(Q)$ to $\lambda_{max} (Q(\pi))$, and change all $C_{min}$ to $\lambda_{\min}(Q(\pi))$. 4) Add footnote for the explanation of modification.5) Never worse than LowESTR. 6) Add column of hard case. 7) Cmin, unify to unscripted version. 8) Our algorithm: give name as LPA-ETC, LPA-ESTR }
\end{table*}

% \ks{To answer these items,
% \begin{itemize}
%     \item 1,3) When I wrote this table, most of the results are based on $\|a\|_F \leq 1$, frobenius norm bounded. I need more detailed analysis, but basically I think we can simply multiply $\min(d_1, d_2)$ for all elements in naive approach since $\|a\|_F \leq \sqrt{\min(d_1, d_2)}\|a\|_{op}$ 
%     \item 2) Yes, Lu has two upper bound, I will fix it.
%     \item 4, 5) I think we need to talk about it.
%     \item 6) As far as I know, the only lower bound that I know is of \citep{lu2021low}, $\Omega(rd\sqrt{T})$.
% \end{itemize}
% }

% with dimensional improvements

%We also find out that the traditional convex optimization is enough for finding sampling distribution to control the error bound of \lpopart.

%This approach is useful in the settings where the learner can control the sampling distribution, such as bandits. 

In this paper, we make meaningful progress in high-dimensional low-rank  trace regression and low-rank
bandits, providing algorithms with arm-set-adaptive exploration and regret analyses for general operator-norm-bounded arm sets.

% \edit{Unlike previous works \citep{jun19bilinear, lu2021low, huang2021optimal, kang2022efficient, lattimore20bandit}, we assume that all our arms are operator norm-bounded, and the hidden parameter $\Theta^*$ is nuclear norm bounded. This bounding norm assumption parallels the linear model case where one could assume the boundedness of $(\ell_\infty, \ell_1)$ vs $(\ell_2,\ell_2)$ norm of (arm set, hidden parameter) pair. There is no consensus in the community on which one is more reasonable, but the former is standard in sparse linear models as it allows an efficient estimation of the sparse unknown parameter. This is true for the matrix bandits as well. }{}

We assume that all arms are operator norm-bounded, and the unknown parameter $\Theta^*$ is nuclear norm bounded as follows:

\begin{assumption}[operator norm-bounded arm set]
\label{assumption:op-bound}
The arm set $\cA$ is such that $\cA \subseteq \cbr{A\in \mathbb{R}^{d_1 \times d_2}: \|A\|_{\op} \leq 1}$. 
\end{assumption}
%expected reward of any arm lies in $[-R_{\max}, R_{\max}]$:
%For any $\Theta$,
\begin{assumption}[Bounded norm on reward predictor]
\label{assumption:nuc-bound}
The reward predictor has a bounded nuclear norm: $\|\Theta^*\|_*\leq S_*$.
\end{assumption}
These two assumptions parallels the standard assumption in the sparse linear model where the covariates are $\ell_\infty$-norm bounded and the unknown parameter is $\ell_1$-norm bounded~\cite{hao2020high}. 

We will also consider the following bounded expected reward assumption in place of Assumption \ref{assumption:nuc-bound}:

\begin{assumption}[Bounded expected reward]\label{assumption:reward-bound} For all $A\in \cA$, $\abs{ \inner{\Theta^*}{A} } \leq \rmax$. 
\end{assumption}

Note that Assumption~\ref{assumption:nuc-bound} implies Assumption~\ref{assumption:reward-bound} with $\rmax = S_*$; however 
the converse is not necessarily true, since $\cA$ is not necessary the unit operator norm ball.

Our contributions are summarized as follows:

\textbf{First,} 
% drawing inspiration from recent works on high-dimensional sparse linear bandits~\citep{hao2020high,jang22popart}, 
under the additional assumption that the measurement distribution $\pi$ is accessible to the learner,
we propose a novel and computationally efficient low-rank estimation method called \lpopart (Low-rank POPulation covariance regression with hARd Thresholding) and prove its estimation error guarantee (Theorem~\ref{thm:wlpopart}) as follows:
%which enjoys an improved operator-norm-based estimation error bound and applicable for general arm sets. \kj{can we say that we, for the first time, provide a precise regret bound that scale with the geometry of the arm set rather than leaving it unspecified like prior works did?} 
% Specifically, the estimation error guarantee (Theorem~\ref{thm:wlpopart}) is of the form 
\[ 
\| \hat{\Theta} - \Theta^* \|_{\op}
\leq 
\tilde{O}\del[2]{ \sigma \sqrt{\frac{B(Q(\pi))}{n_0}} },
\]
where $n_0$ is the number of samples used, and ${B(Q(\pi))}$ (see Eq.~\eqref{eq: optimization of blambda}) is a quantity that depends on the covariance matrix $Q(\pi)$ of the data distribution $\pi$ over the measurement set $\cA$. 
We show that the recovery guarantee of \lpopart is not worse and can sometimes be much better than the classical nuclear norm penalized least squares method~\citep{tsybakov2011nuclear} (see Section \ref{sec:new analysis}).
%the measurement set $\cA$
%arm-set dependent parameter that captures the geometry of

%adaptive compressed sensing and
\textbf{Second,} motivated by the
operator norm recovery bound of \lpopart, we propose a design of experiment objective $B(Q(\pi))$ for finding a sampling distribution that minimizes the error bound of \lpopart. This is useful in settings when we have control on the sampling distribution, such as low-rank linear bandits, the focus of the latter part of this paper.
Applying the recovery bound to the optimal design distribution, we obtain a recovery bound of 
\[ 
\| \hat{\Theta} - \Theta^* \|_{\op}
\leq 
\tilde{O}\del[2]{ \sigma \sqrt{\frac{ \ldyad }{n_0}} },
\]
where $B_{\min}(\cA) := \min_{\pi \in \Delta(\cA)} B(Q(\pi))$ depends on the geometry of the measurement set $\cA$. For example, letting $d := \max\{d_1, d_2\}$, we have $\ldyad = \Theta(d^2)$ and $\Theta(d^3)$ when $\cA$ is the unit operator norm ball and unit Frobenius norm ball, respectively (See Appendix~\ref{sec:b-min-c-min} for the proof).
%arm
Moreover, optimizing our experimental design criterion is computationally tractable. 
In contrast, many prior works on low-rank matrix recovery require finding a sampling distribution that satisfies properties such as restricted isometry property and restricted eigenvalue \citep{hamdi2022low, tsybakov2011nuclear, wainwright_2019} - all these are computationally intractable to compute or verify and thus hard to optimize \citep{Bandeira_2013, juditsky2011verifiable}, which is even harder when the measurements must be limited to an arbitrarily given set $\cA$.

\textbf{Finally, }using \lpopart, we propose two computationally efficient and arm set geometry-adaptive algorithms, for low-rank bandits with general arm sets:
\vspace{-.6em}
% Since $\ldyad \leq \frac{d}{\lambda_{\min}(Q(\pi))}$ holds for any sampling distribution $\pi \in \cP (\cA)$, our LPA-ESTR always outperforms previously known low-rank bandit algorithms, especially LowESTR.
% Further, compared with the best $\lam_{\min}(Q(\pi))$-free guarantee, which attains $\tO(rd^3 T)$, LPA-ETC attains an improved regret when the dimension is relatively large compared to $T$, i.e., $T< O\del{\frac{d^9}{r\ldyad^2}}$.
\begin{itemize}
\item Our first algorithm, LPA-ETC (LowPopArt-Explore-Then-Commit; Algorithm~\ref{alg:etc-lowrank-dr}), leverages the classic explore-then-commit strategy to achieve a regret bound of $\tilde{O}((\rmax r^2 \ldyad T^2)^{1/3})$ (Theorem~\ref{thm:etc regret bound}). 
Compared with the state-of-the-art low-rank bandit algorithms that {allow generic arm sets}~\citep{lu2021low} that guarantees a regret order $\tilde{O}(\sqrt{rd^3T})$, Algorithm~\ref{alg:etc-lowrank-dr}'s guarantee is better when {$T \ll O( \frac{d^9}{\ldyad^2 r})$} (see Remark~\ref{rem:etc-improve} for a more precise statement). 

%for variable selection 
\item Our second algorithm, LPA-ESTR (LowPopArt-Explore-Subspace-Then-Refine; Algorithm~\ref{alg:estr-lowrank-dr}), works under the extra condition that the nonzero minimum eigenvalue of $\Theta^*$, denoted by $\lambda_{\min}$, is not too small.
% This is analogue to the ``minimum signal condition'' exploited in sparse linear bandits~\citep{hao2020high,jang22popart}. 
Algorithm~\ref{alg:estr-lowrank-dr} uses the Explore-Subspace-Then-Refine (ESTR) framework~\citep{jun19bilinear} and achieves a regret bound of $\tilde{O}( \sqrt{d^{1/2}\ldyad T}S_* /\lambda_{\min}))$ (Theorem~\ref{thm:estr bound}). %\chicheng{Define $S_*$ here -- something like ``where $S_*$ is an upper bound of .. ''}
LPA-ESTR gives a strictly better regret bound than previously-known computationally efficient algorithms.
For example, compared to LowESTR \citep{lu2021low}, the regret of our LPA-ESTR algorithm makes not only a factor of $\sqrt{r}$ improvement, but also the dependence on the arm set dependent quantity from $\frac{1}{\lambda_{\min}(Q(\pi))^2}$ to $\ldyad$; {we show that for any $\cA \subset \cB_{\op}(1)$, $B_{\min} (\cA) \leq \frac{1}{\lambda_{\min}(Q(\pi))^2}$ (Lemma \ref{lem: comparison between ldyad and cmin} and Corollary \ref{appcor: Cmin is smaller than 1d}) and there exists an instance $\cA_{\text{hard}}$ such that $d B_{\min} (\cA_{\text{hard}})\leq \frac{1}{\lambda_{\min}(Q(\pi))^2}$} (Lemma \ref{lem: comparison between ldyad and cmin}).

\item Both of our algorithms work for general arm sets, unlike many other low-rank bandit algorithms tailored for specific arm sets such as unit sphere \cite{huang2021optimal}, symmetric unit vector pairs $\{u u^\top: u \in \mathbb{S}^{d-1} \}$ \cite{kotlowski19bandit,lattimore20bandit}, or even one-hot matrices $\{e_i e_j^\top: i,j\in[d] \}$\cite{katariya17bernoulli, trinh2020solving}.

%When $\cA$ is the {unit operator norm ball}, compared with the regret guarantee of state-of-the-art low-rank bandit algorithms that also assumes minimum eigenvalue condition~\citep{jun19bilinear,lu2021low} (which is $\tilde{O}( \sqrt{rd^3 T}/\lambda_{\min})$), our regret bound $\tilde{O}( r^{1/4}d^{5/4}T^{1/2}/\lambda_{\min}))$  is lower. 
\end{itemize}
\vspace{-.6em}
We compare our regret bounds with existing results in Table~\ref{table:comparison}, which showcase how our arm set-dependent regret bounds improve upon prior art in specific arm sets.
We also make a meticulous examination of arm set-dependent constants on regret analysis from previous results, which we believe will help future studies.
\footnotetext[1]{Our bound here is a $d^{1/4}$ factor larger from the original paper since our setting is operator norm bounded action set, which is different from their Frobenius norm bounded action set. For details, see Appendix \ref{appsubsec:d14 modification}. }

\section{Preliminaries}

\paragraph{Basic Notations.} For a matrix $M \in \RR^{d_1 \times d_2}$ and a set of matrices $\cM \subseteq \RR^{d_1 \times d_2}$, let $\vec(M) \in \RR^{d_1 d_2}$ be the vectorization of the matrix $M$ by vertically stacking its columns and  $\vec(\cM) := \{\vec(M): M \in \cM\}$. 
Denote by $\reshape(\cdot)$ the inverse map of $\vec(\cdot)$; i.e., $\reshape(v) = M$ if and only if $\vec(M)=v$.
We assume that $\cA$ spans $\RR^{d_1 \times d_2}$. Define $d = \max(d_1, d_2)$. 
\footnotetext[2]{\cite{kang2022efficient} focused on cases where the arm-set allocation $\pi$ is a continuous distribution with a differentiable probability density. However, $\cA_{\text{hard}}$ is a discrete action set, and theoretical analysis for discrete action sets is not covered in their paper, thus we left this part unaddressed.}
We denote by $v_i$ the $i$-th component of the vector $v$ and by $M_{ij}$ the entry of a matrix $M$ located at the $i$-th row and $j$-th column. 
Let $\lambda_k(M)$ be the $k$-th largest singular value, and define $\lambda_{\max}(M)= \lambda_1(M)$, which is also known as $\| M \|_{\op}$, the operator norm of $M$. Let $\lambda_{\min}(M)$ be the smallest nonzero singular value of $M$.
Let $\| M \|_F = \sqrt{\sum_{i=1}^{d_1} \sum_{j=1}^{d_2} M_{ij}^2}$
and 
$\| M \|_* = \sum_{i=1}^{\min\del{d_1, d_2}} \lambda_i(M)$ be the Frobenius norm of $M$ and nuclear norm, respectively. $\tilde{O}$ is the order notation that hides  logarithmic factors. For any set $S$, let $\cP(S)$ be the set of probability distributions on $S$. For any $\pi \in \mathcal{P}(\cA)$, define the population covariance matrix of the vectorized matrix $Q(\pi) \in \mathbb{R}^{d_1 d_2 \times d_1 d_2}$ as follows: 
\begin{align}\label{eq: covariance matrix}
    Q(\pi) = \EE_{a \sim \pi} \sbr{ \ve(a) \ve(a)^\top }
\end{align}
%\sqrt{\sum_{i=1}^k \lambda_k(M)^2}
%
We define $\cB_{\op} (R) := \{ a \in \mathbb{R}^{d_1 \times d_2}: \|a\|_{\op} \leq R\}$.
\paragraph{Low-rank bandits.} Throughout, we assume that the learning agent interacts with the environment in the following manner. At every time step $t \in \cbr{1, \ldots, T}$, the learner chooses an arm $A_t$ from the arm set $\cA \subset \RR^{d_1 \times d_2}$ and receives reward $y_t = \inner{\Theta^*}{A_t} + \eta_t$, where $\Theta^*$ is an unknown matrix with a known upper bound of the rank at most $r\ll \min(d_1, d_2)$. $\eta_t$ is an independent zero-mean $\sigma$-subgaussian noise, and the inner product of two matrices are defined as $\inner{A}{B}=\inner{\vec(A)}{\vec(B)} = \mathsf{tr}(A^\top B)$. 
The goal of the learner is to minimize its (pseudo-)regret:
\begin{align*}
\Reg(T) := T \max_{A \in \cA} \inner{\Theta^*}{A} - \sum_{t=1}^T \inner{\Theta^*}{A_t}.
\end{align*}
The following matrix generalization of Catoni's robust mean estimator proposed by \citep{minsker2018sub} will be useful for our estimator.
\begin{definition}\label{def:matrix catoni}
    Given a symmetric matrix $M$ with its eigenvalue decomposition $M=U \Lambda U^\top$ where $\Lambda=\diag (\lambda_1, \cdots, \lambda_d)$, we first define $\phi_0: \mathbb{R} \rightarrow \mathbb{R}$ as
    \begin{align*}
        \phi_0 (x) = \begin{cases}
            \log (1+x+ \fr{x^2}{2}) & \text{if } x>0\\
            -\log (1-x+ \fr{x^2}{2}) & \text{otherwise}
        \end{cases}
    \end{align*}
    and $\phi: \mathbb{R}^{d \times d} \rightarrow \mathbb{R}^{d \times d} $ as 
    $$\phi(M) = U \begin{bmatrix}
        {\diag} (\phi_0(\lambda_1), \phi_0(\lambda_2), \cdots, \phi_0 (\lambda_d))
    \end{bmatrix}
    U^\top
    $$
    Finally, for any matrix $A\in \mathbb{R}^{d_1 \times d_2}$, define the dilation operator $\cH: \mathbb{R}^{d_1 \times d_2} \rightarrow \mathbb{R}^{(d_1+d_2) \times (d_1+d_2)}$ as 
    $$ \cH(A) = \begin{bmatrix}
        0_{d_1 \times d_1} & A\\
        A^\top & 0_{d_2 \times d_2}
    \end{bmatrix}.$$
    %\chicheng{Explanations in AISTATS rebuttal can be incorporated here}
\end{definition}
Dilation is a common trick to allow existing estimation tools built for real symmetric matrices to work on rectangular matrices, as in \cite{huang2021optimal, minsker2018sub}. For a dilated matrix $M\in \RR^{(d_1 + d_2)\times (d_1 + d_2)}$, $(M)_{\mathsf{ht}}$ refers to the shorthand of $M_{1:d_1, d_1+1:d_1+d_2}$. 

%for any $$ 
% \paragraph{The warmup length.}
% It is known that bandit algorithms must suffer linear regret at the beginning~\citep{garivier19explore}.
% This means that there are certain warmup periods during which the algorithms do not show much meaningful performance, which is related to \textit{learnability} in statistical learning theory~\cite{shalev-schwartz14understanding}.
% We formally define this quantity as the \textit{warmup length}
% \begin{align*}
%   \tau := \min \cbr[2]{T: \fr{\EE \Reg_T}{T} \le \fr12}~.
% \end{align*}
% For example, if $\Reg_T \le d\sqrt{T}$, then the warmup length $O(d^2)$ since $T\le d^2 \implies d\sqrt{T} \ge T$.
% Note that $\Reg_T \le \sqrt{d T} + d^2$ also has the warmup length $O(d^2)$, even if the leading term is smaller than the above.

% \paragraph{Optimal Experimental design.} In the optimal experimental design problem~\citep{fedorov2013theory}, one is given a measurement set $\cA$, and can make measurements $A_i \in \cA$ and observe noisy linear responses $y_i = \inner{\Theta^*}{A_i} + \epsilon_i$, where $\epsilon_i$ is independent zero-mean random noise. 
% Given a measurement budget $n$, optimal experimental design aims to find a sampling distribution $\pi$ over $\cA$ such that estimation based on $(A_i, y_i)_{i=1}^n$ has the smallest recovery error. Classical experimental design criteria include $A$-, $D$-, $E$- optimality and many others. 
%\input{sections/related_works}
\section{\lpopart: A novel low-rank matrix estimator}
\label{sec:new analysis}
%exploration and

%\chicheng{The writing seems bandit-centric. Can be made more general?}
%\edit{Our main algorithmic tool for designing the bandit algorithm is the following}{}

In this section, we will present our novel low-rank matrix estimation algorithm, LOW-rank Population Covariance regression with hARd Thresholding (\lpopart; Algorithm \ref{alg:lowPopart}), which is inspired by a recent sparse linear estimation algorithm called \textsf{PopArt}~\cite{jang22popart}. We discuss the differences between \lpopart and \textsf{PopArt} in detail at the end of this section.

%with a critical difference that 

\lpopart takes samples $\{X_i , Y_i \}_{i=1}^{n_0}$, sample size $n_0$, the population covariance matrix of the vectorized matrix $Q(\pi)$, pilot estimator $\Theta_0$ and pilot estimation error bound $R_0$ s.t. $\max_{A \in \cA} \abs{ \inner{\Theta_0 - \Theta}{A} } \leq R_0$ as its input.
% \lpopart is different from conventional estimators used in statistical learning because \lpopart requires the population covariance matrix represented as $Q$ (Eq. \eqref{eq: covariance matrix}) as its input, and can exploit the prior knowledge such as pilot estimator $\Theta_0$ if exists. %\lpopart
It consists of three stages. In the first stage, PopArt creates a collection of one-sample estimator $\{\tilde{\Theta}_i\}_{i=1}^{n_0}$ from the input data $\{(X_i, Y_i) \}_{i=1}^{n_0}$ as follows:
\begin{align}\label{eq: one-sample estimator}
    \tilde{\Theta}_i := Q(\pi)^{-1} (Y_i - \inner{\Theta_0}{X_i}) \ve(X_i) 
\end{align}
Note that each $\tilde{\Theta}_i$ is an unbiased estimator of $\Vec(\Theta^* - \Theta_0)$. 

Naively, one could use the average $\bar\Th:= \frac{1}{n_0} \sum_{i=1}^{n_0}\tilde{\Theta}_i$ as an estimator for $\Theta^* - \Theta_0$. 
When the number of samples is large enough, the empirical covariance matrix $\tilde{Q} = \frac{1}{n_0} \sum_{i=1}^{n_0} \ve(X_t) \ve(X_t)^\top$ is close to $Q(\pi)$, which makes $\bar\Th$  close to the $d_1 d_2$-dimensional ordinary least squares (OLS) estimator. 
However, it is not easy to control the tail behavior of $\bar\Th$, and consequently it is hard to exploit the low-rank property when one naively uses $\bar\Th$.
Instead, we use the estimator of \citet[Corollary 3.1]{minsker2018sub} which symmetrizes the original matrix and computes the Catoni function for each eigenvalue (Definition \ref{def:matrix catoni}), which has the effect of lightening the tail distribution of singular values. We call the resulting matrix $\Theta_1$. Finally, we run SVD on $\Theta_1$ and zero out all the singular values smaller than a threshold, to exploit the knowledge that $\Theta^*$ is low-rank. 

\begin{remark}
    In the general estimation problem, we do not have prior knowledge of the inverse covariance matrix of the data, but one may attempt to estimate it if having sample access to the covariate distribution; e.g., matrix geometric sampling \citep{neu2020efficient}. On the other hand, there are some problems (such as bandits or compressed sensing) where the agent has full control over the distribution of the dataset.
    %such as bandit problems
    In these cases, \lpopart can be directly applied. Obtaining a precise performance guarantee when the covariance matrix is estimated from the observed samples is left as future work.
\end{remark}

\begin{algorithm}[t]
\caption{\lpopart}
\label{alg:lowPopart}
\begin{algorithmic}[1]
\STATE {\bfseries Input:} Samples $\{X_i , Y_i \}_{i=1}^{n_0}$, sample size $n_0$, the population covariance matrix of the vectorized matrix $Q(\pi)$, pilot estimator $\Theta_0$ and pilot estimation error bound $R_0$.

\mycommfont{\textbf{Step 1}: Compute one-sample estimators.}

\FOR{$t=1,\ldots,n_0$}
\STATE Compute $\tilde{\Theta}_i$ as in Eq. \eqref{eq: one-sample estimator}. 
\ENDFOR

\mycommfont{\textbf{Step 2}: Compute the matrix Catoni estimator \citep{minsker2018sub} using $\{\tilde{\Theta}_i\}_{i=1}^{n_0}$}
%\chicheng{I think the (inefficient) procedure below for estimating $\Theta$
%is now superceded by our effcient algorithm inspired by~\cite{minsker2018sub}?}

\STATE Compute:
    \[
    \Theta_1    
    =     \Theta_0    +    
    \del[2]{\frac{1}{n_0 \nu} \sum_{i=1}^{n_0} 
    \psi\del[2]{ \nu \cH\del[1]{ \reshape\del[1]{ \tilde{\Theta}_i } } }  }_{\mathsf{ht}}    \]
    where $\nu = 
\frac{1}{\sigma + R_0} \sqrt{\frac{2}{\blambda n_0 }\ln \frac{2d}{\delta}}$ ~.
%\kj{how about replace 12 with $\mathsf{ht}$ where h is for head and t is for tail; or, $0\perp$.}

% \FOR{$(u,v) \in \cN(d, \frac14) \times \cN(d, \frac14)$} 
% \STATE Let $\hat{f}_{u,v} = \catoni( (u \otimes v)^\top \del{ Q(\pi^\star)^{-1} \tilde{A}_t (Y_t - \inner{A_t}{\tau}) + \tau}, t \in (n_1, n_2]) $
% \ENDFOR

% \STATE Let 
% \[ 
% \Theta_1 = \argmin_{\Theta} \max_{u,v \in \cN(d, \frac14) \times \cN(d, \frac14)} 
% \abs{ u^\top \Theta v - \hat{f}_{u,v} }
% \]

%\STATE Let $\tilde{\theta} = \frac{1}{n_2} \sum_{t=n_1+1}^{n_1+n_2} \del{ Q(\pi)^{-1} \tilde{A}_t (Y_t - \inner{A_t}{\tau}) + \tau } \in \RR^{d_1d_2}$ (a vector), and let $\hat{\Theta} \in \RR^{d_1 \times d_2}$ be the rearranged version of $\tilde{\theta}$.
%\Theta_1 = 
\mycommfont{{\textbf{Step 3}: Hard-thresholding eigenvalues.}}

\STATE Let $U_1 \Sigma_1 V_1^\T$ be $\Theta_1$'s SVD. Let $\tilde{\Sigma}_1$ be a modification of $\Sigma$ that zeros out its diagonal entries that are at most $  \threshold:= 2(R_0 + \sigma)\sqrt{\frac{\del{\blambda\ln \frac{2d}{\delta} } }{n_0}}$ where $\blambda$ is in Eq. \eqref{eq: def of blambda}. 
%_{n_1+n_2}

\STATE {\bfseries Return:} Estimator $\hat\Theta = U_1 \tilde{\Sigma}_1 V_1^\top$.
\end{algorithmic}
\end{algorithm}
\paragraph{Analysis of Algorithm~\ref{alg:lowPopart}} 

We start by stating the following recovery guarantee of the estimator $\Theta_1$. Detailed proofs of this part are mainly in Appendix \ref{appsec: proof of new analysis section}.

\begin{figure}[t]      
\centering
\begin{tabular}{cc}
   \includegraphics[width=0.60\linewidth]{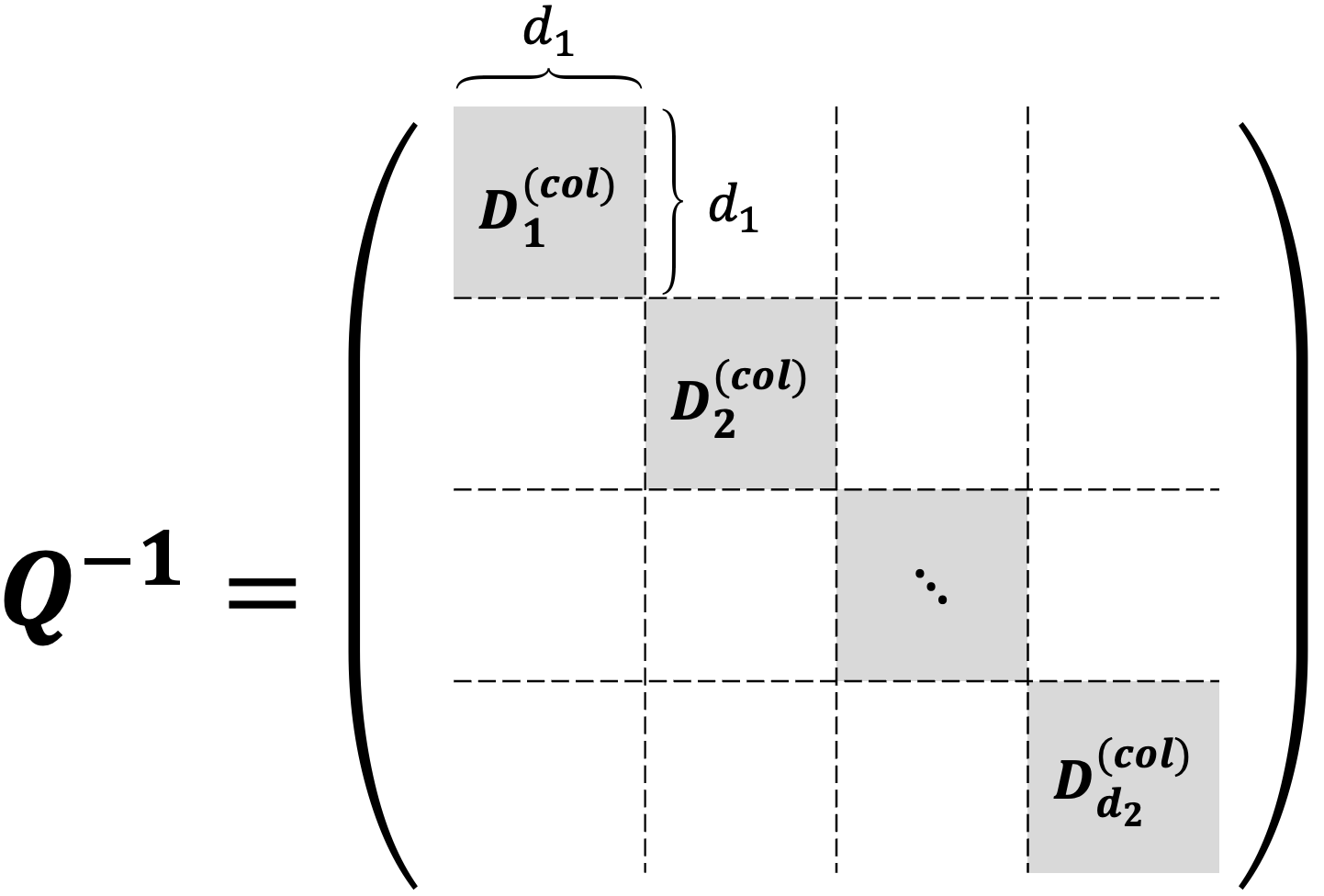} \\  
   \includegraphics[width=0.60\linewidth]{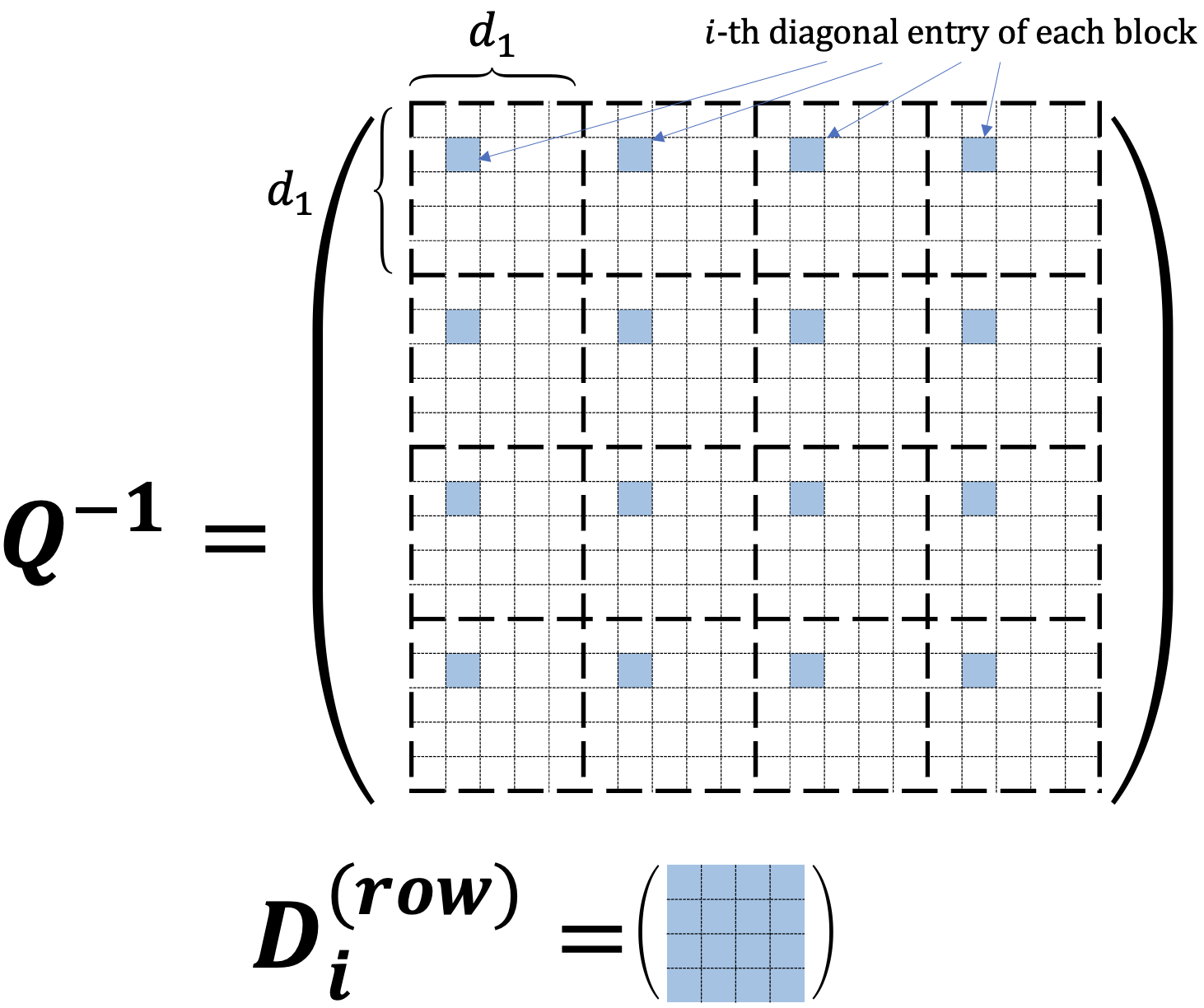}
\end{tabular}
\vspace{-1em}
\caption{Illustration of $D_i^{(\mathrm{col})}$ and $D_i^{(\mathrm{row})}$} 
\label{fig:q-d_1}
\end{figure}

\begin{theorem}\label{thm:lpopart-theta1}
Suppose we run Algorithm \ref{alg:lowPopart} with the arm set $\cA$ \edit{}{which satisfies Assumption \ref{assumption:op-bound}}, sample size $n_0$, population covariance matrix of vectorized matrices $Q$, pilot estimator $\Theta_0$ and pilot estimation error bound $R_0$, such that  $\max_{A \in \cA} \abs{ \inner{\Theta_0 - \Theta^*}{A} } \leq R_0$, then ${\Theta_1}$ satisfies the following {error} bound with probability at least $1-\delta$:  
    \begin{align}
        \| {\Theta}_1 -\Theta^*\|_{\op}
        \leq 
        O\del[2]{ 
        (\sigma+R_0)\sqrt{\tfrac{\blambda}{n_0}\ln \tfrac{2d}{\delta}}
        }~.
    \end{align}
    where 
    \begin{align}\label{eq: def of blambda}
        \blambda := \max \del[3]{
        \lambda_{\max} \del[2]{\sum_{i=1}^{d_2} D_i^{(\mathrm{col})}}, 
        \lambda_{\max}\del[2]{\sum_{i=1}^{d_1} D_i^{(\mathrm{row})}} }
    \end{align}
    where
    $D_i^{(\mathrm{col})}=(Q^{-1})_{[i\cdot d_s+1: (i+1)\cdot d_s],[i\cdot d_s+1: (i+1)\cdot d_s]}$ and
$D_i^{(\mathrm{row})} := [(Q^{-1})_{jk}]_{j,k \in \{i + d_1(\ell-1): \ell \in [d_2]\}};$ see Figure~\ref{fig:q-d_1} for illustrations.
\end{theorem}\label{thm:from Minsker}

%you can take it
\begin{remark}
The intuition underlying $\blambda$ is as follows. When $d=1$,  $\blambda$ is proportional to the variance of $\tilde\Theta_1$; for $d\ge 1$, $\blambda$ is, informally, at most proportional to the largest variance of $\tilde\Theta_1$ projected onto rank-1 dyads $\cbr{u v^\top: u, v \in \mathbb{S}^{d-1}}$; see the proof of Lemma~\ref{lem:bounding-sigma-n} for details.
\end{remark}
% \begin{figure}[h]      
% \centering
% \includegraphics[width=0.8\linewidth]{figures/figure0_def_of_partial_block.png}
% \caption{Definition of $D_i^{(d_1)}$} 
% \label{fig:q-d_1}
% \end{figure}

% \begin{figure}[h]      
% \centering
% \includegraphics[width=0.8\linewidth]{figures/figure0_def_of_partial_block2.png}
% \caption{Definition of $D_i^{(d_2)}$}
% \label{fig:q-d_2}
% \end{figure}

%\edit{$\mathsf{rank}(\hat{\Theta})\leq r$, and the following operator norm bounds hold with probability at least $1-\delta$: }{

From the above Theorem \ref{thm:lpopart-theta1}, one could deduce the final operator norm bound of the output $\hat{\Theta}$.
\begin{theorem}\label{thm:lpopart}
Under the same assumption in Theorem \ref{thm:lpopart-theta1}, 
the following holds with probability at least $1-\delta$: $\mathsf{rank}(\hat{\Theta})\leq r$, and
\begin{align}
        \| \hat{\Theta} -\Theta^*\|_{\op}
        \leq 
        O\del[3]{(\sigma+R_0)
        \sqrt{\frac{\blambda}{n_0}\ln \frac{2d}{\delta}}}
\end{align}
\end{theorem}

Theorem \ref{thm:lpopart} implies the following error bounds in nuclear norm and Frobenius norm recovery errors:

\begin{corollary}\label{cor:nuc anf frob}
    Under the same assumption as in Theorem \ref{thm:lpopart-theta1}, the following nuclear norm and Frobenius norm bounds hold with probabiliy at least $1-\delta$:
    \begin{align}
        \| \hat{\Theta} -\Theta^*\|_{*}
        \leq 
        O\del[3]{(\sigma+R_0)
        \sqrt{\frac{r^2 \blambda}{n_0}\ln \frac{2d}{\delta}}}\\
        \| \hat{\Theta} -\Theta^*\|_{F}
        \leq 
        O\del[3]{(\sigma+R_0)
        \sqrt{\frac{r\blambda}{n_0}\ln \frac{2d}{\delta}}}
\end{align}
\end{corollary}

%Now suppose that
\edit{}{In practical applications, the learner may not have a specific pilot estimator.} A naive application of $\textsf{LowPopArt}$ with pilot estimator $0_{d_1 \times d_2}$ gives an estimator $\hat{\Theta}$ such that  
$\| \hat{\Theta} -\Theta^*\|_{\op}
\leq 
\tilde{O}\del[2]{ (\sigma + \rmax)
\sqrt{\frac{\blambda}{n_0}}}$ \edit{}{with Assumption \ref{assumption:reward-bound}}; the dependence on $\rmax$ is somewhat undesirable when $\rmax \gg \sigma$.
Motivated by this, we propose an improved version of $\textsf{LowPopArt}$ whose estimation error guarantee is $\tilde{O}\del[2]{ \sigma
\sqrt{\frac{\blambda}{n_0}}}$ under mild assumptions, i.e. $\textsf{Warm-LowPopArt}$ (Algorithm~\ref{alg:warm-popart}). 
Its key idea is to first use $\textsf{LowPopArt}$ to 
construct a coarse estimator $\Theta_0$ such that $\| \Theta_0 - \Theta^* \|_{*} \leq \sigma$, which ensures that $\max_{A \in \cA} \abs{ \inner{\Theta_0 - \Theta^*}{A} } \leq \sigma$; it subsequently calls $\textsf{LowPopArt}$ again with $\Theta_0$ as a pilot estimator,  to obtain the final estimate $\hat{\Theta}$. 
Formally, we have the following theorem:

\begin{theorem}\label{thm:wlpopart}
Suppose that \edit{}{Assumption \ref{assumption:op-bound} and \ref{assumption:reward-bound} hold, and} Algorithm \ref{alg:warm-popart} is run with arm set $\cA$, sample size $n_0$, failure rate $\delta$,
and $n_0 \geq \tilde{O}\del{ r^2 \blambda \cdot (\frac{\sigma + \rmax}{\sigma})^2 }$,
then its output $\hat{\Theta}$ is such that $\rank(\hat{\Theta}) \leq r$, and:  
    \begin{align}
        \| \hat{\Theta} -\Theta^*\|_{\op}
        \leq 
        O \del[3]{ \sigma
        \sqrt{\frac{2\blambda}{n_0 }\ln \frac{2d}{\delta}} }.
    \end{align}
\end{theorem}\label{thm:from Minsker}

\paragraph{Comparison with nuclear norm penalty methods} An alternative and popular approach for matrix estimation is nuclear norm penalized least squares~\citep{tsybakov2011nuclear}, which yields a  recovery guarantee of $\| \hat{\Theta} - \Theta^* \|_F \leq \tilde{O}(\sqrt{ \frac{r }{n \lambda_{\min}(Q)^2} })$ and $\| \hat{\Theta} - \Theta^* \|_* \leq \tilde{O}(\sqrt{ \frac{r^2}{n \lambda_{\min} (Q)^2} })$.
We show in Appendix~\ref{appsubsec:lemma proof valko} that under Assumption~\ref{assumption:op-bound}, $\lambda_{\min}(Q) \leq \frac{1}{d}$, and by Lemma \ref{lem: Upper bound Bmin} below, our error bound of $\lpopart$ is always tighter than that of \cite{tsybakov2011nuclear}.

\begin{lemma}\label{lem: Upper bound Bmin}
    $B(Q) \leq \frac{d}{\lambda_{\min}(Q)}$
\end{lemma}

% \chicheng{I wonder if we could discuss what happens when $\pi$ is the uniform distribution over the Frobenius ball here, in which case we have $\lambda_{\min}(Q(\pi)) \leq \frac{1}{d^2}$. 

% On a second thought, this may be an unfair comparison? Since in this case we can get a much better bound of $\|\sum_{i=1}^n A_i \epsilon_i\|_{\op}$ than the naive $O( \sigma \sqrt{ \frac{\ln\frac{d}{\delta}}{n} } )$ outlined in Lemma~\ref{lem:m-concentration}. 
% }

Thus, $B(Q)$ can be viewed as a tighter measurement-distribution-dependent quantity that characterizes the hardness of the low-rank matrix recovery,
%which provides a tighter characterization than the one appeared with $\lam_{\min}^{-1}(Q)$ in prior studies.

However, we can go even further -- it is a natural question to consider how much the recovery error can be reduced when applying the best experimental design tailored to each estimation method.

%An somewhat undesirable property of $\mathsf{LowPopArt}$'s guarantee is its dependency on the range

%\ln \frac{2d}{\delta}}

\begin{algorithm}[t]
\caption{$\textsf{Warm-LowPopArt}$: a bootstrapped version of $\mathsf{LowPopArt}$}
\label{alg:warm-popart}
\begin{algorithmic}[1]
\STATE {\bfseries Input:} Samples $\{X_i, Y_i\}_{i=1}^{n_0}$, sample size $n_0$, population covariance matrix of the vectorized matrix $Q$, failure rate $\delta$.

\STATE $\Theta_0\!\gets\! \mathsf{LowPopArt} (\{X_i, Y_i\}_{i=1}^{\tfrac{n_0}{2}}\!, n_0/2, Q, 0_{d_1\! \times\! d_2}, S_*,\!  \delta/2)\!\!\!\!$

\STATE $\hat{\Theta} \gets\! \mathsf{LowPopArt} (\{X_i, Y_i\}_{i=\tfrac{n_0}{2}+1}^{n_0}, n_0/2, Q, \Theta_0, \sigma, \delta/2)\!\!\!\!\!$

\STATE {\bfseries Return:} $\hat{\Theta}$
\end{algorithmic}
\end{algorithm}

\paragraph{Experimental design} 
As can be seen from Theorem \ref{thm:lpopart}, the recovery guarantee of the \lpopart algorithm depends on the hardness $\blambda$. Therefore, if the agent can design the sampling distribution over the given measurement set $\cA$, a natural choice would be one that minimizes the $\blambda$ value. Formally, we define the optimal $\blambda$ as:
\begin{align}\label{eq: optimization of blambda}
    \ldyad := \min_{\pi \in \mathcal{P}(\cA)} B(Q(\pi))
\end{align}
where $Q(\pi)$ is defined in Eq. \eqref{eq: covariance matrix}. 

Intuitively, this quantity can be understood as a single metric capturing the geometry of the measurement set. This optimization problem is convex and can be efficiently computed using common convex optimization tools such as cvxpy \citep{diamond2016cvxpy}.

Research on the experimental design for low-rank matrix estimation is surprisingly scarce.
%We believe a suitable benchmark 
One reasonable comparison point for our experimental design is the classical E-optimal design~\cite{lattimore21bandit, hao2020high, soare14best}, well-known in experimental design for linear regression. 
E-optimality aims to maximize the minimum eigenvalue of the sampling distribution's covariance matrix, with optimal objective value formally defined as follows:
\begin{align}\label{eq: def of Cmin}
    C_{\min} (\cA) = \max_{\pi \in \mathcal{P}(\cA)} \lambda_{\min} (Q(\pi))
\end{align}
Now, the important question is {how the recovery bounds of $\lpopart$ and nuclear norm penalized least squares differ when written in terms of $C_{\min} (\cA)$ and $\ldyad$, respectively}. We have established the following results between $C_{\min} (\cA)$ and $\ldyad$:
\begin{lemma}\label{lem: comparison between ldyad and cmin}Suppose Assumption~\ref{assumption:op-bound} holds. Then 
    $d^2 \leq \ldyad \leq \frac{d}{C_{\min}}$, and there exists an arm set $\cA_{\textsf{hard}}$ for which $B_{\min}(\cA_{\textsf{hard}}) \approx \frac{1}{C_{\min}}$.
\end{lemma}

%\chicheng{Do we want to ensure that $\cA_{\textsf{hard}}$ satisfies Assumption~\ref{assumption:op-bound}? I think the current construction, ensures only that all elements in $\cA_{\textsf{hard}}$ has operator norm at most 2 (say). }
See Appendix \ref{appsec: ldyad vs cmin, and Ahard} for the proof of Lemma \ref{lem: Upper bound Bmin}, \ref{lem: comparison between ldyad and cmin} and the construction of $\cA_{\textsf{hard}}$. For the arm set $\cA_{\textsf{hard}}$ our guarantee is $\frac{1}{d^{3/2}}$ 
times tighter than the guarantee of \cite{tsybakov2011nuclear}, which shows the importance of using the right arm set geometry quantity.
\paragraph{Main novelty of \lpopart compared to \textsf{PopArt} \cite{jang22popart}.} The major challenge is the absence of the knowledge of a well-structured basis that the agent could exploit a low-rank property of $\Theta^*$ to do better estimation. %low-dimensional subspace 
%proper subspace estimation and the guarantees on the signal recovery
In sparse linear bandits, the basis for testing the zeroness is known to the agent (i.e. the canonical basis), so the estimation procedure can simply focus on controlling the estimation error over the $d$ coordinates. On the other hand, in low-rank bandits, we need to control the subspace estimation error, but the potential number of subspace directions (i.e., $\mathcal{F} = \{u v^\top: u \in \mathbb{S}^{d_1-1}, v \in \mathbb{S}^{d_2-1}\}$ or its $\epsilon$-net) is infinite or exponentially large ($\sim \exp (d_1 + d_2)$).
Indeed, one of the naive extension of \cite{jang22popart} for estimation, which considers all possible directions in an $\epsilon$-net of $\mathcal{F}$.  However, this causes computational intractability. % since the optimizer needs to consider exponentially many directions ($\sim \exp (d_1 + d_2)$). 
To get around this issue, we propose to directly upper bound $\|\hat{\Theta} - \Theta^* \|_{\mathrm{op}}$ for establishing Frobenius and nuclear norm recovery error guarantees, which can be performed via the method of \cite{minsker2018sub} in a computationally efficient manner.
This was the key observation that led to our main result.

\section{Low rank bandit algorithms}

%Algorithm \ref{alg:lowPopart}
We now leverage \lpopart to design two computationally efficient algorithms for low-rank bandits. 

\paragraph{Explore-then-commit based algorithm.}
Algorithm \ref{alg:etc-lowrank-dr} is based on the well-known Explore-then-Commit framework. It uses $\textsf{Warm-LowPopArt}$ as its exploration method to obtain $\hat{\Theta}$, an estimate of $\Theta^*$, and subsequently takes the greedy arm with respect to $\hat\Theta$.

%performs exploitation

\begin{algorithm}[t]
\caption{LPA-ETC (\lpopart based Explore then commit)}
\label{alg:etc-lowrank-dr}

\begin{algorithmic}[1]

\STATE {\bfseries Input:} time horizon $T$, arm set $\cA$, exploration lengths $n_0$, regularization parameter $\nu$, pilot estimator $\Theta_0$

\STATE Solve the optimization problem in Eq. \eqref{eq: optimization of blambda} and denote the solution as $\pi^*$

\FOR{$t = 1, \ldots,n_0$} 
\STATE Independently pull the arm $A_t$ according to $\pi^*$ and receives the reward $Y_t$
\ENDFOR
\STATE Run $\textsf{Warm-LowPopArt} (\{A_i, Y_i\}_{i=1}^{n_0}, n_0, Q(\pi^*), \delta)$ and get $\hat\Theta$

\FOR{$t=n_0+1,\ldots,T$}
\STATE Pull the arm $A_t = \argmax_{A \in \cA} \inner{\hat{\Theta}}{A}$
\ENDFOR
\end{algorithmic}
\end{algorithm}

We prove the following regret guarantee:

%, $\nu = \sqrt{\frac{2\cC_{\min}}{d n_0 }\ln \frac{2d}{\delta}}$

%S_*

\begin{theorem}[Regret upper bound]\label{thm:etc regret bound} 
Suppose \edit{}{that Assumption \ref{assumption:op-bound} and \ref{assumption:reward-bound} hold, and} $T \geq r{\ldyad} (\frac{\sigma + \rmax}{\sigma})^4$.
The regret upper bound of Alg. \ref{alg:etc-lowrank-dr} with $n_0 = \min(T, \del[2]{\sigma^2 r^2 \ldyad T^2/\rmax^2}^{1/3})$ is as follows:
\begin{align}
    \normalfont{\Reg}(T) \leq \tilde{O}((\sigma^2 \rmax r^2 T^2 \ldyad)^{1/3})
\label{eqn:reg-ub-etc}
\end{align}
\end{theorem}

\begin{remark}\label{rem:etc-improve}
To the best of our knowledge, the only algorithms that can handle general arm sets with $\lambda_{\min}(\Theta^*)$-free regret bounds 
% and provide regret bounds independent of $\lambda_{\min}(\Theta^*)$
are LowLOC~\citep{lu2021low} and rO-UCB~\citep{jang21improved}.
Both algorithms have regret bounds of $O ( \sigma r^{1/2} d^{3/2} \sqrt{T} )$ but are not computationally tractable.
% \kj{\textbf{(B)} should we include the norm dependence of these algorithms?}
On the other hand, our ETC-based algorithm is computationally efficient and achieves a better regret bound when $T \leq O\del{ {\sigma^2 d^9}{\rmax^{-2} \ldyad^{-2} r^{-1}} }$.
%Furthermore, the warm up time of LowLOC and rO-UCB is $O(d^3 r)$ whereas our result in Theorem~\ref{thm:etc regret bound} is $O(d^2 r^2)$ when $B_{\min}(\cA) = \Theta(d^2)$ (e.g., the unit nuclear norm ball arm set).
\end{remark}

%\chicheng{Just a note: if we want to choose $n_0$ without the $\min(T, )$ operation, we can impose  the additional constraint that $T \geq \frac{r^2 d}{C_{\min}} \frac{\sigma^2}{S_*^2}$.}

%\subsection{A $\Omega(T^{2/3})$ lower bound }

%\chicheng{Todo: add $\Omega(T^{2/3})$ lower bound here.}

%One might wonder if the Theorem~\ref{thm:etc regret bound}'s regret bound~\eqref{eqn:reg-ub-etc} dependence on $\cC_{\min}$ is fundamental. 

%Commented at 231009
% \textcolor{red}{
% \begin{remark}
% As we shall see in Section~\ref{sec:lower-bound}, our lower bound Theorem~\ref{thm:lower-bound-lr} implies that the dependence on $\cC_{\min}$ in Eq.~\eqref{eqn:reg-ub-etc} is fundamental: it is impossible to design an algorithm with a regret upper bound of only $\tilde{O}((\sigma^2 S_* r^2 d T^2)^{1/3})$, without dependence on  $\cC_{\min}^{-1/3}$. In other words, any attempt in removing the dependence on $\cC_{\min}$ will come at a price of having a higher dependence on other parameters (e.g. $d$ or $T$). 
% \end{remark}
% }
\paragraph{Explore-Subspace-Then-Refine (ESTR) based algorithm.}

\newcommand\minsig{S_r}
%\chicheng{We can say, ``throughout this section, we make an extra assumption that the learner has the knowledge that $\| \Theta^* \|_{\op} \leq S_2$''}
Although general, Algorithm~\ref{alg:etc-lowrank-dr} 
overlooks 
a favorable structure underlying many low-rank bandit problems: $\Theta^*$ is well-conditioned in many settings, e.g. $\lambda_{\min} \geq \Omega( S_* / r )$. Such structure has been exploited by many prior works~\cite{jun19bilinear,lu2021low,kang2022efficient} to design $\sqrt{T}$-regret algorithms. {In this part, \edit{}{in addition to Assumption \ref{assumption:op-bound} and \ref{assumption:nuc-bound},} we assume that $\lambda_{\min}(\Theta^*) \geq \minsig$ for some known $\minsig>0$. }
\begin{algorithm}[t]
\caption{LPA-ESTR (\lpopart based Explore Subspace Then Refine)}
\label{alg:estr-lowrank-dr}
\begin{algorithmic}[1]

\STATE {\bfseries Input:} time horizon $T$, arm set $\cA$, exploration lengths $n_0$, singular value lower bound $\minsig$
%, regularization parameter $\nu$, pilot estimator $\Theta_0$

\STATE Solve the optimization problem in Eq. \eqref{eq: optimization of blambda} and denote the solution as $\pi$

\FOR{$t = 1, \ldots,n_0$} 
\STATE Independently pull the arm $A_t$ according to $\pi$ and receives the reward $Y_t$
\ENDFOR

\STATE Run $\textsf{Warm-LowPopArt}(\{A_i, Y_i\}_{i=1}^{n_0}, n_0, Q(\pi), \delta)$ and get $\hat\Theta$ with SVD result $\hat{\Theta}=\hat{U}\hat{\Sigma} \hat{V}^\top$. 

\STATE Let $\hat{U}_\perp$ and $\hat{V}_\perp$ be the orthonormal bases of {the orthogonal complement} subspaces of $\hat{U}$ and $\hat{V}$, respectively.

\STATE Rotate whole arm feature set $\cA':=\{[\hat{U} \; \hat{U}_\perp] A [\hat{V} \;\hat{V}_{\perp}]^\top : A \in \cA \}$

\STATE Define a vectorized arm feature set so that the last $(d_1 - r)(d_2 - r)$components are from the complementary subspaces:
%\vspace{-1em}
\begin{align*}
    \cA_{vec}' &:= \{  ( \vec(A'_{1:r,1:r}); 
\vec(A'_{r+1:d_1,1:r}); \\
 & \vec(A'_{1:r,r+1:d_2}); 
\vec(A'_{r+1:d_1,r+1:d_2}) ) : A'\in \cA'\}
\end{align*} 
%\vspace{-2em}
\STATE Invoke LowOFUL with time horizon $T - n_0$, 
arm set $\cA_{\vec}'$, the low dimension $k=r(d_1 + d_2 -r)$, $\lambda = \frac{\sigma^2}{S_*^2} d r$, 
    $\lambda_\perp = \frac{T}{r \log (1+ \frac{dT}{\lambda})}$, $B= S_*$, and $B_\perp = \frac{\ldyad \sigma^2 S_* }{n_0 \minsig^2}$.
%\ENDFOR
\end{algorithmic}
\end{algorithm}

In this section, we use $\wlpopart$ to design an efficient algorithm with $O(\sqrt{T})$ regret (Algorithm~\ref{alg:estr-lowrank-dr}).
%Throughout this section, we make the following extra assumption:
%\begin{assumption}
%$\| \Theta^* \|_{\op} \leq %S_2$. 
%\end{assumption}
Algorithm \ref{alg:estr-lowrank-dr} is based on the Explore-Subspace-Then-Refine (ESTR) framework~\cite{jun19bilinear}. 
In ESTR, we use \wlpopart to find an estimate $\hat{\Theta}$ such that it closely approximates $\Theta$ in operator norm.
We then estimate the row and column spaces of $\Theta$ using an SVD over $\hat{\Theta}$, represented by their orthonormal bases $\hat{U}$ and $\hat{V}$. 
Then, we rotate the arm set using $\hat{U}$ and $\hat{V}$.
After this transformation, the original linear bandit problem becomes a $d_1 d_2$-dimensional linear bandit problem with arm set $\cA'$ and reward predictor

\begin{align*}
    \theta^*
= 
( &\vec(\hat{U}^\top \Theta^* \hat{V}); 
\vec(\hat{U}_\perp^\top \Theta^* \hat{V});\\
&\vec(\hat{U}^\top \Theta^* \hat{V_\perp});
\vec(\hat{U}_\perp^\top \Theta^* \hat{V_\perp})
)
\end{align*}

Crucially, by the recovery guarantee of \wlpopart and Wedin's Theorem \citep{stewart1990matrix}, $\| \hat{U}_\perp^\top U \|_{\op}$
and $\| \hat{V}_\perp^\top V \|_{\op}$ are both small; as a consequence, 
$\| \theta^*_{r(d_1+d_2-r)+1:d_1d_2} \|_2 = \| \vec(\hat{U}_\perp^\top \Theta^* \hat{V_\perp}) \|_F \leq \| \hat{U}_\perp^\top U \|_{\op} \| \Theta^* \|_F \| \hat{V}_\perp^\top V \|_{\op}$, which is also small. 
In other words, we are now faced with a linear bandit problem with the prior knowledge that a large subset of the coordinates of the reward predictor is small.

% \[
% r_t = \inner{ 
% \begin{bmatrix} 
% \hat{U}^\top \Theta^* \hat{V} & \hat{U}^\top \Theta^* \hat{V_\perp} \\
% \hat{U}_\perp^\top \Theta^* \hat{V} & \hat{U}_\perp^\top \Theta^* \hat{V_\perp}
% \end{bmatrix} }{}
% \]

%[\hat{U} \; \hat{U}_\perp]^\top \Theta^* [\hat{V} \; \hat{V}_\perp]

%After that, the Algorithm \ref{alg:estr-lowrank-dr} run
This motivates the usage of the 
LowOFUL algorithm~\cite{jun19bilinear}\footnote{Pseudocode of LowOFUL is in Appendix \ref{appsec:lowoful}, Algorithm~\ref{alg:lowoful}.} in the second stage, which is a modification of OFUL~\citep{ay11improved} with heavy penalizations on the reward predictor on insignificant coordinates. Theorem \ref{thm:estr bound} states the overall regret upper bound of Algorithm \ref{alg:estr-lowrank-dr}.

%helps us to

%\ks{Add assumption on the operator norm bound on $\Theta^*$ around here. }
%\chicheng{Now that we assume the arm set has operator norm bound 1, I think we should just set $X_\infty = 1$.}

%[\ks{Am I right?}], 
    % \chicheng{I think we need to set $\lambda$ to ensure that 
    % $\lambda \sqrt{B} \leq d r$ so that the first two terms of $\beta_t \leq dr$?
    % }

% d X_\infty^2

\begin{theorem}\label{thm:estr bound} Suppose \edit{}{that Assumptions \ref{assumption:op-bound} and \ref{assumption:nuc-bound} hold, $\lambda_{\min}(\Theta^*) \geq \minsig$ for some known $\minsig>0$, and} $T \geq \frac{16\ldyad \sigma^4}{d^{0.5}\minsig(\Theta^*)^2}$. The regret upper bound of Algorithm \ref{alg:estr-lowrank-dr} with $n_0 = \sqrt{\frac{d^{0.5}\ldyad}{\minsig^2}T}$ is 
$$ 
  \Reg(T) \leq \tilde{O}\del[3]{\sigma \sqrt{ \tfrac{ S_*^2}{\minsig^2} \ldyad d^{0.5}T }}
$$
with probability at least $1-2\delta$.
\end{theorem}
Algorithm~\ref{alg:estr-lowrank-dr} attains a $\sqrt{T}$-order regret bound, at the cost of introducing a dependence of  $S_r$ factor in the regret bound.
\begin{remark}
%For the case where $S_* \leq O(\|\Theta^* \|_{\op}) = O( \lambda_1)$ \kj{example: $\lam_{i} = 1/(i\ln^2(1+i))$}, the regret bound can be simplified to $O( \sigma \sqrt{ \frac{ \lambda_1^2}{\lambda_r^2} {d^{0.5} \ldyad} T } )$.
When $\Theta^*$ is well conditioned, i.e. 
$S_r \geq \Omega( S_* / r )$, the above regret bound can be simplified to $O( \sigma \sqrt{ {r^2 d^{0.5} \ldyad} T } )$. For the case {where} $\cA = \cB_{\op}(1)$, we can prove $\ldyad \leq d^2$, and we have the upper bound of order $\tilde{O}(\sqrt{r^{2}d^{2.5}T})$ when $\Theta^*$ is well-conditioned, which is an improved result compared to $\sqrt{r^3 d^{2.5} T}$ of \citet{lu2021low}
and even to the computationally inefficient result $\sqrt{r d^3 T}$ of \citet{lu2021low}. {Plus, our algorithm is strictly better than LowESTR \citep{lu2021low} in any cases because $\ldyad \leq \frac{1}{\lambda_{\min}(Q(\pi))^2}, \forall \pi \in \cP(\cA)$ by Lemma \ref{lem: comparison between ldyad and cmin}}.

%Note that we made nuclear norm bound assumption on $\Theta^*$ so $\lambda_r < \frac{S_*}{r}$, while most of the prior works assume Frobenius norm bound of $\Theta^*$ so that $\lambda_r \leq \frac{\|\Theta^*\|_F}{\sqrt{r}}$. \kj{$\larrow$ move this sentence to earlier part of the paper, somewhere relevant.}
%\kj{instead of saying `new', can we elaborate whether it is better bound or the tightest bound under the same setup? what we have now : $\sqrt{d^{2.5}r^2 T}$; compare this with $\sqrt{d^3 r^2 T}$ of Jun et al. and $\sqrt{d^3 r T}$ (inefficient) from Lu et al.} in comparable with prior works \cite{jun19bilinear, lu2021low}.

%\chicheng{I would say something weaker like ``this is a new result in comparable with prior works, in that ..''}
%is a significantly better result than \citet{jun19bilinear, lu2021low} exploiting the regret bound. 
    %Informally, we have the regret bound of order
    %$$ \Reg(T) \leq

    % \chicheng{I suggest that removing the remark below as this paper does not directly tackle the Frobenius norm bounded arm set setting.}
    % We also stated the result when the arm set is Frobenius norm bounded, and we successfully recovered the regret bound of $\tilde{O}(\sqrt{rd^3T})$ for one of the general cases when $\cC_{\min}=d_1 d_2$ (which is also assumed on \citet{jun19bilinear, lu2021low}). For the readers who are interested, please check Appendix. 
\end{remark}

{\begin{remark}
In addition to arm set dependent constant, LPA-ESTR also achieves an improved regret guarantee over LowESTR~\citep{lu2021low} w.r.t. $r$. 
This is because our $\lpopart$ estimator provides improved bounds on $\| \hat{U}_\perp^\top U \|_{\op}$
and $\| \hat{V}_\perp^\top V \|_{\op}$, which are a factor of $\sqrt{r}$ lower than their respective bounds in~\citep{lu2021low}. 
This is enabled by the unique operator-norm based recovery guarantee of $\lpopart$ and the operator norm-version of Wedin's Theorem; to the best of our knowledge, we are not aware of an operator-norm-based recovery guarantee for nuclear norm penalized least squares regression.
\end{remark}
}

%\chicheng{Todo for CZ: add a lower bound section here.}

\section{Experiments}
\label{sec:expr}
\vspace{-6pt}

We now evaluate the empirical performance of \lpopart and our proposed experimental design to validate our improvement. 
\edit{}{For all experiments, we set ground truth $\Theta^*=uv^\top$ where $,u \sim \texttt{Unif}(\mathbb{S}^{d_1-1})$ and $v\sim\texttt{Unif}(\mathbb{S}^{d_2-1})$ and we sample $\Theta^*$ before each experiment starts. The noise of the reward $\eta_t \sim N(0,1)$. All plots are generated by averaging over 60 number of random instances.}
We defer unimportant details of the experimental setup in Appendix \ref{appsec: experiment settings}, and please check \url{https://github.com/jajajang/LowPopArt} for the code. %See Appendix \ref{appsec: experiment settings} for settings. 

\begin{figure}[h]        
    \centering
    \begin{tabular}{ll}
                 \includegraphics[width=0.45\linewidth]{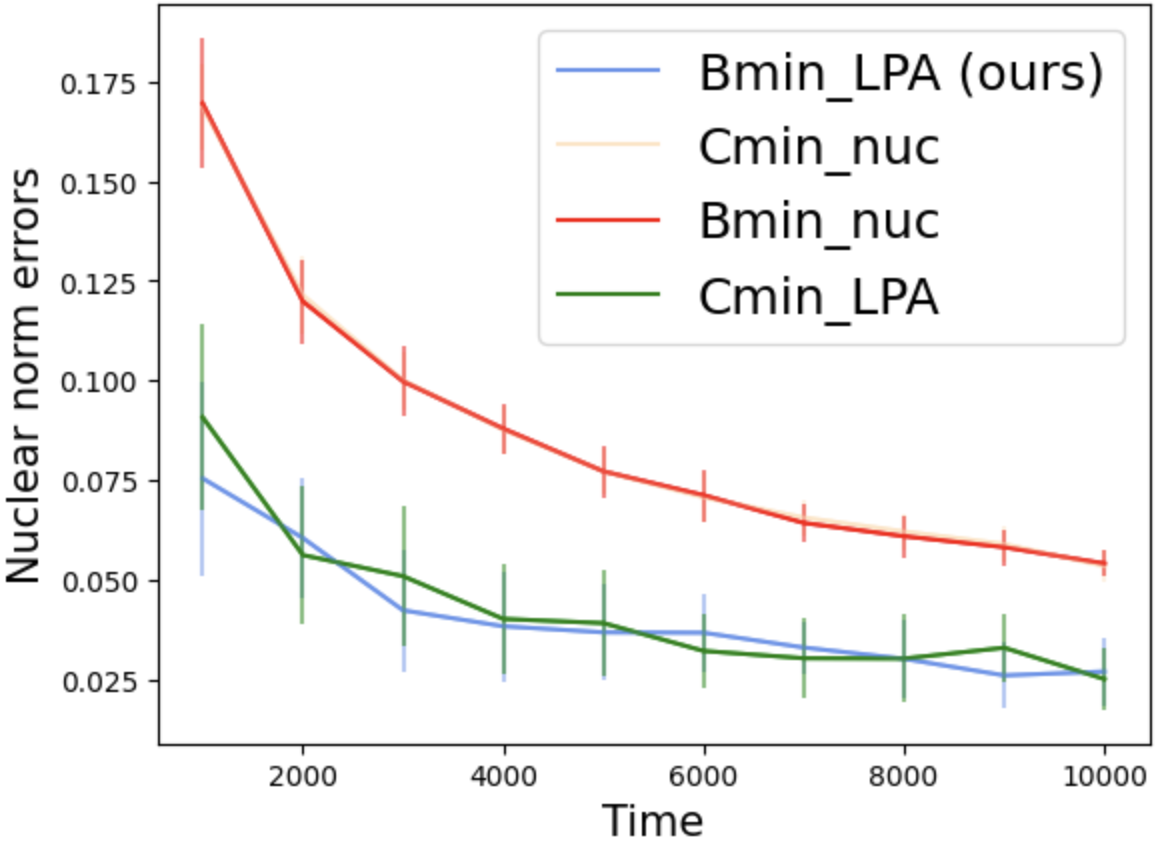}
                 &
                 \includegraphics[width=0.45\linewidth]{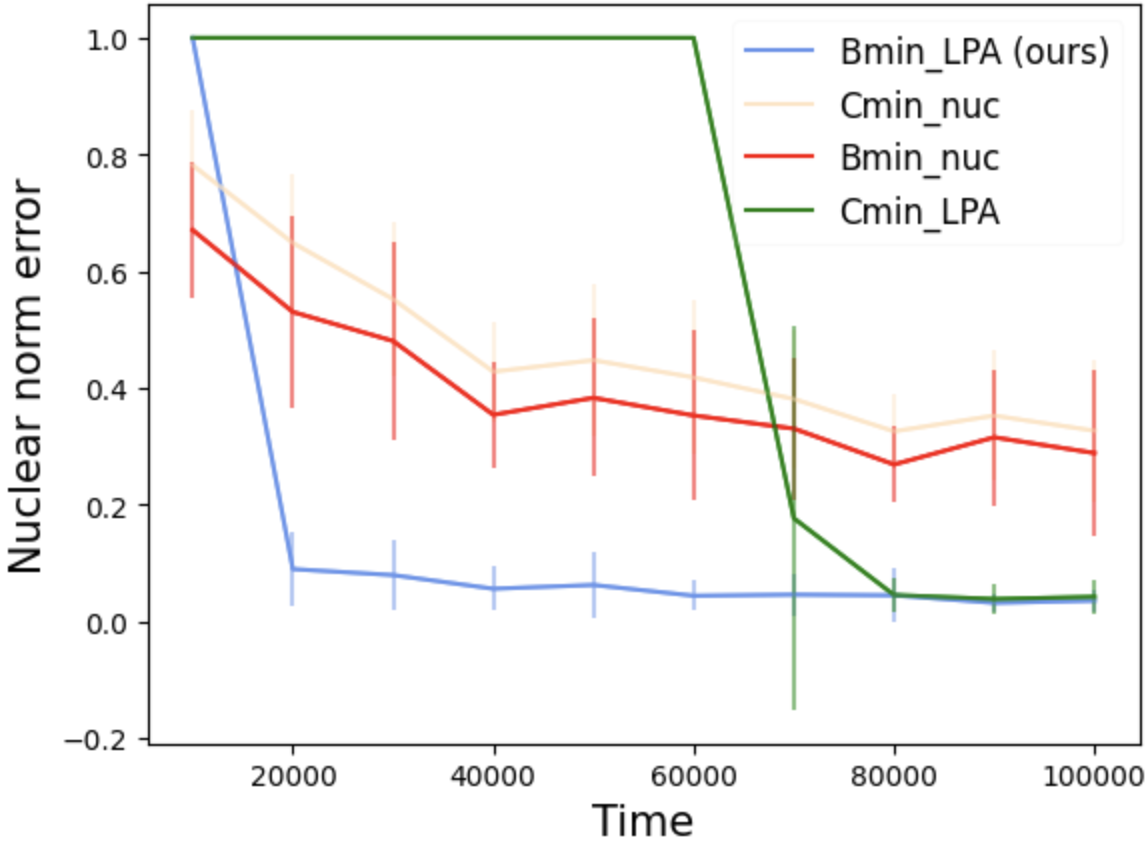}
    \end{tabular}
    \vspace{-1.5em}
\caption{Experiment results on nuclear norm error %We run \popart with the solution of Eq. \ref{def: H2}. 
      %Cmin-Lasso is the algorithm of \cite{hao2020high} that solves ... Eq. xx, and H2-Lasso is the algorithm that performs exploration based on the solution of Eq. \ref{def: H2}.
    %   \kj{increase the resolution; you can do it by exporting the plot to pdf from python and then use pdf viewer to enlarge it and then screencapture it} 
    }
    \label{fig:table_of_figures}
\end{figure}

\paragraph{Low-rank matrix recovery.}
%\chicheng{What is the ground truth $\Theta^*$?}
Figure~\ref{fig:table_of_figures} presents the results {on the nuclear norm recovery error (y-axis) as a function of the sample size (x-axis)}. {In this matrix recovery experiments, $d_1 = d_2 = 3$.}
The prefix of each line (Cmin, Bmin) represents the experimental design for the sampling distribution (optimal solutions of Eq. \eqref{eq: def of Cmin} and Eq. \eqref{eq: optimization of blambda}, respectively). 
The suffix (LPA, \textsf{nuc}) indicates the estimation method employed (\lpopart and nuclear norm regularized least squares, respectively.) 
In the left plot, \edit{}{Arm set $\cA$ has 150 arms and the elements of $\cA$ are drawn uniformly at random from $\cB_{\mathrm{Frob}}(1)$. } 
%\chicheng{What is the size of $\cA$? 100? Also I think it would be good to mention $d_1, d_2$.
%It would be good to mention that the plots are generated by averaging over X number of \emph{random instances}
%}
%we draw arm matrices uniformly at random from $\cB_{\mathrm{Frob}}(1)$.
{In the right figure, we consider the arm set $\cA_{\text{hard}}$ from Lemma \ref{lem: comparison between ldyad and cmin} that has a significant disparity between $\ldyad$ and $C_{\min}(\cA)$ values (see Appendix \ref{appsec: ldyad vs cmin, and Ahard} for the definition).} 

As one can see in the above figures, in all cases, $\ldyad$ based exploration generally outperforms naive E-optimal design, and \lpopart tends to show a better nuclear norm recovery error than \textsf{nuc}.
\begin{figure}[h]        
    \centering
\begin{tabular}{ll}
    \includegraphics[width=0.45\linewidth]{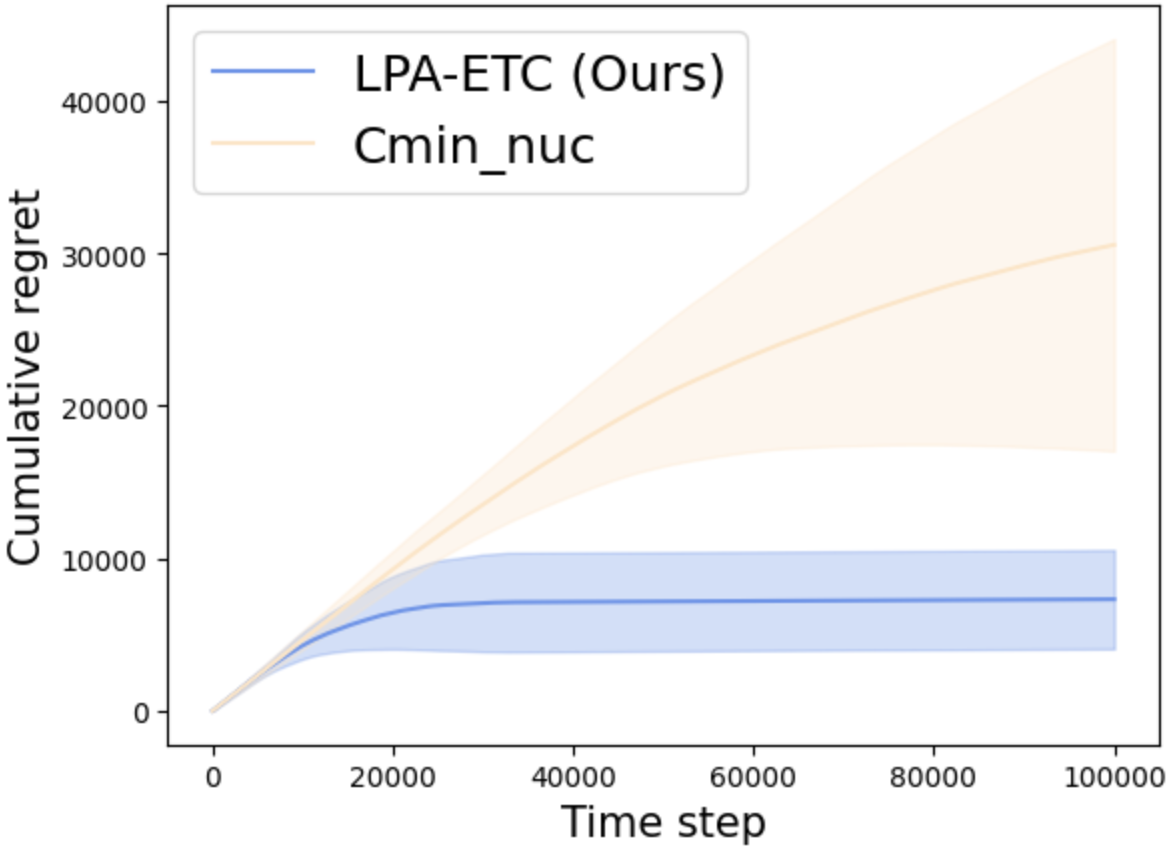}
    &\includegraphics[width=0.45\linewidth]{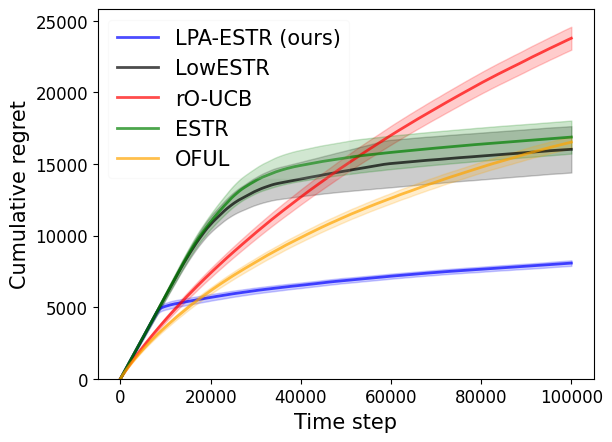}
    \end{tabular}
    \vspace{-1.5em}
    \caption{Experiment results on bandits with ETC-based (left) and ESTR-based algorithms (right)}
    \label{fig:matrix-bandits}
\end{figure}

% \chicheng{Again I think it will be beneficial to briefly mention the setups.
% This will help answer questions like: why ETC based algorithms are evaluated in one environment, and ESTR based algorithms are evaluated in another?
% }

\paragraph{Low-rank matrix bandits.} \edit{}{We consider a low-rank bandit setting with $d_1 =d_2 =5$, and for each experiment arm set $\cA$ has 100 elements which are drawn from $\texttt{Unif}(\cB_{Frob}(1))$.}
Figure ~\ref{fig:matrix-bandits} presents the results of applying \lpopart-based algorithms (Algorithm \ref{alg:etc-lowrank-dr} and \ref{alg:estr-lowrank-dr}) to the low-rank bandit problem. The first graph (left) compares Algorithm \ref{alg:etc-lowrank-dr} with another ETC-based algorithm, which is based on nuclear norm regularized least squares. Please check Appendix \ref{appsec: experiment settings} for the pseudocode of this algorithm. 
Algorithm \ref{alg:etc-lowrank-dr} achieves a significantly lower regret with {a much shorter exploration length}, demonstrating more stable results than nuclear norm regularization.

%we used , and since the original ESTR algorithm \citep{jun19bilinear} only works for the bilinear setting, the arm set $\cA$ follows the bilinear setting. More specifically,  are drawn 
\edit{}{We next consider a bilinear bandit setting 
where the arm set has structure $\cA=\{xz^\top: x\in\cX, z\in\cZ\}$. We draw $\cX$ and $\cZ$ uniformly at random from the $\mathbb{S}^{d_1 -1}$ and $\mathbb{S}^{d_2 -1}$, respectively, with $d_1 =d_2=6$.
}
The second graph (right) compares our Algorithm \ref{alg:estr-lowrank-dr} with state-of-the-art algorithms based on OFUL, such as ESTR \citep{jun19bilinear}, rO-UCB \citep{jang21improved}, LowESTR\citep{lu2021low}, and OFUL on the flattened $d_1d_2$-dimensional linear bandit problem itself \citep{ay11improved}.  Once again, it is apparent that our LPA-ESTR (Algorithm \ref{alg:estr-lowrank-dr}) outperforms other OFUL based algorithms, showing lower and more stable cumulative regret. For more experiments, \edit{such as checking the necessity}{including validating the utility} of \lpopart's thresholding step and a real-world dataset experiment, see Appendix \ref{appsec: experiment settings}. 
%with {a considerably shorter exploration length.}
\section{Lower bound}
\label{sec:lower-bound}

%\chicheng{Kyoungseok and Kwang, please take a look at this section when you have time -- thanks!}

%$\tilde{O}( \sigma^{2/3} R_{\max}^{1/3} r^{2/3} d^{1/3} T^{2/3} \cC_{\min}(\cA)^{-1/3})$,
We show via the following theorem that in Algorithm~\ref{alg:etc-lowrank-dr}'s regret upper bound~\eqref{eqn:reg-ub-etc}, its
dependence on some structural parameters of the action set is fundamental.

%Theorem~\ref{thm:etc regret bound}
%\cC_{\min}(\cA)^{-1/3}

% \chicheng{I think the original bound is expressed using $\rmax$ instead of $S_*$ -- do we have a preference on one or another?
% (I slightly prefer $\rmax$ since~\eqref{eqn:reg-ub-etc} depends on $\rmax$).
% }

\begin{theorem}
\label{thm:lower-bound-lr}
For any $d$, $r$ such that $2r-1 \leq d-1$,
$T \geq 1$, 
%$\kappa \in [ \frac{\sqrt{r}}{d} , \frac{1}{\sqrt{d}}]$, 
$C \in [\frac{r}{d^2} , \frac{1}{d}]$, 
$\sigma > 0$, 
%$R_{\max} \in [ \sigma \sqrt{ \frac{r}{n\kappa^2}}, \sigma \sqrt{ \frac{d^6 \kappa^4}{n r^2} } ]$,
$\rmax \in [ \sigma \sqrt{ \frac{r}{T C}}, \sigma \sqrt{ \frac{d^6 C^2}{T r^2} } ]$,
any bandit algorithm $\cB$, there exists a $(2r-1)$-rank $d$-dimensional bandit environment with $\sigma$-subgaussian noise, action space $\cA \subset \cbr{ a: \| a \|_{\op} \leq 1 }$ such that $C_{\min} (\cA) \geq C$, and $\Theta^*$ which satisfies $\max_{A\in \cA} |\inner{\Theta^*}{A}|\leq \rmax$ such that 
\[
\EE_{\Theta, \cB}[\Reg(\Theta, T)]
\geq 
\Omega(\sigma^{2/3} \rmax^{1/3}  r^{1/3} T^{2/3} C^{-1/3})
\]
\end{theorem}
%, and $\max_{a \in \cA} |\inner{\Theta}{a}| \leq R_{\max}$
Specifically, the theorem implies that, we cannot hope to design an algorithm with a regret bound of say, $\tilde{O}( \sigma^{2/3} \rmax^{1/3} r^{2/3} d^{1/3} T^{2/3})$, without dependence on $C_{\min}(\cA)$. To see this, we choose $C = \Theta(\frac{r}{d^2})$ in Theorem~\ref{thm:lower-bound-lr}, which yields a regret lower bound of $\Omega(\sigma^{2/3} \rmax^{1/3} d^{2/3} T^{2/3})$, which is $\gg \tilde{O}( \sigma^{2/3} \rmax^{1/3} r^{2/3} d^{1/3} T^{2/3})$ when $d \gg r^2$. 

\begin{remark}
    In Theorem \ref{thm:lower-bound-lr}, for the sake of clarity in notation, the lower bound was expressed in terms of $C$ which is a lower bound of $C_{\min}(\cA)$. However, if one desires a lower bound based on $\ldyad$, one can simply substitute every $C$ in Theorem \ref{thm:lower-bound-lr} with $\frac{d}{B}$. \edit{Then}{This is because,} by Lemma \ref{lem: comparison between ldyad and cmin}, \edit{$\ldyad\leq B$}{$C_{\min}(\cA) \geq \frac{d}{B}$ implies that $\ldyad\leq B$} 
    \edit{and a lower bound based on \( B \) can be obtained, which is as follows}{the lower bound in terms of $B$ is as follows}:
    $$ \EE_{\Theta, \cB}[\Reg(\Theta, T)]
\geq 
\Omega(\sigma^{2/3} \rmax^{1/3}  r^{1/3} T^{2/3} B^{1/3}d^{-1/3})$$

\edit{}{Compared with Theorem~\ref{thm:estr bound}'s upper bound, there is a 
$(rd)^{1/3}$ gap between the upper and lower bounds. We conjecture that our upper bound is tight and lower bound is loose. Indeed, our lower bound construction follows the construction in~\cite{hao2020high} which reduces regret lower bound to lower bounding error of a two-hypothesis testing problem; it would be interesting to see if better lower bounds can be developed using advanced techniques such as~\citet{lattimore21bandit,jang22popart}.
}
\end{remark}

\paragraph{Comparison with prior work.} By a direct adaptation of regret lower bound for $d$-dimensional stochastic bandits with unit-ball action spaces to the $dr$-dimensional setting, ~\cite{lu2021low} shows a regret lower bound of $\Omega( \sigma d r\sqrt{T} )$ for rank-$r$ matrix bandit for the action space $\cA$ being the unit Frobenius ball. 
A close examination of their lower bound reveals that, 
their lower bound fits into our Assumption~\ref{assumption:nuc-bound} with $S_* \geq \Omega(\frac{d r}{\sqrt{T}})$.
As our lower bound allows $S_*$ to take values as small as $\sigma \sqrt{\frac{r}{TC}}$, which in turn can be as small as\edit{$\sigma \sqrt{\frac{(dr)^3}{T}}$}{$\sigma \sqrt{\frac{dr}{T}}$},
\chicheng{double check -- this is $\sigma \sqrt{\frac{dr}{T}}$?}
our lower bound covers different regimes of parameter settings from~\cite{lu2021low}, which is of independent interest. 

\section{Conclusion}
We have proposed a novel low-rank estimation algorithm called \lpopart, along with a novel experimental design that aims at minimizing \lpopart's recovery guarantees. This new algorithm utilizes the geometry of the arm set to conduct estimation in a different manner than conventional approaches. Based on \lpopart, we have designed two low-rank bandit algorithms with general arm sets, improving the dimensionality dependence in regret bounds. 

%Lastly, we show via a lower bound that our upper bound's dependence on the arm set structure parameter $C_{\min}(\cA)$ is unavoidable.

%, by presenting a lower bound.
%$\Omega(T^{2/3})$

Although general, one drawback of our algorithms is that, when applied to special arm sets (e.g. the unit Frobenius norm ball), its guarantees are inferior than algorithms designed specifically for these settings~\citep{lattimore21bandit,huang2021optimal}.
Designing algorithms that can match these guarantees in these specialized settings while maintaining generality is an interesting future direction. 
Another interesting open question is establishing regret lower bound that depends on the geometry of the arm set in the low-rank bandit problem.

\section*{Acknowledgements}

Much of the work was done while the first author was at the University of Arizona and New York University. We thank Yue Kang for insightful discussions about \cite{kang2022efficient}, and the ICML reviewers for their valuable feedback.
Chicheng Zhang acknowledges support by the University of Arizona FY23 Eighteenth Mile TRIF Funding. 
Kwang-Sung Jun was supported in part by the National Science Foundation under grant CCF-2327013.

\section*{Impact statement}

This paper presents work whose goal is to advance the field of Machine Learning. There are many potential societal consequences of our work, none which we feel must be specifically highlighted here.
%\edit{}{Also, as far as we know, there is no known lower bound depending on the geometry of the arm set in low-rank bandit studies, we leave this as an interesting open question.} 

% For the future work, one possible adaptation is how to dominate the special case algorithms such as \citet{lattimore21bandit} and \citet{huang2021optimal}. 

% It turns out in their specific setup, our algorithm is not performing as well as those algorithms. Therefore, we can say there's a gap for improvement for this low-rank bandit studies. 

\bibliographystyle{icml2024}
\bibliography{main}
% \putbib[sections/bibliography]
% \bibliographystyle{icml2024}

% \end{bibunit}
%%%%%%%%%%%%%%%%%%%%%%%%%%%%%%%%%%%%%%%%%%%%%%%%%%%%%%%%%%%%%%%%%%%%%%%%%%%%%%%
%%%%%%%%%%%%%%%%%%%%%%%%%%%%%%%%%%%%%%%%%%%%%%%%%%%%%%%%%%%%%%%%%%%%%%%%%%%%%%%
% APPENDIX
%%%%%%%%%%%%%%%%%%%%%%%%%%%%%%%%%%%%%%%%%%%%%%%%%%%%%%%%%%%%%%%%%%%%%%%%%%%%%%%
%%%%%%%%%%%%%%%%%%%%%%%%%%%%%%%%%%%%%%%%%%%%%%%%%%%%%%%%%%%%%%%%%%%%%%%%%%%%%%%
\newpage
\appendix
\onecolumn

\addcontentsline{toc}{section}{Appendix} % Add the appendix text to the document TOC
\part{Appendix} % Start the appendix part
\parttoc % Insert the appendix TOC

\section{Additional Related Work}
\label{sec:relwork}

% for each time step the agent choose a pair of arm $x_t \in \mathcal{X}\subset \RR^{d_1}$ and $z_t \in \mathcal{Z} \subset \RR^{d_2}$, and receives a noisy reward whose expected value is the bilinear product $x_t^\top \Theta^* z_t$
\paragraph{Low-rank bandits with general arm sets} The first low-rank bandit algorithm that can work with a broad range of arm sets is proposed by \citet{jun19bilinear}. 
They studied the bilinear bandit model, where the arm set $\cA$ is of the form $\cbr{x z^\top, x \in \cX, z \in \cZ}$, and $\cX, \cZ$ are subsets of {$\cbr{x \in \RR^{d_1}: \| x \|_2 \leq 1}, \cbr{z \in \RR^{d_2}: \| z \|_2 \leq 1}$}, respectively. They proposed the Explore-Subspace-Then-Refine algorithm that has a regret of $\tilde{O}(\sqrt{\frac{rdT}{\lambda_{\min}(Q(\pi))}}\frac{\lambda_{\max}(\Theta^*)}{\lambda_{\min}(\Theta^*)})$; 
this is the first algorithm that enjoys regret rate improvements over the naive rate of $\tilde{O}(d^2 \sqrt{T})$ obtained by a direct reduction to $d_1 d_2$-dimensional linear bandits, which ignores the low-rank structure. 
%\textsf{tr}(\Theta^* (x_t z_t^\top)^\top )
\citet{lu2021low} extended the bilinear arm set to generic matrix arm sets
% this work to low-rank generalized linear bandit setting and consider general arm sets $\cA \subset \RR^{d_1 \times d_2}$, 
and proposed LowLOC, a computationally inefficient algorithm with $\tilde{O}(\sqrt{r d^3 T})$ regret and a computationally efficient algorithm LowESTR with $\tilde{O}(\sqrt{r d^3 T} / \lambda_{\min})$ regret. 
They also proved a $\Omega(r d \sqrt{T})$ regret lower bound for this setting. 
\citet{kang2022efficient} designed low-rank bandit algorithms by combining Stein's method for matrix estimation and the Explore-Subspace-Then-Refine framework of~\cite{jun19bilinear}, assuming the existence of a nice exploration distribution over the arm set; their regret bound is $\tilde{O}(\sqrt{r d^2 M T}/\lambda_{\min})$, where $M$ is an arm set-dependent constant.
%\chicheng{Emphasize that their setting is the contextual bandits with changing arm set setting, which is different from ours. (To me, this is the main reason we don't need to compare with this algorithm in the table.} \ks{I just checked their result, and seems like they have $\frac{1}{\cC_{\min}^{2/3}}$ dependence. This means our algorithm is usually stronger than theirs. To be precise, they are depending on the compatibility constant.}
%These works assume the existence of some nice sampling distribution over the arm set, 
%It should be noted that in the \citet{kang2022efficient} didn't address enough on the bound of the sampling distribution related constant 
However, the $M$ from their given example can have hidden dimensionality dependence -- when specialized to the setting of $\cA$ being the unit Frobenius norm ball, it is of order $d_1 d_2$, which induces higher regret compared to the previous works with general arm sets \citep{jun19bilinear, lu2021low}. See Appendix \ref{Appendix:counter examples} for a detailed derivation. {In addition, there is no known method to optimize $M$. }{As far as we know, \cite{kang2022efficient} is the first low-rank bandit paper that applies the techniques of \cite{minsker2018sub}. For the Catoni's estimator, several studies use Catoni's estimator to get a variance-dependent bound on regret bound, such as \cite{camilleri2021high,mason2021nearly}.}

%\chicheng{I think it is better to have a placeholder section in the appendix so that we can make this clear when we work on the supplementary.} 

%especially \citet{li2022simple} assumed it is given from the nature and
%a unifying framework of the

%, in that we study low-rank bandits 

%which has same effect as assuming the existence of the fixed random distribution of the arms. 

\paragraph{Low-rank bandits with specific arm sets} There {have been} lots of other variants of the low-rank bandit, exploiting more specific structures. Some researchers \cite{katariya17bernoulli, trinh2020solving, jedra2024low} mainly focused on low-rank bandit problems with canonical arms, which means $\cA = \cbr{e_i e_j^\top: i \in [d_1], j \in [d_2]}$; \citet{katariya17bernoulli} and \citet{trinh2020solving} even added rank-1 assumption on $\Theta^*$ over this setting.
%\chicheng{By this we mean arms that are canonical basis?} 
%\chicheng{I see, can we make this explicit, say $\cA = \cbr{e_i e_j^\top: i \in [d_1], j \in [d_2]}$? Thanks!} 
\citet{kveton17stochastic_arxiv} studied about low-rank bandit where the hidden matrix is a hott topic matrix and arm set is $\{ UV^\top : U^\top=[u_1; u_2; \cdots; u_r], u_i\in \Delta([d_1]), V^\top=[v_1; v_2; \cdots; v_r], v_i\in \Delta([d_2]) \}$. 
where $[u_1; u_2; \cdots; u_r]$ refers to concatenation of $r$ vectors to create a matrix. 
%\chicheng{Make the assumption explicit.. Thanks!}\ks{Okay, I will check the assumption soon. }
%Compared to these results, ours are more general in that we allow for generic arm sets. 
\citet{kotlowski19bandit, lattimore21bandit, huang2021optimal} studied the low-rank bandit with a sphere or unit ball arm set. Though \citet{lattimore21bandit} and \citet{huang2021optimal} dramatically improved the regret bounds (see Table~\ref{table:comparison}), as \citet{rusmevichientong10linearly} have pointed out, the curvature property of the arm set~\cite{huang2016following} can help the agent to improve the regret bound - the regret bound of ETC can be $\sqrt{T}$ when the arm set satisfies certain curvature property. {We show in Appendix \ref{Appendix:counter examples} that even when the arm set is modified slightly, the regret analysis in these works may no longer go through.}
In contrast, our algorithm is applicable to general arm sets. 

\paragraph{Low-rank contextual bandits with time-varying arm sets} \citet{li2022simple} studied  high-dimensional contextual bandits where at each time step,  the set of available arms are drawn iid from some fixed distribution; when specialized to the low-rank linear bandit setting, their setup is different ours due to the nature of time-varying arm sets in their work.

%so it is not easy to compare the performances directly, and 

\paragraph{Sparse linear bandits} As previously discussed in Section \ref{sec:intro}, the algorithm presented in this paper draws inspiration from sparse linear bandit algorithms. Reserachers have made significant development on the field of sparse linear bandit algorithms, e.g. \cite{hao2020high,jang22popart}. These papers extensively utilize the geometry of the arm set and effectively mitigate the dependence on dimensionality in the regret bound.
\paragraph{Low-rank matrix estimation} It is natural to apply efficient low-rank matrix recovery results for solving low-rank bandit, since smaller estimation error leads fewer samples for exploration which leads smaller cumulative regret in bandit problems. \citet{keshavan10matrix} provides recovery guarantees for projection based rank-$r$ matrix optimization for matrix completion, and \citet{rohde2011estimation,tsybakov2011nuclear} provide analysis of nuclear norm regularized estimation method for general trace regression, with~\citet{rohde2011estimation} providing further analysis on the (computationally inefficient) Schatten-$p$-norm penalized least squares method. In this paper, we mainly use the robust matrix mean estimator of \cite{minsker2018sub} us it to provide efficient matrix recovery.

\section{Proof of Section \ref{sec:new analysis}}\label{appsec: proof of new analysis section}

\subsection{Proof of Theorem \ref{thm:lpopart-theta1}}

\begin{proof}
    First, we recall the following lemma of \citet{minsker2018sub} on robust matrix mean estimation: 
    \begin{lemma}[Modification of Corollary 3.1, \citet{minsker2018sub}]\label{lem:from Minsker}
        For a sequence independent, identically distributed random matrices $(M_i)_{i=1}^n$, let 
        $$\sigma_n^2 = \max\del{\norm{\sum_{i=1}^n \EE [M_i M_i^\top ]}_{\op}, \norm{\sum_{i=1}^n \EE [M_i^\top M_i] }_{\op}}$$
        Given $\nu=\frac{t\sqrt{n}}{\sigma_n^2}$, let $X_i = \phi (\nu \cH(M_i ))$ and let $\hat{T}=\frac{1}{n \nu} (\sum_{i=1}^n 
    X_i)_{\mathsf{ht}}$. Then, with probability at least $1-2(d_1 +d_2)\exp\del{-\frac{t^2 n }{2\sigma_n^2}}$,
    \begin{align*}
        \| \hat{T} - \EE[M_i]\|_{\op}\leq \frac{t}{\sqrt{n}}
    \end{align*}
    \end{lemma}

To utilize this Lemma \ref{lem:from Minsker}, we choose $M_i$'s so that
\begin{itemize}
    \item $\EE\sbr{M_i} = \Theta^* - \Theta_0$ so that $\hat{T}$ estimates the hidden parameter $\Theta^*-\Theta_0$
    \item $\sigma_{m}^2$ is well-controlled. 
\end{itemize}

It can be checked that $M_i = \reshape \del{\tilde{\Theta}_i}$ satisfies the condition with $\sigma_{n}^{2} \leq 2(\sigma^2 + R_{0}^2) {\blambda n_0}$ (See Appendix \ref{Appendix: proof of main theorem} for the proof). Substituting $\sigma_{n}^2$ by $2(\sigma^2 + R_{0}^2)\blambda n_0$, and setting $t=\sqrt{\frac{2\sigma_n^2}{n_0}\ln \frac{2d}{\delta}}$ leads the desired result.  
\end{proof}

\subsection{Proof of \texorpdfstring{$\sigma_n^2 \leq 2 (\sigma^2 + R_0^2) \blambda$}{} in Theorem~\ref{thm:lpopart-theta1}}\label{Appendix: proof of main theorem}

\begin{lemma}
\label{lem:bounding-sigma-n}
    $$\sigma_n^2 = \max (\sum_{i=1}^n \|\EE[M_i M_i^\top ]\|_{\op},\sum_{i=1}^n \|\EE[M_i^\top M_i ]\|_{\op})\leq {2n\blambda(\sigma^2 + R_0^2)}$$
\end{lemma}
\begin{proof}
Note that $M_i=\reshape \del{Q(\pi^*)^{-1} (Y_i - \inner{\Theta_0}{X_i}) \ve(X_i)}$, and all $M_i$ are i.i.d. Therefore, $\sigma_n^2 = n \cdot \max(\|\EE[M_1 M_1^\top ]\|_{\op},\|\EE[M_1^\top M_1 ]\|_{\op})$, and to compute the first term in the max, 
    \begin{align*}
    \EE[M_i M_i^\top ]&= \EE\sbr{ (Y_i - \inner{\Theta_0}{X_i})^2 \reshape\del{ Q(\pi)^{-1} \ve(X_i) } \reshape\del{ Q(\pi)^{-1} \ve(X_i) }^\top } \\
    &\preceq 2\EE\sbr{ (\eta_i^2 + \inner{\Theta_0- \Theta^*}{X_i}^2) \reshape\del{ Q(\pi)^{-1} \ve(X_i) } \reshape\del{ Q(\pi)^{-1} \ve(X_i) }^\top } \\
    &\preceq 
    2(\sigma^2 + R_{0}^2) \cdot \EE\sbr{ \reshape\del{ Q(\pi)^{-1} \ve(X_i) } \reshape\del{ Q(\pi)^{-1} \ve(X_i) }^\top }\\
%    &=
%    2(\sigma^2 + R_{0}^2) \cdot \EE\sbr{ \reshape\del{ Q(\pi)^{-1} \ve(X_i) } \reshape\del{ Q(\pi)^{-1} \ve(X_i) }^\top }
    \end{align*}
    where the first inequality holds since $(Y_i-\inner{\Theta_0}{X_i})^2 = (\eta_i+\inner{\Theta^*-\Theta_0}{X_i})^2 \leq 2 \eta_i^2 + 2\inner{\Theta^*-\Theta_0}{X_i}^2$.  Now the main task is how to compute $\|\EE\sbr{ \reshape\del{ Q(\pi^*)^{-1} \ve(X_i) } \reshape\del{ Q(\pi^*)^{-1} \ve(X_i) }^\top }\|_{\op}$. Here, we will simply use the definition of the operator norm.
  \begin{align*} 
        & \norm{\EE\sbr{ \reshape\del{ Q(\pi^*)^{-1} \ve(X_i) } \reshape\del{ Q(\pi^*)^{-1} \ve(X_i) }^\top }}_{\op} \\
        &= \max_{u \in \mathbb{S}^{d_1-1}} u^\top \EE\sbr{ \reshape\del{ Q(\pi^*)^{-1} \ve(X_i) } \reshape\del{ Q(\pi^*)^{-1} \ve(X_i) }^\top } u\\
        &{= \max_{u \in \mathbb{S}^{d_1-1}} u^\top \EE\sbr{ \reshape\del{ Q(\pi^*)^{-1} \ve(X_i) } \cdot \del{\sum_{i=1}^{d_2} e_i^{d_2} (e_i^{d_2})^\top } \cdot \reshape\del{ Q(\pi^*)^{-1} \ve(X_i) }^\top } u}\\
        &= \max_{u \in \mathbb{S}^{d_1-1}} \EE \sbr{ \sum_{i=1}^{d_2} \langle (e_i^{d_2} \otimes u), \del{ Q(\pi^*)^{-1} \ve(X_i) } \rangle^2 } \\
        &=\max_{u \in \mathbb{S}^{d_1-1}} \sbr{ \sum_{i=1}^{d_2} (e_i^{d_2} \otimes u)^\top Q^{-1} (\pi) (e_i^{d_2} \otimes u) }
        \\&=\max_{u \in \mathbb{S}^{d_1-1}} \sbr{ u^\top \del{\sum_{i=1}^{d_2} D_i^{(\mathrm{col})}} u}= \lambda_{\max}(\sum_{i=1}^{d_2} D_i^{(\mathrm{col})})
    \end{align*}
    Therefore, we can conclude $ \|\EE [M_i M_i^\top ]\|_{\op}\leq (\sigma^2 + R_0^2) \lambda_{\max}(\sum_{i=1}^{d_2} D_i^{(\mathrm{col})})$, and similarly $ \|\EE [M_i^\top M_i ]\|_{\op}\leq d_1 (\sigma^2 + R_0^2) \lambda_{\max}(D_i^{(\mathrm{row})})$. Thus, $$ \sigma_n^2 \leq 2\max \del{\lambda_{\max}(\sum_{i=1}^{d_2} D_i^{(\mathrm{row})}),\lambda_{\max}(\sum_{i=1}^{d_1} D_i^{(\mathrm{col})})} (\sigma^2 + R_0^2) n= 2B(Q)(\sigma^2 + R_0^2) n.$$
    This concludes the proof.
\end{proof}

\subsection{Proof of Theorem \ref{thm:lpopart}}

\begin{proof}
Note that for all $j \geq r+1$, $\sigma_j(\Theta^*) = 0$.
    By Weyl's Theorem \citep{horn2012matrix}, for all $j \geq r+1$, we have that $\sigma_j(\Theta_1) \leq 2\sqrt{\frac{((\sigma^2 + R_{0}^2) )\blambda \del{\ln \frac{2d}{\delta} } }{n_0}}=\threshold$. As a consequence, $\hat{\Theta}$ has rank at most $r$. 
    
    Moreover, by construction, $\| \hat{\Theta} - \Theta_1 \|_{\op} \leq \threshold$. By triangle inequality, we have
    $\| \hat{\Theta} - \Theta^* \|_{\op} 
    \leq 
    2\threshold$.
    \qedhere
\end{proof}

\subsection{Proof of Corollary \ref{cor:nuc anf frob}}

\begin{proof}
    For any matrix $M$, $\|M\|_*\leq r\|M\|_{\op}$ and $\|M\|_*\leq \sqrt{r}\|M\|_{F}$. Substitute $M$ to $\hat{\Theta}-\Theta^*$ leads the desired property.  
\end{proof}

\subsection{Proof of Theorem \ref{thm:wlpopart}}
\begin{proof}
    By Corollary \ref{cor:nuc anf frob}, the assumption $n_0 \geq \tilde{O}\del{ r^2 \blambda \cdot (\frac{\sigma + S_*}{\sigma})^2 }$ guarantees that $\|\Theta_0 - \Theta^*\|_*\leq O(\sigma)$ where $\Theta_0$ is the pilot estimator in Line 2 of Algorithm \ref{alg:warm-popart}. Therefore, $\max_{A \in \cA} |\inner{\Theta_0 - \Theta}{A}|\leq \max_{A \in \cA} \|\Theta_0 - \Theta\|_*\|A\|_{\op} \leq O(\sigma)$. We can get our final result by substituting $R_0$ to $O(\sigma)$ in Theorem \ref{thm:lpopart}.
\end{proof}

\section{Proofs of Lemma \ref{lem: Upper bound Bmin} and \ref{lem: comparison between ldyad and cmin} }\label{appsec: ldyad vs cmin, and Ahard}

\subsection{Preliminaries - Relationship between \texorpdfstring{$D_i^{(\mathrm{col})}$ and $D_i^{(\mathrm{row})}$}{}}\label{appsec:prelim of dcol and drow}

In Figure \ref{fig:q-d_1}, $D_i^{(\mathrm{col})}$ and $D_i^{(\mathrm{row})}$ looks quite different. However, it turns out that they are coming from the similar logic, due to the nature of the low-rank bandit problem.

Recall the definition of the low-rank bandit problem. For each time, the agent pulls action $A_t \in \RR^{d_1 \times d_2}$ and receives reward $\langle \Theta^*, A_t \rangle + \eta_t$. However, one could simply transpose all the actions and define $\cA^\top := \{ a^\top: a\in \cA\}$, and think of the reward as $\langle (\Theta^*)^\top, A_t^\top \rangle + \eta_t$. This does not change the nature of the problem. The definition of $D_i^{(\mathrm{col})}$ and $D_i^{(\mathrm{row})}$ comes from this fact.

To compare the original low-rank bandit problem with 'transposed version' of the low-rank bandit problem, let $Q_{trans}(\pi):= \EE_{a\sim \pi} [\ve(a^\top) {\ve(a^\top)^\top}]$. Then, the following properties also hold:

\begin{itemize}
    \item $\ve(a) = P \ve(a^\top)$ for a fixed permutation matrix $P \in \mathbb{R}^{d_1 d_2 \times d_1 d_2}$.
    \item $\lambda_{\min}(Q) = \lambda_{\min}(Q_{trans})$ since $Q_{trans} = P^\top Q P$.
    \item One could check $D_{i}^{(row)}(Q) = D_i^{(col)} (Q_{trans})$ and $D_{i}^{(col)}(Q) = D_i^{(row)} (Q_{trans})$.
\end{itemize}
Which means, though $D_i^{(\mathrm{col})}$ and $D_i^{(\mathrm{row})}$ looks quite different, $D_i^{(\mathrm{row})}$ is the matrix that come from the same logic as $D_i^{(\mathrm{col})}$, but from the transposed problem. 

Therefore, from now on, we will only compute $D_i^{(\mathrm{col})}$ related quantity for the scale comparison in this Section \ref{appsec: ldyad vs cmin, and Ahard}. 

\subsection{Proof of Lemma \ref{lem: Upper bound Bmin}}
\begin{proof}
For any vector $v \in \RR^{d_1}$, define $\mathsf{Ext}(v,i) \in \mathbb{R}^{d_1 d_2}$ as follows:
\begin{align*}
    \mathsf{Ext}(v,i) := e_i^{d_2} \otimes v %\ve([\underbrace{\overrightarrow{0}_{d_1}; \overrightarrow{0}_{d_1}; \cdots; \overrightarrow{0}_{d_1}}_{i-1\text{ columns}}; v; \overrightarrow{0}_{d_1}; \cdots \overrightarrow{0}_{d_1}])
\end{align*}

Then, 
\begin{align*}
    \lambda_{\max}(\sum_{i=1}^{d_2} D_i^{(\mathrm{col})}) &\leq \sum_{i=1}^{d_2} \lambda_{\max}(D_i^{(\mathrm{col})}) \tag{Homogeneity of degree 1 and convexity of maximum eigenvalue.}
    %\chicheng{Also homoegenity of $\lambda_{\max}(\cdot)$?}
    \\
    &=\sum_{i=1}^{d-2} \max_{v \in \mathbb{S}^{d_1 -1}} v^\top (D_i^{(\mathrm{col})})v \\
    &=\sum_{i=1}^{d-2} \max_{v \in \mathbb{S}^{d_1 -1}} \mathsf{Ext}(v,i)^\top Q^{-1} \mathsf{Ext}(v,i)\\
    &\leq \sum_{i=1}^{d-2} \max_{u \in \mathbb{S}^{d_1 d_2 -1}} u^\top Q^{-1} u\\
    &= d_2 \lambda_{\max}(Q^{-1}) = \frac{d_2}{\lambda_{\min}(Q)}
\end{align*}
and the proof follows. 
\end{proof}
\subsection{Proof of Lemma \ref{lem: comparison between ldyad and cmin}}
\newcommand{\cAhard}{\cA_{\textsf{hard}}}

In this section, we will consider a setting where $d_1 = d_2 = d$, and the following action set, $\cAhard = \{\reshape(a_1), \cdots, \reshape(a_{d^2})\} \subset \mathbb{R}^{d \times d}$ where

\begin{align*}
    a_i := \begin{cases}
         l \cdot e_1 &\text{For $i=1$}\\
        e_1 + m \cdot e_i &\text{Otherwise}
    \end{cases}
\end{align*}

Eventually, we will choose $l=\frac{1}{\sqrt{d}}, m=1$ for our final $\cAhard$, but to demonstrate the effect of each scaling factor, we will leave $l,m$  unspecified and assume $l,m \leq 1$ 
throughout this proof. 

In this subsection, we will also use following definitions for the brevity.

\begin{itemize}
    \item $D:=d^2$, 
    \item $\pi_i:= \pi(a_i)$, and $\hat{\pi}:=(\pi_1, \pi_2, \cdots, \pi_D)$ for any $\pi \in \cP(\cA)$

    \item $\mathsf{Sym}(n)$ be a permutation group of $[n]$. 
    \item For any permutation $\sigma \in \mathsf{Sym}(n)$ 
    \begin{itemize}
        \item For any $v \in \mathbb{R}^{d}$, let $\sigma(v) := (v_{\sigma(1)}, \cdots, v_{\sigma(n)})$
        \item For any $\pi \in \cP(\cA)$, define $\sigma(\pi)\in \cP(\cA)$ to be such that $\sigma(\pi)(a_i) := \hat{\pi}_{\sigma(i)}$      
    \end{itemize}
\end{itemize} for the brevity. 

Now, one could check that
\begin{align}\label{appeq: cov matrix of Ahard}
    Q(\pi)=\begin{bmatrix}l^2 \pi_1 + \sum_{i=2}^D \pi_i & m \hat{\pi}_{2:D}^\top \\m \hat{\pi}_{2:D} & m^2 \diag(\hat{\pi}_{2:D})  \end{bmatrix}
\end{align}

For the notational convenience, let $\hat{q} = (\pi_1^{-1}, \cdots , \pi_D^{-1})$. Then, 

$$Q(\pi)^{-1}=\begin{bmatrix}\frac{1}{l^2 \pi_1}& -\frac{1}{m l^2 \pi_1}\textbf{1}_{D-1}^\top \\ -\frac{1}{m l^2 \pi_1}\textbf{1}_{D-1}& \frac{1}{m^2} \diag(\hat{q}_{2:D})+ \frac{1}{l^2 m^2 \pi_1} \textbf{1}_{D-1} \textbf{1}_{D-1}^{\top} \end{bmatrix} $$

\subsubsection{Calculate \texorpdfstring{$C_{\min}(\cA_{\mathsf{hard}})$}{}}

Suppose that $\Pi^C$ is the set of optimal experimental designs for $C_{\min}$ (which means, the solution of Eq. \eqref{eq: def of Cmin}).
Below, we will show that there exists some $\pi^C$ in $\Pi^C$ such that $\pi_2^C = \cdots = \pi_D^C$.

%\chicheng{I think there can be ties for the optimal solution? What we really want to say is that, there exists some optimal solution $\pi^C$ such that $\pi_2^C = \cdots = \pi_D^C$?}

\paragraph{Prove that $\exists \pi^C \in \Pi^C$ such that $\pi_2^C = \cdots = \pi_D^C$}

Note that $\lambda_{\max}$ and the matrix inversion are both convex functions. Moreover, from the symmetry of the arm set $\cAhard$, for any permutation $\sigma' \in \mathsf{Sym}(D)$ which satisfies $\sigma'(1)=1$, for all $\pi\in \cP(\cAhard)$, $\lambda_{\max} (Q(\pi)^{-1}) = \lambda_{\max}(Q(\sigma'(\pi))^{-1})$. Let 
% \in \Pi^C
\begin{align*}
    \sigma_1 (i) =\begin{cases}
    1 & \text{if $i=1$}\\
    2 & \text{if $i=D$}\\
    i+1 & \text{Otherwise}
\end{cases}
\end{align*}

Now, fix $\pi \in \Pi^C$; Define $\pi^C$ to be
\begin{align*}
    \pi^C(a_i)
    = \begin{cases}
        \pi(a_1) & \text{if $i=1$}\\
        \frac{\sum_{s=2}^{D} \pi(a_s)}{D-1} & \text{Otherwise}
    \end{cases}
\end{align*}
Then, 
\begin{align*}
    \lambda_{\max}(Q(\pi)^{-1}) &= \frac{1}{D-1} \sum_{s=1}^{D-1}\lambda_{\max}(Q(\sigma_1^{s}(\pi))^{-1})\\
    &\geq \lambda_{\max}(Q(\frac{1}{D-1} \sum_{s=1}^{D-1}\sigma_1^{s}(\pi))^{-1}) \tag{Convexity}\\
    &\geq \lambda_{\max}(Q(\pi^C)^{-1})\\
    &\geq \lambda_{\max}(Q(\pi))^{-1}) \tag{Minimality of $\pi$}\\
\end{align*}

Therefore, $\pi^C \in \Pi^C$, and to calculate $C_{\min}(\cAhard)$, it suffices to consider only distributions $\pi \in \cP (\cAhard)$ which satisfies $\pi_2 = \pi_3 = \cdots = \pi_D$. Let $\pi_1 = a$, and $\pi_2 = b$ for brevity. 

Then, the characteristic matrix looks like this: if we let $I_n$ be the $n \times n$ dimensional identity matrix,  

\begin{align*}
    Q(\pi)- \lambda I_D = \begin{bmatrix}
    l^2 a + (D-1)b -\lambda & mb & \cdots & mb\\
    mb &  & & \\
    \vdots & & (m^2 b-\lambda)I_{D-1} & \\
    mb &  & &
    \end{bmatrix}
\end{align*}
and by the row operation, 
\begin{align*}
    \det(Q(\pi)- \lambda I_D) = \det \begin{bmatrix}
    l^2 a + (D-1)b -\lambda - (D-1) \frac{m^2 b^2}{m^2 b - \lambda} & 0 & \cdots & 0\\
    mb &  & & \\
    \vdots & & (m^2 b-\lambda)I_{D-1} & \\
    mb &  & &
    \end{bmatrix}
\end{align*}

The characteristic polynomial is therefore

$$\det(Q(\pi)-\lambda I) = (m^2 b - \lambda)^{D-2} (\lambda^2 - ((D-1)b+l^2 a + m^2 b) {\lambda} + l^2 a m^2 b) $$

We can therefore get two eigenvalues from quadratic equation, and \edit{$D-2$ repeated eigenvalue $m^2 b$}{eigenvalue $m^2 b$ has multiplicity $D-2$}.

For eigenvalues from the quadratic equation, note that for the quadratic equation of the form $\lambda^2 - B \lambda + C = 0 (B,C>0)$, the smaller eigenvalue has the order of $\Theta(\frac{C}{B})$ since $B<B+\sqrt{B^2 - 4C} < 2B$. Therefore, the order of the eigenvalue is $\Theta(\frac{l^2 a m^2 b}{((D-1)b+l^2 a + m^2 b)})$ and for the inverse, it's of order $\Theta(\frac{B}{C})$. 

When $l,m<1$, one can note that the dominating terms are $Db$ and $l^2 a$ on the denominator. Therefore,
\[
\lambda_{\max}(Q(\pi)^{-1})
=
\Theta ( \max( \frac{1}{m^2 b}, \frac{D}{m^2 l^2 a} ) )
\]

Using the fact that $a+(D-1)b=1$, one can get we get optimal rate when $\frac{a}{b} = \Theta (\frac{d}{l^2})$ and the final $C_{\min}^{-1} = \Theta(\frac{D}{m^2 l^2})$.

\subsubsection{Calculate \texorpdfstring{$B_{\min}(\cAhard)$}{}}
From Eq. \eqref{appeq: cov matrix of Ahard},

$$[Q(\pi)^{-1}]_{1:d, 1:d}=\begin{bmatrix}\frac{1}{l^2 \pi_1}& -\frac{1}{m l^2 \pi_1}\textbf{1}_{d-1}^\top \\ -\frac{1}{m l^2 \pi_1}\textbf{1}_{d-1} & \frac{1}{m^2} \diag(\hat{q}_{2:d})+ \frac{1}{l^2 \pi_1 m^2} \textbf{1}_{d-1} \textbf{1}_{d-1}^{\top} \end{bmatrix} $$

and 

$$[Q(\pi)^{-1}]_{d(i-1)+1:di, d(i-1)+1:di}=\frac{1}{m^2} \diag(\hat{q}_{d(i-1)+1:di})+ \frac{1}{l^2 \pi_1 m^2} \textbf{1}_{d} \textbf{1}_{d}^{\top}$$

for $i=2,\cdots, d$. Therefore, if we let $G_i (\pi) = \sum_{j=0}^{d-1} \frac{1}{\pi_{dj+i}}$ for $i=2,\cdots, d-1$, and $G_1 (\pi)= \sum_{j=1}^{d-1} \frac{1}{\pi_{dj+1}}$, 
\begin{equation}\label{appeqn: initial form of B(Q) in Ahard}
    \sum_{i=1}^d D_i^{(\mathrm{col})} (\pi) = \sum_{i=1}^d [Q(\pi)^{-1}]_{d(i-1)+1:di, d(i-1)+1:di} = \begin{bmatrix}\frac{1}{l^2 \pi_1}+ \frac{1}{m^2} G_1 + \frac{(d-1)^2}{l^2 m^2 \pi_1}& \frac{d-m-1}{m^2 l^2 \pi_1}\textbf{1}_{d-1}^\top \\ \frac{d-m-1}{m^2 l^2 \pi_1}\textbf{1}_{d-1} & \frac{1}{m^2} \diag(G_{2:d})+ \frac{d-1}{l^2 m^2 \pi_1} \textbf{1}_{d-1} \textbf{1}_{d-1}^{\top} \end{bmatrix} 
\end{equation}

Suppose that $\Pi^B$ is the set of optimal experimental designs for $B_{\min}$ (which means, the solution of Eq. \eqref{eq: optimization of blambda}).
Below, we will show that there exists some $\pi^B$ in $\Pi^B$ such that 
\begin{itemize}
    \item $\pi_i^B = \pi_j^B$ for all $i,j \not\equiv 1 \pmod d$
    \item $\pi_{d+1}^B = \pi_{2d+1}^B = \cdots \pi_{(d-1)d+1}^B$
\end{itemize}

\subsubsection{Proving that $\exists \pi^B \in \Pi^B$ such that $\pi_i^B = \pi_j^B$ for all $i, j\not\equiv 1 \pmod d $}

%Suppose $\pi_{di +1}$ $(i=0,\cdots, d-1$) are fixed; we look at how optimal $\pi_j (j\not\equiv 1 \pmod d)$ should behave. 

Let $G= \frac{1}{d-1}\sum_{i=2}^d G_i$, and let $\pi \in \Pi^B$. Let $\sigma$ be the permutation of $[d]$ which is defined as
$$ \sigma (n) = \begin{cases}
    1 & \text{if $n \equiv 1 \pmod d$}\\
    2 & \text {if $n \equiv 0 \pmod d$}\\
    n+1 & \text{otherwise}
\end{cases}$$

and $\rho$ be the permutation of $[D]$ which is defined as
$$ \rho (n) = \begin{cases}
    n & \text{if $n \equiv 1 \pmod d$}\\
    n-d+2 & \text {if $n \equiv 0 \pmod d$}\\
    n+1 & \text{otherwise}
\end{cases}$$

Then, 

\begin{align*}
B(Q(\pi))&=\frac{1}{d-1}\sum_{s=1}^{d-1} B(Q(\rho^s ({\pi})) \tag{From the symmetry of Eq. \eqref{appeqn: initial form of B(Q) in Ahard}}
    \\
    &=\frac{1}{d-1} \sum_{s=1}^{d-1} \lambda_{max} \del{\begin{bmatrix}\frac{1}{l^2 \pi_1}+ \frac{1}{m^2} G_1 + \frac{(d-1)^2}{l^2 m^2 \pi_1}& \frac{d-m-1}{m^2 l^2 \pi_1}\textbf{1}_{d-1}^\top \\ \frac{d-m-1}{m^2 l^2 \pi_1}\textbf{1}_{d-1} & \frac{1}{m^2} \diag(\sigma^s (G)_{2:d})+ \frac{d-1}{l^2 \pi_1 m^2} \textbf{1}_{d-1} \textbf{1}_{d-1}^{\top} \\ \end{bmatrix}} \tag{Property of permutation $\sigma$}\\
    &\geq \lambda_{max} \del{\frac{1}{d-1}\sum_{s=1}^{d-1}\begin{bmatrix}\frac{1}{l^2 \pi_1}+ \frac{1}{m^2} G_1 + \frac{(d-1)^2}{l^2 m^2 \pi_1}& \frac{d-m-1}{m^2 l^2 \pi_1}\textbf{1}_{d-1}^\top \\ \frac{d-m-1}{m^2 l^2 \pi_1}\textbf{1}_{d-1} & \frac{1}{m^2} \diag(\sigma^s (G)_{2:d})+ \frac{d-1}{l^2 m^2 \pi_1} \textbf{1}_{d-1} \textbf{1}_{d-1}^{\top} \\ \end{bmatrix}} \tag{Jensen's inequality and convexity}\\
    &= \lambda_{max} \del{\begin{bmatrix}\frac{1}{l^2 \pi_1}+ \frac{1}{m^2} G_1 + \frac{(d-1)^2}{l^2 m^2 \pi_1}& \frac{d-m-1}{m^2 l^2 \pi_1}\textbf{1}_{d-1}^\top \\ \frac{d-m-1}{m^2 l^2 \pi_1}\textbf{1}_{d-1} & \frac{1}{m^2} \diag(G\textbf{1}_{d-1})+ \frac{d-1}{l^2 m^2 \pi_1} \textbf{1}_{d-1} \textbf{1}_{d-1}^{\top} \end{bmatrix}}
\end{align*}

Plus, note that when $C' > C>0$, 
\begin{align*}
    \lambda_{\max} &\del{\begin{bmatrix}\frac{1}{l^2 \pi_1}+ \frac{1}{m^2} G_1 + \frac{(d-1)^2}{l^2 m^2 \pi_1}& \frac{d-m-1}{m^2 l^2 \pi_1}\textbf{1}_{d-1}^\top \\ \frac{d-m-1}{m^2 l^2 \pi_1}\textbf{1}_{d-1} & \frac{1}{m^2} \diag(C'\textbf{1}_{d-1})+ \frac{d-1}{l^2 m^2 \pi_1} \textbf{1}_{d-1} \textbf{1}_{d-1}^{\top} \end{bmatrix}} \\
    &\geq \lambda_{\max} \del{\begin{bmatrix}\frac{1}{l^2 \pi_1}+ \frac{1}{m^2} G_1 + \frac{(d-1)^2}{l^2 m^2 \pi_1}& \frac{d-m-1}{m^2 l^2 \pi_1}\textbf{1}_{d-1}^\top \\ \frac{d-m-1}{m^2 l^2 \pi_1}\textbf{1}_{d-1} & \frac{1}{m^2} \diag(C\textbf{1}_{d-1})+ \frac{d-1}{l^2 m^2 \pi_1} \textbf{1}_{d-1} \textbf{1}_{d-1}^{\top} \end{bmatrix}}.
\end{align*}

Now consider the following distribution $\pi^B$
\begin{align*}
    \pi^B(a_n) = \begin{cases}
        \pi(a_n) & \text{if $n \equiv 1 \pmod d$}\\
        \frac{1-\sum_{i: i \not\equiv 1 \pmod d} \pi_i}{D-d} & \text{otherwise}
    \end{cases}
\end{align*}

By AM-HM inequality, we have 
$$(d-1) G= \sum_{i=2}^d G_i = \sum_{i \not\equiv 1 \pmod d} \frac{1}{\pi_i} \geq \frac{(D-d)^2}{\sum_{i \not\equiv 1 \pmod d} \pi_i} = \frac{(D-d)^2}{1-\pi_1-{\sum_{i=2}^d \pi_i}} =\frac{d(d-1)}{\pi_2^B}$$
This means $G \geq \frac{d}{\pi_2^B}$, and naturally

\begin{align*}
    B(Q(\pi)) &\geq \lambda_{max} \del{\begin{bmatrix}\frac{1}{l^2 \pi_1}+ \frac{1}{m^2} G_1 + \frac{(d-1)^2}{l^2 m^2 \pi_1}& \frac{d-m-1}{m^2 l^2 \pi_1}\textbf{1}_{d-1}^\top \\ \frac{d-m-1}{m^2 l^2 \pi_1}\textbf{1}_{d-1} & \frac{1}{m^2} \diag(G\textbf{1}_{d-1})+ \frac{d-1}{l^2 m^2 \pi_1} \textbf{1}_{d-1} \textbf{1}_{d-1}^{\top} \end{bmatrix}}\\
    &\geq \lambda_{max} \del{\begin{bmatrix}\frac{1}{l^2 \pi_1}+ \frac{1}{m^2} G_1 + \frac{(d-1)^2}{l^2 m^2 \pi_1}& \frac{d-m-1}{m^2 l^2 \pi_1}\textbf{1}_{d-1}^\top \\ \frac{d-m-1}{m^2 l^2 \pi_1}\textbf{1}_{d-1} & \frac{1}{m^2} \diag(\frac{d}{\pi_2^B}\textbf{1}_{d-1})+ \frac{d-1}{l^2 m^2 \pi_1} \textbf{1}_{d-1} \textbf{1}_{d-1}^{\top} \end{bmatrix}}\\
    &= B(Q(\pi^B))
\end{align*}
By the minimality of $\pi$, $B(Q(\pi))=B(Q(\pi^B))$ and $\pi^B \in \Pi^B$. Therefore we can conclude that one of the optimal allocation $\pi$ should satisfy $\pi_i=\pi_j$ for all $i,j \not\equiv 1 \pmod d$.

% when $\{\pi_{di+1}\}_{i=0}^{d-1}$ are fixed,

\subsection{Proving that $\exists \pi^B \in \Pi^B$ such that $\pi_{d+1}^B = \pi_{2d+1}^B = \ldots = \pi_{(d-1)d+1}^B$ and $\pi_i^B = \pi_j^B$ for all $i, j\not\equiv 1 \pmod d $}

Suppose that $\pi\in \Pi^B$ which satisfies $\pi_i = \pi_j$ for all $i, j\not\equiv 1 \pmod d$). We aim to construct a $\pi^B$ such that in addition to this property, $\pi^B$ satisfies $\pi_{d+1}^B = \pi_{2d+1}^B = \ldots = \pi_{(d-1)d+1}^B$. 

%In this time, suppose that $\pi_1$ and $\pi_j (j \not\equiv 1 \pmod d$ are all fixed. 

Define $\pi^B$ as 
\begin{align*}
    \pi_i^B = \begin{cases}
        \frac{\sum_{j=2}^{d} \pi_j}{d-1} & \text{if $i=2, \cdots, d$}\\
        \pi_i & \text{Otherwise}
    \end{cases}.
\end{align*} Then, from AM-HM we can note that 

$$G_1 (\pi) = \sum_{i=2}^d \frac{1}{\pi_i} \geq (d-1)^2 \frac{1}{\sum_{i=2}^d {\pi_i}} = \sum_{i=2}^d \frac{1}{\pi_i^B}:=G_1 (\pi^B)$$

and therefore, $B(Q(\pi)) \geq B(Q(\pi^B))$ (we need to change only $G_1$ to $G_1(\pi^B)$ from the above calculation) and therefore $\pi^B \in \Pi^B$.

%and by the minimality we could notice that the optimal allocation should satisfy $\pi_2=\pi_3=\cdots = \pi_d$.
\subsubsection{Calculating \texorpdfstring{$\ldyad$}{}}

From the above observations, to calculate $B_{\min}(\cAhard)$, it suffices to restrict to those $\pi$'s of the following form: 
\begin{itemize}
    \item $\pi_1=a$
    \item $\pi_{d+1} = \pi_{2d+1} = \cdots = \pi_{(d-1)d+1} = b$
    \item $\pi_i = \cdots \pi_D = c$ for all $i \not\equiv 1 \pmod d$.
    \item $a+ (d-1)b + (D-d)c = 1$
    \item $G_2 = \cdots = G_d = G := \frac{d}{b}$, $G_1 = \frac{d-1}{c}$. 
\end{itemize}
To compute the maximum eigenvalue, we should solve the following characteristic equation:

\begin{align*}
&\det\del{\del{\begin{bmatrix}\frac{1}{l^2 a}+ \frac{1}{m^2} G_1 + \frac{(d-1)^2}{m^2 l^2 a}& \frac{d-m-1}{m^2 l^2 a}\textbf{1}_{d-1}^\top \\ \frac{d-m-1}{m^2 l^2 a}\textbf{1}_{d-1} & \frac{1}{m^2} \diag(G\textbf{1}_{d-1})+ \frac{d-1}{l^2 a m^2} \textbf{1}_{d-1} \textbf{1}_{d-1}^{\top} \end{bmatrix}}-\lambda I } = 0\\
\Leftrightarrow &\det\del{\del{\begin{bmatrix}\frac{1}{l^2 a}+ \frac{1}{m^2} G_1 + \frac{(d-1)^2}{m^2 l^2 a} - \lambda& \frac{d-m-1}{m^2 l^2 a}\textbf{1}_{d-1}^\top \\ \frac{d-m-1}{m^2 l^2 a}\textbf{1}_{d-1} & \diag((\frac{G}{m^2} -\lambda)\textbf{1}_{d-1})+ \frac{d-1}{l^2 a m^2} \textbf{1}_{d-1} \textbf{1}_{d-1}^{\top} \end{bmatrix}}} = 0\\
\Leftrightarrow &\det\del{\del{\begin{bmatrix}\frac{1}{l^2 a}+ \frac{1}{m^2} G_1  + \frac{(d-1)^2}{m^2 l^2 a}- \lambda - (\frac{d-m-1}{m^2l^2 a})^2 \frac{d-1}{{\frac{G}{m^2}-\lambda}+\frac{(d-1)^2}{l^2 a m^2}}& 0 \\ \frac{d-m-1}{m^2 l^2 a}\textbf{1}_{d-1} &  \diag((\frac{G}{m^2}-\lambda)\textbf{1}_{d-1})+ \frac{d-1}{l^2 a m^2} \textbf{1}_{d-1} \textbf{1}_{d-1}^{\top} \end{bmatrix}}} = 0 
\tag{Determinant is invariant under row operation}\\
\Leftrightarrow &(\frac{1}{l^2 a}+ \frac{1}{m^2} G_1  + \frac{(d-1)^2}{m^2 l^2 a}- \lambda -  \frac{(\frac{d-m-1}{m^2 l^2 a})^2\cdot (d-1)}{\frac{G}{m^2}-\lambda+\frac{(d-1)^2}{l^2 a m^2}}) \cdot \det\del{ \diag((\frac{G}{m^2}-\lambda)\textbf{1}_{d-1})+ \frac{d-1}{l^2 a m^2} \textbf{1}_{d-1} \textbf{1}_{d-1}^{\top}} = 0 \tag{Determinant cofactor formula}\\
\Leftrightarrow &(\frac{1}{l^2 a}+ \frac{1}{m^2} G_1  + \frac{(d-1)^2}{m^2 l^2 a}- \lambda - \frac{(\frac{d-m-1}{m^2 l^2 a})^2 \cdot (d-1)}{\frac{G}{m^2}-\lambda+\frac{(d-1)^2}{l^2 a m^2}}) \cdot ({\frac{G}{m^2}-\lambda+\frac{(d-1)^2}{l^2 a m^2}})\cdot (\frac{G}{m^2}-\lambda)^{d-2}= 0 \\
\Leftrightarrow &[(\frac{1}{l^2 a}+ \frac{1}{m^2} G_1  + \frac{(d-1)^2}{m^2 l^2 a}- \lambda)\cdot ({\frac{G}{m^2}-\lambda+\frac{(d-1)^2}{l^2 a m^2}}) - (d-1)(\frac{d-m-1}{m^2l^2 a})^2  ]\cdot ({\frac{G}{m^2}-\lambda})^{d-2}= 0 
\end{align*} 

From the above characteristic polynomial, we can notice there are $d-2$ repeated eigenvalues of size $G$, and the remaining two eigenvalues are the solution of the following quadratic equation:

$$[(\frac{1}{l^2 a}+ \frac{1}{m^2} G_1  + \frac{(d-1)^2}{m^2 l^2 a}- \lambda)\cdot ({\frac{G}{m^2}-\lambda+\frac{(d-1)^2}{l^2 a m^2}}) - (d-1)(\frac{d-m-1}{m^2l^2 a})^2  ]=0$$

After rearrangement, this formula looks like this:

$$ \lambda^2 - B\lambda + C =0$$

where $B= \frac{G_1}{m^2} + \frac{1}{l^2 a} + \frac{G}{m^2} + \frac{2(d-1)^2}{m^2 l^2 a}$ and $C = \del{\frac{1}{l^2 a}+ \frac{1}{m^2} G_1  + \frac{(d-1)^2}{m^2 l^2 a}}\cdot \del{\frac{G}{m^2}+\frac{(d-1)^2}{l^2 a m^2}} -  (d-1)(\frac{d-m-1}{m^2l^2 a})^2$. Now note that $C>0$, since $C> \del{\frac{(d-1)^2}{m^2 l^2 a}}^2 - (d-1)(\frac{d-m-1}{m^2l^2 a})^2>0$.

Since $C>0$ and $0<B^2 - 4C <B^2 $, $B \leq \frac{B+\sqrt{B^2 -4C}}{2}\leq 2B$ which means that the largest solution of the above quadratic equation is of order $B$. Now one could note that $B= \Theta (\max(\frac{G_1}{m^2}, \frac{G}{m^2}, \frac{d^2}{m^2 l^2 a}))$, or
$$B = \Theta (\max \del{\frac{d}{m^2 b}, \frac{d}{m^2 c}, \frac{d^2}{m^2 l^2 a}})$$

After optimizing the scale, $a = \Theta(\frac{db}{l^2}), c= \Theta(b)$ and from the constraint $a+ (d-1)b + (D-d)c=1$, 

$$\frac{1}{b} = \Theta{\del{\frac{d}{l^2} + D}}$$
and 
$B = \Theta \del{\frac{d^2}{m^2 l^2} +\frac{d^3}{m^2}}$ and so $B_{\min}(\cAhard) = \Theta \del{\frac{d^2}{m^2 l^2} +\frac{d^3}{m^2}}$. When applying $l=\frac{1}{\sqrt{d}}$ and $m=1$, we get $B_{\min}(\cAhard)=\Theta(d^3)$

{Recall that we have shown that $C_{\min}^{-1} (\cAhard)= \Theta \del{\frac{d^2}{m^2 l^2}}$; with this choice of $l$ and $m$, $C_{\min}^{-1} (\cAhard) = \Theta(d^3)$. Therefore, for $\cAhard$, $B_{\min}(\cAhard) = \Theta(C_{\min}^{-1}(\cAhard))$.}

\section{Examples of \texorpdfstring{$\ldyad$}{} and \texorpdfstring{$\cC_{\min}(\cA)$}{}}
\label{sec:b-min-c-min}

\subsection{\texorpdfstring{$\cA$}{} is Frobenius norm unit ball}

\begin{claim}
If $\cA$ is the unit ball in Frobenius norm: 
$\cA=\{A\in \RR^{d_1 \times d_2}: \|A\|_F \leq 1\}$, then $\cC_{\min}(\cA) = \frac{1}{d_1 d_2}$ and $B_{\min}(\cA) = d_2 d_2 d$. 
\end{claim}

\begin{proof}
%When $\cA$ is the unit ball in Frobenius norm: 
%$\cA=\{A\in \RR^{d_1 \times d_2}: \|A\|_F \leq 1\}$, 
We will prove $\cC_{\min}(\cA)=\frac{1}{d_1 d_2}$ by proving $\cC_{\min}(\cA)\leq \frac{1}{d_1 d_2}$ and $\cC_{\min}(\cA)\geq \frac{1}{d_1 d_2}$. 

\paragraph{Proving $\cC_{\min}(\cA)\geq \frac{1}{d_1 d_2}$: }Let $\cB=\{\reshape(e_{k}): k=1, \cdots, d_1 d_2\}$. 
%Since $\cB \subset\cA$ and by definition $\cC_{\min}(\cB)\leq \cC_{\min}(\cA)$. 
Note that $\Vec(\cB)$ is a $d_1 d_2$ dimensional canonical basis, and for any $\pi \in \Delta (\cB)$, $Q(\pi) = \sum_{i=1}^{d_1 d_2} \pi_i e_i e_i^\top = \diag(\pi_1, \cdots, \pi_{d_1 d_2})$ and $\lambda_{\min} (Q(\pi))=\min \{\pi_i\}_{i=1}^{d_1 d_2}$. Let $\pi$ be a uniform distribution over $\cB$. Then, $\lambda_{\min} (Q(\pi))=\frac{1}{d_1 d_2}$ and this fact leads to $\cC_{\min}(\cA)\geq \frac{1}{d_1 d_2}$.
%\geq \cC_{\min}(\cB)
%On the other hand, 
\paragraph{Proving $\cC_{\min}(\cA) \leq \frac{1}{d_1 d_2}$: }
Fix any distribution $\pi$ over $\cA$. 
%We know that nuclear norm is a convex function. 
Therefore, $\tr(\EE_{a\sim \pi} [\Vec(a) \Vec(a)^\top ]) =   \EE_{a\sim \pi} \tr(\|\Vec(a) \Vec(a)^\top) ]\leq 1$ since for all $a\in \cA$, $\|a\|_F \leq 1$ and $\tr(\Vec(a)\Vec(a)^\top) = \|\Vec(a)\|_2^2 = \|a\|_F^2 \leq 1$. Therefore, by the minimality of $\lambda_{d_1 d_2}$ we get $\lambda_{d_1 d_2} (Q(\pi))\leq \frac{1}{d_1 d_2} \tr(Q(\pi)) = \frac{1}{d_1 d_2}$.

\paragraph{Proving $\ldyad\geq d_1 d_2 d$: }

From the definition of $\blambda$ (Eq. \ref{eq: def of blambda}), 

    \begin{align*}
        \blambda &= \max \del{
        \lambda_{\max} \del{\sum_{i=1}^{d_2} D_i^{(\mathrm{col})}}, 
        \lambda_{\max}\del{\sum_{j=1}^{d_1} D_j^{(\mathrm{row})}} }\\
        &\geq \max \del{ \frac{1}{d_1}
        \textsf{tr} \del{\sum_{i=1}^{d_2} D_i^{(\mathrm{col})}}, 
        \frac{1}{d_2}\textsf{tr} \del{\sum_{j=1}^{d_1} D_j^{(\mathrm{row})}} } \tag{$\lambda_{\max}(M) \geq \frac{1}{d}\textsf{tr}(M)$ for any matrix $M \in \RR^{d \times d}$}\\
        &= \max \del{\frac{1}{d_1}
        \textsf{tr} \del{Q(\pi)^{-1}}, 
        \frac{1}{d_2}\textsf{tr} \del{Q(\pi)^{-1}}}\tag{From the definition of $D_i^{(\mathrm{col})}$ and $D_i^{(\mathrm{row})}$}\\
        &= \frac{1}{\min (d_1, d_2)} \textsf{tr} \del{Q(\pi)^{-1}}\\
        &\geq \frac{1}{\min (d_1, d_2)} \frac{(d_1 d_2)^2}{\textsf{tr} \del{Q(\pi)}} \tag{AM-HM inequality on the spectrum of $Q(\pi)^{-1}$}
    \end{align*}
Here, note that $\cA \subset \cB_{Frob}(1)$, which means 

\begin{align*}
    \textsf{tr} \del{Q(\pi)} &= \textsf{tr}(\EE_{a \sim \pi}[\ve(a)\ve(a)^\top])\\
    &= \EE_{a \sim \pi}[\textsf{tr}(\ve(a)\ve(a)^\top)] \tag{Linearity of expectation}\\
    &= \EE_{a \sim \pi}[||a||_F^2] \\
    &\leq \EE_{a \sim \pi}[1] \tag{$a \in \cA \subset \cB_{Frob}(1)$}\\
    &=1
\end{align*}

Therefore, $\blambda \geq \frac{(d_1 d_2)^2}{\min (d_1, d_2)} = d_1 d_2 d$ for any $\pi \in \cP(\cA)$

%On the other hand, 
\paragraph{Proving $\ldyad \leq d_1 d_2 d$: }

Consider 

\begin{align*}
    \pi(a) := \begin{cases}
        \frac{1}{d_1 d_2} & \text{ if $\ve(a) \in \{ e_i: i=1, \cdots d_1 d_2\}$}\\
        0 & \text{ Otherwise}
    \end{cases}
\end{align*}

(Recall that $e_i$ is a canonical basis where only $i$-th entry is 1 and all other entries are 0.) Obviously $\pi \in \cP (\cA)$. On the other hand, $Q(\pi) = \frac{1}{d_1 d_2} I_{d_1 d_2}$, which means $Q(\pi)^{-1} = d_1 d_2 I_{d_1 d_2}$ and $\blambda = d_1 d_2 d$. Therefore, $\ldyad \leq d_1 d_2 d$ by the minimality of $\ldyad$.
\end{proof}

\subsection{\texorpdfstring{$\cA$}{} is operator norm unit ball}\label{appsubsec: Bmin and Cmin in operator norm ball}

\begin{claim}
If $\cA$ is the unit ball in operator norm: 
$\cA=\{A\in \RR^{d_1 \times d_2}: \|A\|_{\op} \leq 1\}$, then $\cC_{\min}(\cA) = \Theta(\frac{1}{\max(d_1, d_2)})$ and $B_{\min}(\cA)=\max(d_1, d_2)^2$.
\end{claim}

\begin{proof}
We will prove that $\cC_{\min}(\cA)=\Theta(\frac{1}{\max(d_1,d_2)})$ by proving $\cC_{\min}(\cA)= O(\frac{1}{\max(d_1,d_2)})$ and $\cC_{\min}(\cA)= \Omega( \frac{1}{\max(d_1,d_2)})$. WLOG $d_2 \geq d_1$. 

\paragraph{Proving $\cC_{\min}(\cA)\geq \frac{1}{\max(d_1, d_2)}$:}
{Without loss of generality, assume that $d_2 \geq d_1$; we will show that $C_{\min}(\cA) \geq \frac{1}{d_2}$.}
Consider a distribution $\pi \in \Delta(\cA)$ which draws a matrix $A\in \cA$ by the following process:
\begin{itemize}
    \item Let $U \sim \sigma(d_1)$ and $V = [v_1; v_2; \ldots ;v_{d_2}] \sim \sigma(d_2)$ where $\sigma(d)$ denotes the Haar measure over $d \times d$ orthogonal matrices and $[v_1; v_2; \ldots ;v_{d_2}]$ is a concatenation of $d_2$ vectors.
    \item Let $\Sigma = \begin{bmatrix}
        I_{d_1} & 0_{d_1 \times (d_2 -d_1)}
    \end{bmatrix}$ where $I_d$ denotes $d$-dimensional identity matrix and $0_{a\times b}$ denotes $a\times b$ dimensional zero matrix. 
    \item Let $A = U \Sigma V^\top = U [v_1; \cdots; v_{d_1}]^\top$. Since $U$ and $V$ are all orthogonal matrices, we have $\|A\|_{\op}=1$. 
\end{itemize}

Note that $A$ has the same distribution as $[v_1; \cdots; v_{d_1}]^\top$. This is because $AA^\top = UU^\top=I_{d_1}$ so those rows are mutually orthonormal, and for any $v_j$ where $j>d_1$, $Av_j = U \Sigma [v_1; \cdots; v_{d_1}]^\top v_j = U\Sigma 0_{d_1\times 1} = 0_{d_1\times 1}$ which implies that all rows in $A$ and $v_{d_1+1}, \cdots, v_{d_2}$ forms an orthogonal basis. Therefore we can conclude
\[
\begin{bmatrix} 
v_1; \ldots ;v_{d_2}
\end{bmatrix}
\cdot
\begin{bmatrix} 
{U^\top} & 0 \\
0 & I
\end{bmatrix}
\overset{d}{=}
\begin{bmatrix} 
v_1; \ldots; v_{d_2}
\end{bmatrix}
\]
and $A \overset{d}{=} [v_1; \cdots; v_{d_1}]^\top$
. Now we should check the covariance matrix of $A$, $\EE\sbr{\Vec(A) \Vec(A)^\top}$. As mentioned in Appendix \ref{appsec:prelim of dcol and drow}, there exists a permutation matrix $P\in \mathbb{R}^{d_1 d_2 \times d_1 d_2}$ such that $P \Vec(A) = \Vec(A^\top)$ and $\EE\sbr{\Vec(A) \Vec(A)^\top}= P^\top \EE \sbr{\Vec(A^\top) \Vec(A^\top)^\top} P$. In our case it is easier to compute $\EE\sbr{\Vec(A^\top) \Vec(A^\top)^\top}$. Since $A \overset{d}{=}[v_1; \cdots;v_{d_1}]$,
\begin{align*}\EE\sbr{\Vec(A^\top) \Vec(A^\top)^\top}= \EE\begin{bmatrix}
    V_{1, 1} & \cdots &V_{1, d_1}\\
    \vdots & \ddots &\vdots\\
    V_{d_1, 1} & \cdots &V_{d_1, d_1}
\end{bmatrix}
\end{align*}
where $V_{ij}=v_i v_j^\top$. 
We can easily note that 
\[
\EE\sbr{V_{ij}} 
=
\begin{cases}
0_{d_2 \times d_2} & i\neq j\\
\frac{1}{d_2} I_{d_2} & i=j
\end{cases}
\]
and therefore $\EE\sbr{ \ve(A^\top) \ve(A^\top)^\top } = \frac{1}{d_2} I_{d_1d_2 \times d_1d_2}.$ As a result,
\[
\EE\sbr{ \ve(A) \ve(A)^\top } = P^\top \del{\frac{1}{d_2} I_{d_1d_2 \times d_1d_2}}P = \frac{1}{d_2} P^\top P = \frac{1}{d_2} I_{d_1d_2 \times d_1d_2}.
\]
This implies that $\cC_{\min}( \cA ) \geq \frac{1}{\max(d_1, d_2)}$. 

%On the other hand, w
\paragraph{Proving $\cC_{\min}(\cA) \leq O(\frac{1}{\max(d_1,d_2)})$: }
We know that nuclear norm is a convex function. Therefore, $\|\EE_{a\sim \pi} [\Vec(a) \Vec(a)^\top ]\|_* \leq  \EE_{a\sim \pi} [\|\Vec(a) \Vec(a)^\top\|_* ]\leq d_1 +d_2$ since for all $a\in \cA$, $\|a\|_{\op} \leq 1$ means $\|a\|_F \leq \sqrt{\min(d_1, d_2)}$, and $\|\Vec(a)\Vec(a)^\top\|_* = \|\Vec(a)\Vec(a)^\top\|_{\op}=\|\Vec(a)\|^2 = \|a\|_F^2 \leq \min(d_1 , d_2)$. Therefore, by the minimality of $\lambda_{d_1 d_2}$ we get $\lambda_{d_1 d_2} (Q(\pi))\leq \frac{1}{d_1 d_2} \|Q(\pi)\|_* = \frac{1}{\max(d_1, d_2)}$.

\paragraph{Proving $\ldyad\geq \max(d_1, d_2)^2$: }

From the definition of $\blambda$ (Eq. \ref{eq: def of blambda}), 

    \begin{align*}
        \blambda &= \max \del{
        \lambda_{\max} \del{\sum_{i=1}^{d_2} D_i^{(\mathrm{col})}}, 
        \lambda_{\max}\del{\sum_{j=1}^{d_1} D_j^{(\mathrm{row})}} }\\
        &\geq \max \frac{1}{d_1}\del{
        \textsf{tr} \del{\sum_{i=1}^{d_2} D_i^{(\mathrm{col})}}, 
        \frac{1}{d_2}\textsf{tr} \del{\sum_{j=1}^{d_1} D_j^{(\mathrm{row})}} } \tag{$\lambda_{\max}(M) \geq \frac{1}{d}\textsf{tr}(M)$ for any matrix $M \in \RR^{d \times d}$}\\
        &= \max \del{\frac{1}{d_1}
        \textsf{tr} \del{Q(\pi)^{-1}}, 
        \frac{1}{d_2}\textsf{tr} \del{Q(\pi)^{-1}}}\tag{From the definition of $D_i^{(\mathrm{col})}$ and $D_i^{(\mathrm{row})}$}\\
        &= \frac{1}{\min (d_1, d_2)} \textsf{tr} \del{Q(\pi)^{-1}}\\
        &\geq \frac{1}{\min (d_1, d_2)} \frac{(d_1 d_2)^2}{\textsf{tr} \del{Q(\pi)}} \tag{AM-HM inequality on the spectrum of $Q(\pi)^{-1}$}
    \end{align*}
Here, note that $\cA = \cB_{\op}(1) \subset \cB_{Frob}(\sqrt{\min(d_1,  d_2)})$.    
Then, 

\begin{align*}
    \textsf{tr} \del{Q(\pi)} &= \textsf{tr}(\EE_{a \sim \pi}[\ve(a)\ve(a)^\top])\\
    &= \EE_{a \sim \pi}[\textsf{tr}(\ve(a)\ve(a)^\top)] \tag{Linearity of expectation}\\
    &= \EE_{a \sim \pi}[||a||_F^2] \\
    &\leq \EE_{a \sim \pi}[\min(d_1, d_2)] \tag{$a \in \cA \subset \cB_{Frob}(\sqrt{\min(d_1,  d_2)})$
    }\\
    &=\min(d_1 ,d_2)
\end{align*}

Therefore, $\blambda \geq \frac{(d_1 d_2)^2}{\min(d_1, d_2)^2} = \max(d_1 , d_2)^2$ for any $\pi \in \cP(\cA)$

%On the other hand, 
\paragraph{Proving $\ldyad \leq \max (d_1,  d_2)^2$: }

From Lemma \ref{lem: comparison between ldyad and cmin}, $\ldyad \leq \frac{\max(d_1, d_2)}{C_{\min}(\cA)} \leq \max(d_1, d_2)^2$. 
% \ks{When $4|d_1 d_2$, using Hadamard matrix we can create diagonal matrices which are in $\cB_{\op}(1)$, and orthogonal to each other w.r.t the matrix inner product. Naturally, we can follow the same logic from Frobenius norm ball above. However, we don't have divisibility guarantee. }
% \chicheng{I think $d^2$ is the best bound we can hope for? (if my comment above is true)}

\end{proof}

This computation result leads to the following:

\begin{corollary}\label{appcor: Cmin is smaller than 1d}
For any $\cA \subset B_{\op}(1)$, $C_{\min}(\cA)\leq \frac{1}{d}$.     
\end{corollary}
\begin{proof}
    By the maximality of the $C_{\min}$, when a set $S$ is a subset of $S'$, then $C_{\min}(S) \leq C_{\min}(S')$. We proved in this subsection that $C_{\min}(B_{\op}(1))=\frac{1}{d}$. Therefore the corollary follows. 
\end{proof}
\section{Proof of Theorem \ref{thm:etc regret bound}}

\begin{proof}
First, if $T \leq \frac{\sigma^2 r^2 B_{\min}(\cA)}{\rmax^2}$, 
we have
$T \rmax \leq (\sigma^2 \rmax r^2 \ldyad T^2)^{1/3})$, therefore 
\[
\Reg(T) \leq T \rmax
\leq \tilde{O}((\sigma^2 \rmax r^2 \ldyad T^2)^{1/3})
\]
trivially holds. 

Therefore, throughout the reset of the proof we focus on the case when $T \geq \frac{\sigma^2 r^2 B_{\min}(\cA)}{\rmax^2}$. In this case, $n_0 = \del{\frac{ 
 \sigma^2 r^2 \ldyad T^2}{\rmax^2}}^{1/3} \leq T$, and by our assumption that $T \geq {r^2} \ldyad(\frac{\sigma + \rmax}{\sigma})^4$, we have $n_0 \geq {r^2}\ldyad(\frac{\sigma + R_{\max}}{\sigma})^2$.
% \chicheng{Is my change correct? This may also affect the main text's theorem statement?} \ks{Oh sorry. I think you are right. We don't need to change the main regret bound, but maybe we should change the lower bound for $T$. }
This range of $n_0$ satisfies the condition of Theorem \ref{thm:lpopart}, which gives the following recovery bound on $\hat{\Theta}$ with probability $1-\delta$:
    \[
    \| \hat{\Theta} - \Theta^* \|_{*}
    \leq 
    2r \| \hat{\Theta} - \Theta^* \|_{\op} 
    \leq 
    2r \sigma
    \sqrt{\frac{\del{ \ldyad\ln \frac{2(d_1 +d_2)}{\delta} } }{n_0}}
    \]
    
    For the rest of the rounds, we can bound the instantaneous regret of the exploitation as follows:
    \begin{align*}
        \inner{\Theta^*}{A^* - A_t} &=\inner{\Theta^* - \hat{\Theta}}{A^*}+\inner{\hat\Theta}{A^*}-\inner{\Theta^*}{A_t}\\
        &\leq \inner{\Theta^* - \hat{\Theta}}{A^*} + \inner{\hat\Theta-\Theta^*}{A_t} \tag{Definition of $A_t$}\\
        &\leq \|\Theta^* - \hat{\Theta}\|_* (\|A^*\|_{\op} + \|A_t\|_{\op}) \tag{Holder's inequality}\\
        &\leq 2 \sigma r \sqrt{\del{ 2\frac{\ldyad}{n_0}\ln \frac{2(d_1 +d_2)}{\delta} } }\times 2
    \end{align*}
    Therefore, we can conclude the upper bound of the total regret bound as follows:
    \begin{align*}
        \Reg(T) &= \sum_{t=1}^T \inner{\Theta^*}{A^* - A_t}\\
        &\leq n_0 \rmax + T \cdot 8 \sigma r \sqrt{\frac{\del{ \ldyad\ln \frac{2(d_1 +d_2)}{\delta} } }{n_0}}
    \end{align*}
 The final regret bound of Theorem \ref{thm:etc regret bound} follows by plugging in the setting of $n_0 = \del{\frac{ 
 \sigma^2 r^2 \ldyad T^2}{\rmax^2 }}^{1/3}$.
\end{proof}

% \chicheng{The assumption on $n$ satisfies that $n_0 \geq \frac{d}{C_{\min}}(\frac{\sigma + R_{\max}}{\sigma})^2$}\ks{We need to explicitly state $n0$ really satisfies the condition.})

\section{Results of \cite{jun19bilinear}}

\section{Proof of Theorem \ref{thm:estr bound}} 
\label{appsec: proof of estr bound}

\subsection{LowOFUL Algorithm}\label{appsec:lowoful}

Before we proceed, we need to state the ESTR algorithm \cite{jun19bilinear} for completeness. Throughout this section, we will use the notations in this algorithm, such as the confidence set $C_t$.

\begin{algorithm}[h]
\caption{LowOFUL \cite{jun19bilinear}}
\label{alg:lowoful}
\begin{algorithmic}[1]
\STATE {\bfseries Input:} time horizon $T'$, arm set $\cA_{vec}'$, lower dimension $k$, regularization parameter $\lambda_1$, failure rate $\delta$, positive constants $B, B_\perp$, $\lambda$, $\lambda_\perp$
%pilot estimator $\Theta_0$, 
\STATE Set $\Lambda= \diag(\lambda,\cdots, \lambda, \lambda_{\perp}, \cdots, \lambda_{\perp})$ where $\lambda$ occupies the first $k$ diagonal entries, and set $V_0=\Lambda$, $\theta_0 = \vec{(0_{d_1 \times d_2})}$.
\FOR{$t = 1,\cdots, T'$}
\STATE $\sqrt{\beta_t} = \sigma 
 \sqrt{\log \frac{|V_{t-1}|}{|\Lambda|\delta^2}} + \sqrt{\lambda} {B} + \sqrt{\lambda_{\perp}} {B_\perp}$
\STATE $C_t = \{\theta: \|\theta-\hat{\theta}_{t-1}\|_{V_{t-1}} \leq \sqrt{\beta_t} \}$
\STATE Compute $a_t = \arg \max_{a\in\cA_{vec}'} \max_{\theta \in C_t} \inner{a}{\theta}$
\STATE Pull arm $a_t$ and receive reward $y_t$.
\STATE Update $V_t = V_{t-1}+a_t a_t^\top$, $A = [a_1;\cdots ;a_t]$, $\textbf{y}=[y_1, \cdots, y_t]$ $\theta_t = V_t^{-1} A \textbf{y}$ 
\ENDFOR
\end{algorithmic}
\end{algorithm}

\subsection{Proof of Theorem \ref{thm:estr bound}}

\begin{proof}
Let's divide the regret of Algorithm \ref{alg:estr-lowrank-dr} into two terms. Let $R_1$ be the regret occured by the procedure before calling Algorithm \ref{alg:lowoful}, and let $R_2$ be the regret occured by invoking LowOFUL. 

\paragraph{Part 1: Bounding $R_1$.} For $R_1$, since each instantaneous regret is bounded as follows: 
$$\inner{\Theta^*}{A^*-A_t}\leq \|\Theta^*\|_{*} \|A^* -A_t\|_{\op}\leq \|\Theta^*\|_{*} (\|A^*\|_{\op} +\|A_t\|_{\op})\leq 2 \|\Theta^*\|_{*}$$ 

Therefore, we can trivially bound $R_1 \leq n_0 \| \Theta^* \|_* \leq n_0 S_*$. 

\paragraph{Part 2: bounding subspace estimation error.} From the analysis on Section \ref{sec:new analysis}, we have $\| \hat{\Theta} - \Theta^* \|_{\op} \leq \sqrt{\frac{\ldyad \sigma^2}{n_0}}$. Here, we will use the following operator norm version of Wedin's Theorem \citep[Theorem 4.4]{stewart1990matrix}; \edit{}{for the purpose of our analysis,} this is \edit{sometimes}{} tighter than the Frobenius norm version of Wedin's Theorem \citep[Theorem 4.1]{stewart1990matrix}.

\begin{theorem}[Wedin Theorem]
Let $M$ and $M^*$ be two $d_1 \times d_2$ matrices with the following SVD:
\begin{align*}
    M = \begin{bmatrix}
    U_1 & U_\perp
\end{bmatrix}\begin{bmatrix}
    \Sigma_1 & 0_{r\times (d_2-r)}\\
    0_{(d_1-r) \times r} & \Sigma_2
\end{bmatrix}\begin{bmatrix}
    V_1^\top \\ V_\perp^\top
\end{bmatrix}\\
    M^* = \begin{bmatrix}
    U^*_1 & U^*_\perp
\end{bmatrix}\begin{bmatrix}
    \tilde{\Sigma}_1 & 0_{r\times (d_2-r)}\\
    0_{(d_1-r) \times r} & \tilde{\Sigma}_2
\end{bmatrix}\begin{bmatrix}
    (V^*_1)^\top \\ (V^*_\perp)^\top
\end{bmatrix}
\end{align*}
Where \begin{itemize}
    \item $\Sigma_1, \tilde{\Sigma}_1$ represents top-r singular values for $M$ and $M^*$
    \item $\Sigma_2, \tilde{\Sigma}_2$ represents the rest of singular values for $M$ and $M^*$
    \item $(U_1, V_1)$, $(U^*_1, V^*_1)$ \edit{be}{are} the corresponding singular vectors for $\Sigma_1$ and $\tilde{\Sigma}_1$
    \item $(U_\perp, V_\perp)$, $(U^*_\perp, V^*_\perp)$ \edit{be}{are} the corresponding singular vectors for $\Sigma_2$ and $\tilde{\Sigma}_2$
\end{itemize}
respectively. Suppose that there are numbers $\alpha, \delta>0$ such that
$$ \lambda_r (\Sigma_1)\geq \alpha+\delta, \text{       and       } \lambda_{\max} (\tilde{\Sigma}_2) \leq \alpha$$
Then, $$ \max\{\|U_\perp^\top U_1^*\|_{\op}^2 , \|V_\perp^\top V_1^*\|_{\op}^2 \}\leq \frac{\max\{\|(U^*_1)^\top (M-M^*)\|_{\op}^2 , \|(M-M^*)V^*_1\|_{\op}^2\}}{\delta^2}$$
\end{theorem}

We can check that by the assumption that $T \geq \frac{16\ldyad \sigma^4}{d^{0.5}\minsig^2}$, $n_0 \geq {\frac{4B_{\min}(\cA) \sigma^2}{\minsig^2}}$. Thus by Weyl's Theorem, 
$\lambda_r(\hat{\Theta}) \geq \lambda_r(\Theta^*) - \sqrt{\frac{B_{\min}(\cA) \sigma^2}{n_0}} \geq \frac{\minsig}{2}$, therefore, choosing
$\delta= \frac{\minsig}{2},\alpha=0$ satisfies the condition, since the rank of $\hat{\Theta}$ is $r$ and therefore $\lambda_j (\Sigma_2)=0$ {for all $j=r+1, \ldots, \min(d_1, d_2)$}. 

Now, substitute parameters as follows: suppose the SVD of $\Theta^* = U^* \Sigma^* V^*$
\begin{align*}
    & M= \hat{\Theta}, M^* = \Theta^*\\
    & U_1= U_1, U^* = U^*\\
    & V_1= V_1, V^* = V^*\\
    & \Sigma_1= \tilde{\Sigma}_1, {\Sigma_1^*} = \Sigma_1^*\\
\end{align*}
Plus, note that 
\begin{align*}
    \|\Theta_1-\Theta^*\|_{\op}^2 =\|\Theta_1-\Theta^*\|_{\op}^2 \|V_1^*\|_{\op}^2\geq \|(\Theta_1-\Theta^*)V_1^*\|_{\op}^2
\end{align*}
and similarly, $\|\Theta_1 -\Theta^*\|_F^2 \geq \|U_1^*(\Theta_1-\Theta^*)\|_F^2$
. Now Wedin's theorem implies that 
$$ {\max(\|U_\perp^\top U_1^*\|_{\op}^2 , \|V_\perp^\top V_1^*\|_{\op}^2)}\leq \frac{\|\Theta-\Theta^*\|_{\op}^2}{\delta^2}$$

With the result of Theorem \ref{thm:wlpopart}, we can conclude
$$ \| \hat{U}_\perp^\top U^* \|_{\op} \leq \frac{1}{\minsig} \sqrt{\frac{\ldyad \sigma^2}{n_0 }}, \| \hat{V}_\perp^\top V^* \|_{\op} \leq \frac{1}{\minsig} \sqrt{\frac{\ldyad \sigma^2}{n_0 }}$$
 Therefore, $\| \hat{U}_\perp^\top \Theta \hat{V}_\perp \|_{F} 
\leq
\| \hat{U}_\perp^\top U\|_{\op} \cdot \| \Sigma \|_F \cdot \| V^\top \hat{V}_\perp \|_{\op} 
\leq 
\frac{\ldyad \sigma^2 \| \Theta^* \|_F }{n_0 \minsig^2} \leq \frac{\ldyad\sigma^2 S_* }{n_0 \minsig^2} =: B_\perp$

\paragraph{Part 3: bounding $R_2$.}
Recall that we set $\lambda_{\perp} = \frac{T}{r}, B_\perp=\frac{\sigma^2 \ldyad S_*}{n_0 \minsig^2}$ in low-OFUL.

Let $reg_t$ be the instantaneous pseudo-regret at time step $t$: $reg_t = \inner{\Theta^*}{A^*-A_t}
= 
\inner{\vec(\Theta^*)}{ \vec(A^*) - \vec(A_t) }
$ where $A^*=\arg \max_{A \in \cA} \inner{\Theta^*}{A}$. From the fact that $\Theta^*\in C_t$ \citep[Lemma 1]{jun19bilinear}
and using Cauchy-Schwarz inequality, we have
\begin{align}
    reg_t &= \inner{\vec(\Theta^*)}{ \vec(A^*) - \vec(A_t) } \nonumber\\
    &\leq \max_{\Theta\in C_{t-1}} \inner{\vec({\Theta})-\vec(\Theta^*)}{\vec(A_t) } \tag{Definition of $A_t$}\\
    &\leq \max_{\Theta \in C_{t-1}} \|\vec({\Theta})-\vec(\Theta^*)\|_{V_{t-1}} \|\vec(A_t)\|_{V_{t-1}^{-1}} \\
    &\leq 2\sqrt{\beta_t} \|\vec(A_t)\|_{V_{t-1}^{-1}} \tag{Definition of $C_t$}\\
    &\leq \sqrt{\beta_T} \|\vec(A_t)\|_{V_{t-1}^{-1}} \label{appeq:instantaneous regret upper bound}
\end{align}

Now, define $H_T := \{t\in[T]: t> n_0, \|A_t\|_{V_{t-1}^{-1}}>1 \}$ and $\bar{H}_T:= \{t\in[T]: t> n_0, \|A_t\|_{V_{t-1}^{-1}}\leq 1 \}$. Then,

\begin{align}
    R_2 &= \sum_{t=n_0+1}^T reg_t \nonumber \\
    &= \sum_{t=n_0+1}^T reg_t \mathbbm{1} \{t \in \bar{H}_T\}+\sum_{t=n_0+1}^T reg_t \mathbbm{1} \{t\in H_T\}\nonumber \\
    &= \sum_{t=n_0+1}^T reg_t \mathbbm{1} \{t \in \bar{H}_T\}+2S_* |H_T| \tag{$reg_t\leq 2S_*$}\nonumber\\
    &\leq \sqrt{|\bar{H}_T|\sum_{t \in \bar{H}_T} reg_t^2} + 2S_*|H_T|\tag{Cauchy-Schwarz}\\
    &\leq \sqrt{|\bar{H}_T|\beta_T \sum_{t \in \bar{H}_T} \|\vec(A_t)\|_{V_{t-1}^{-1}}^2} + 2S_*|H_T|\tag{Eq. \eqref{appeq:instantaneous regret upper bound}}\\
    &\leq \sqrt{|\bar{H}_T|\beta_T \sum_{t=n_0+1}^T \min(1,\|\vec(A_t)\|_{V_{t-1}^{-1}}^2)} + 2S_*|H_T| 
\label{appeq: estr regret formula pre}
\end{align}
Now for the first term of Eq. \eqref{appeq: estr regret formula pre}, we can use the elliptic potential lemma \cite{ay11improved, lattimore20bandit}:

\begin{lemma}[\cite{lattimore20bandit}, Lemma 19.4]
    $\sum_{t=1}^n \min(1,\|\vec(A_t)\|_{V_{t-1}^{-1}}) \leq 2\log \frac{|V_T|}{|\Lambda|}$
\end{lemma}

For the second term, $S_* |H_T|$, we can use the slight modification of the elliptical potential count lemma in \cite{gales2022norm}:
\begin{lemma}[Modification of Lemma 7, \cite{gales2022norm}]
    $|H_T|\leq \frac{2d_1 d_2}{\log 2} \max\del{1, \log \del{\frac{\omega_1}{\omega_2}+\frac{\min(d_1, d_2)}{\omega_2 \log 2 }}}$
\end{lemma}

\begin{proof}
    Let $M_T = \Lambda + \sum_{t \in H_T} \vec (A_t) \vec (A_t)^\top$. Then,
    \begin{align*}
        \det(M_T)&\leq \del{\frac{1}{d_1 d_2} \tr(M_T)}^{d_1 d_2}\\
        &= \del{\frac{\tr (\Lambda) + \tr(\sum_{t \in H_T} \vec (A_t) \vec (A_t)^\top)}{d_1 d_2} }^{d_1 d_2}\\
        &\leq \del{\frac{\tr(\Lambda) + \min(d_1, d_2) |H_T|)}{d_1 d_2} }^{d_1 d_2}\tag{$\|\vec (A_t)\|_2 \leq \sqrt{\min(d_1 , d_2)}$}
    \end{align*}
    Also, using the trick in the proof of \citep[Lemma 11]{ay11improved}, one can also achieve a lower bound of $\det (M_T)$
    \begin{align*}
        \det (M_T) &= \det (\Lambda) \cdot \prod_{t \in H_T} (1+ \|\vec (A_t)\|_{M_{t-1}^{-1}})\\
        &\geq \det (\Lambda) \cdot \prod_{t \in H_T} (1+ \|\vec (A_t)\|_{V_{t-1}^{-1}}) \tag{$V_{t-1}^{-1} \succeq M_{t-1}$}\\
        &\geq \det (\Lambda) 2^{|H_T|} \tag{Definition of $H_T$} 
     \end{align*}
    Therefore, we have $\det (\Lambda) 2^{|H_T|} \leq \det(M_T) \leq \del{\frac{k \lambda + (d_1 d_2 -k) \lambda_{\perp} + \min(d_1, d_2)  |H_T|)}{d_1 d_2}}^{d_1 d_2}$, or after taking log on both sides we have
     $$ |H_T|\log 2 + \log \det(\Lambda) \leq d_1 d_2 \log \frac{\tr (\Lambda) + \min(d_1, d_2)  |H_T|}{d_1 d_2} $$
     Now let $\omega_1= \max(\lambda, \lambda_{\perp})$ and $\omega_2 = \min(\lambda, \lambda_{\perp})$. Then $\log \det (\Lambda) \geq d_1 d_2 \omega_2$ and $\tr(\Lambda)\leq d_1 d_2 \omega_1$ which leads
     $$ |H_T|\leq \frac{d_1 d_2}{\log 2} \log \del{\frac{\omega_1}{\omega_2} + \frac{|H_T|}{d \omega_2} }$$
     Using Lemma \ref{applem: lemma 8 of norm agnostic} with $\eta=\frac{1}{2}$, $A= \frac{d_1 d_2}{\log 2}, B=\frac{1}{d\omega_2}$, $C=\frac{\omega_1}{\omega_2}$ and $X=|H_T|$ leads
     $$ |H_T|\leq \frac{2d_1 d_2}{\log 2} \log \del{\frac{d_1 d_2}{\log 2} \del{\frac{\omega_1}{\omega_2 |H_T|}+\frac{1}{d\omega_2}}}.$$
     Now suppose $|H_T|>\frac{2d_1 d_2}{\log 2}$. Then, from above inequality we have
     $$ |H_T|\leq \frac{2d_1 d_2}{\log 2} \log \del{\frac{\omega_1}{\omega_2}+\frac{\min(d_1, d_2)}{\omega_2 \log 2 }}.$$ Therefore, $$|H_T|\leq \frac{2d_1 d_2}{\log 2} \max\del{1, \log \del{\frac{\omega_1}{\omega_2}+\frac{\min(d_1, d_2)}{\omega_2 \log 2 }}}$$
\end{proof}

\begin{lemma}[Modification of Lemma 8, \citep{gales2022norm}]\label{applem: lemma 8 of norm agnostic}
    Let $X,A,B,C\geq 0$. Then $X\geq A \log (C+BX)$ implies that for all $\eta\in(0,1)$,
    $$ X \leq \frac{A}{1-\eta} \log \del{\frac{A}{2\eta} \del{\frac{C}{X}+B}}$$
\end{lemma}
\begin{proof}
    Simply change $1+BX$ to $C+BX$ and following the proof in \citet{gales2022norm} leads the desired result.
\end{proof}
For the reasonable case we have $T\gg d_1 d_2\gg \Theta(1)$ and therefore we can safely say $|H_T|\leq O(d_1 d_2\log T)$. Overall, we have 
\begin{align}\label{eq: estr regret formula}
R_2\leq 4 \sqrt{\beta_T}\sqrt{\log \frac{|V_T|}{|\Lambda|}}\sqrt{T} + O(d_1 d_2 S_* \log (T))
\end{align}
% \begin{align}\label{eq: estr regret formula}
%     R_2 &= \sum_{t=n_0+1}^T reg_t \leq 2 \sqrt{\max\{2, \frac{1}{\lambda}\}}\sqrt{\beta_T}\sqrt{\log \frac{|V_T|}{|\Lambda|}}\sqrt{T}
% \end{align}
where $\sqrt{\beta_t} = B\sqrt{\lambda} + B_\perp \sqrt{\lambda_\perp} + \sigma \sqrt{\log \frac{|V_t|}{|\Lambda|}}$.

%\chicheng{Do we nail down the choice of $\lambda$?}

Now the minor difference from \cite{jun19bilinear, lu2021low} comes from the computation of $\log \frac{|V_T|}{|\Lambda|}$, simply because we have different bounds on the $l_2$ norm of the actions (note that 
%for all $a \in \cA$, $\| a \|_{F} \leq \sqrt{d} \| a\|_{\op} \leq \sqrt{d}$). 
for all $a \in \edit{\cA'}{\cA_{vec}'}$, $\| a \|_2 = \| \reshape(a) \|_F \leq \sqrt{d} \| \reshape(a) \|_{\op} \leq \sqrt{d}$.)

\begin{lemma}[Modification of \citet{valko14spectral}, Lemma 5]\label{lem:valko logdet}
    For any $T$, let $\Lambda = \diag([\lambda_1, \cdots , \lambda_p])$. Then,
    $$ \log \frac{|V_T|}{|\Lambda|} \leq \max \{\sum_{i=1}^p \log (1+\frac{d t_i}{\lambda_i})\} $$
where the maximum is taken over all possible positive real numbers $t_1, \cdots , t_p$ such that $\sum_{i=1}^p t_i= T$. 
\end{lemma}

Note that in comparison with~\cite{valko14spectral} (which originally assumes $\|a_t\|_2 \leq 1$ for all $t$), we added a factor of $d$ inside the $\log$ because $V_T = \sum_{t=1}^T a_t a_t^\top$ and each $\| a_t \|_2 \leq \sqrt{d}$. Detailed proof is in Appendix \ref{appsubsec:lemma proof valko}
%\chicheng{explain a bit on where the factor of $d$ (inside log) comes from? I think it is from the fact that $V_T = \sum_{t=1}^T a_t a_t^\top$ and each $\| a_t \|_2 \leq \sqrt{d}$? (Also, we should probably mention that this $t$ here is not the global time index for the contextual bandit problem..)}\ks{TODO: When we need to prepare camera-ready version, add an explicit proof for this lemma. }

The only difference from the original lemma is that our Frobenius norm of $\|a\|_F$ is bounded by $\sqrt{d}$, so we need to compensate that scale difference inside the logarithm. Using our $\Lambda= \diag(\lambda,\cdots, \lambda, \lambda_{\perp}, \cdots, \lambda_{\perp})$ with Lemma \ref{lem:valko logdet} we can induce the following result:

\begin{lemma}[Modification of \citet{jun19bilinear}, Lemma 3]\label{lem:jun logdet}
    If $\lambda_{\perp}=\frac{T}{r \log (1+ \frac{dT}{\lambda})}$, then
$$\log \frac{|V_T|}{|\Lambda|} \leq 2k \log (1+\frac{dT}{\lambda})$$
\end{lemma}
\begin{proof}
\begin{align*}
    \log \frac{|V_T|}{|\Lambda|} &\leq \max \{\sum_{i=1}^p \log (1+\frac{d t_i}{\lambda_i})\} \\
    &\leq k \log (1+ \frac{dT}{\lambda}) + \sum_{i=k+1}^p \log (1+\frac{d t_i}{\lambda_\perp})\\
    \sum_{i=k+1}^p \log (1+\frac{d t_i}{\lambda_i}) &\leq \sum_{i=k+1}^p (\frac{d t_i}{\lambda_\perp}) \leq \frac{dT}{\lambda_\perp}\leq k \log (1+\frac{dT}{\lambda})
\end{align*}
\end{proof}
One can note that the additional $d$ factor from Lemma \ref{lem:valko logdet} leads $\lambda_\perp$ should have order $\frac{T}{r}$, not like $\frac{T}{k}$ in \citet{jun19bilinear}. 

%; this ensures that $\ln \frac{\det(V_T(\Lambda))}{\det(V_0(\Lambda))} \leq O(rd)$.)

Combining Lemma \ref{lem:jun logdet} with Eq. \eqref{eq: estr regret formula}, regret occured by the LowOFUL algorithm is
\begin{align*}
    R_2 &\leq \tilde{O}((\sigma k\sqrt{T} + B\sqrt{k\lambda T} + B_\perp \sqrt{k\lambda_{\perp}T}))\\
    &\leq \tilde{O}(\sigma r d \sqrt{T} + T\sqrt{d} B_{\perp} ) \\
    &\leq \tilde{O}(\sigma r d \sqrt{T} + T\frac{\sigma^2 d^{0.5} \ldyad S_*}{n_0 \minsig^2} )
\end{align*}

\paragraph{Part 4: putting it together.} Therefore, the total regret of ESTR can be bounded by 
\begin{align*}
\Reg_T = R_1 + R_2 
\leq & 
\tilde{O}\del{ n_0 S_*
+
\sigma r d \sqrt{T}
+ 
\frac{T d^{0.5} \ldyad\sigma^2 S_* }{\edit{n_2}{n_0} \minsig^2}}
\\
\leq & 
\tilde{O}\del{\sigma r d \sqrt{T}
+ 
\sigma \sqrt{ S_*^2  \frac{d^{0.5} \ldyad}{ {\minsig^2}} T }}
\end{align*}
with the setting of $n_0 = \sqrt{\frac{d^{0.5}\ldyad}{S_r^2}T}$ in the algorithm.
\end{proof}

\subsection{Proof of Lemmas we have used in this section}
\subsubsection{Proof of Lemma \ref{lem:valko logdet}} \label{appsubsec:lemma proof valko}

\begin{proof} 
We need the following lemma of \citet{valko14spectral}:
\begin{lemma}[Modification of \citet{valko14spectral}, Lemma 4]\label{lem:valko eigenvec}
Let $\Lambda = \diag(\lambda_1, . . . , \lambda_p )$ be any diagonal matrix with strictly positive entries. Then for any vectors $(a_t)_{1\leq t\leq T}$ such that $\|a_t\|_2 \leq C$ for some constant $C$ for all $1 \leq t \leq T$, we
have that the determinant $|V_T|$ is maximized when all $a_t$ are aligned with the axes.
\end{lemma}
The proof of Lemma \ref{lem:valko eigenvec} is exactly the same as \citet{valko14spectral}, Lemma 4. Now, in our case, for each $1\leq t\leq T$, $x_t = \Vec{X_t}$ and $\|x_t\|_2 \leq \|{X_t}\|_F \leq \sqrt{d}\|X_t\|_{op}\leq \sqrt{d}$. Now,
\begin{align*}
    |V_T| &= |\Lambda + \sum_{t=1}^T x_t x_t^\top|\\
    &\leq \max_{(a_i)_{i=1}^t: \|a_i\|_2 \leq \sqrt{d}} |\Lambda + \sum_{t=1}^T a_t a_t^\top|\\
    &= \max_{(a_i)_{i=1}^t: a_i \in \{\sqrt{d}e_1, \cdots, \sqrt{d}e_p\}} |\Lambda + \sum_{t=1}^T a_t a_t^\top| \tag{Lemma \ref{lem:valko eigenvec}}\\
    &\leq \max_{(t_i)_{i=1}^t: t_i\geq 0, \sum_{i=1}^t t_i = T} (\lambda_i + d t_i)
\end{align*}
and dividing $|V_T|$ by $|\Lambda|$ and taking logarithm leads the result of Lemma \ref{lem:valko logdet}.
\end{proof}

\section{Additional discussions of related works}
\label{Appendix:counter examples}

\subsection{{Discussion of} \citet{huang2021optimal}}

The result of \citet{huang2021optimal} is mainly based on the noisy power method \cite{hardt2014noisy}. 
After using noisy power method to estimate $\hat{\Theta}$ such that $\| \hat{\Theta} - \Theta^* \|_F \leq \epsilon \| \Theta \|_F$, they use the fact that their arm set is a sphere and therefore the empirical best arm (greedy) is explicitly $\hat{A}=\hat{\Theta}/\|\hat{\Theta}\|_F$ and the true best arm ${A}^*={\Theta^*}/\|{\Theta^*}\|_F$.
\begin{align*}
    &\| \hat{\Theta} - \Theta^* \|_F \leq \epsilon \| \Theta^* \|_F \\
    \iff & \| \frac{\hat{\Theta}}{\|\Theta^*\|_F} - A^* \|_F \leq \epsilon
\end{align*}
and by trigonometry, one can {deduce} that $\|\hat{A}-A^*\|_F \leq \epsilon$. See \citet[][Appendix B.2]{huang2021optimal} for details.

They then use the fact $\hat{A}=\hat{\Theta}/\|\hat{\Theta}\|_F$ and ${A}^*={\Theta^*}/\|{\Theta^*}\|_F$ to achieve the instantaneous regret bound of $\epsilon^2$ as follows:
\begin{align}\label{eqn:core of huang}
\inner{\Theta^*}{A^*} -\inner{\Theta^*}{\hat{A}}= \frac{\inner{\Theta^*}{A^*}}{2} (2 - \inner{\frac{\Theta^*}{\|\Theta^*\|_F}}{\hat{A}})
= \frac{\|\Theta^*\|_F}{2}  \norm{ \frac{\Theta^*}{\|\Theta^*\|_F}-{\hat{A}}  }_F^2 
\leq \| \Theta \|_F \epsilon^2
\end{align}
This small $\epsilon^2$ error guarantee (as opposed to, say, $\eps$ described below) is crucial for obtaining their regret bound.

% helped them to achieve small bandit regret bound. 

To summarize, a key property~\citet{huang2021optimal} used was the fact when $\Theta^*$ and $\hat{\Theta}$ are close enough, then $A^*$ and $\hat{A}$ is also close enough in their setting. This is true when the arm set $\cA$ has a smooth curvature. However, without curvature on the arm set, the greedy arm $\hat{A} = \argmax_{A \in \cA} \inner{\hat{\Theta}}{A}$ can only be guaranteed such that 
\[
\inner{\Theta^*}{A^*} -\inner{\Theta^*}{\hat{A}}
\leq 
2 \max_{a \in \cA} |\inner{\hat{\Theta} - \Theta^*}{A}|
\leq 
O( \epsilon )
\]

Here's one example that shows the importance of the Frobenius norm unit ball arm set for their anaylsis. Suppose that arm set $\cA=\cB \cup \{\diag(1,1,0,\cdots,0)\}$, where $\mathcal{B}=\{M\in\RR^{d\times d}:\|M\|_F\leq 1\}$. Consider  $\Theta^*=\diag(1,\epsilon, 0, \cdots, 0)$ for some small $\epsilon$. Suppose that we run the algorithm of \citet{huang2021optimal} using $\cB$. Then, for an arbitrary estimation error $\epsilon_h$, the for the estimator using \citet{huang2021optimal}, $\hat{\Theta}_{h}$, we have guarantee $\|\hat{\Theta}_h-\Theta^*\|\leq \epsilon_h \|\Theta^*\|_F$ when $n_0= \tilde{O}(d^2 r \lambda_r^{-2}\epsilon_h^{-2})$ is number of total exploration steps (From Theorem 3.8 of~\cite{huang2021optimal}). 
As we have stated above, \citet{huang2021optimal} converted this to a bound of $\|\hat{A}-A^*\|_F$ when the arm set was $\cB$. 
However, in the case when the arm set is $\cA$, $\hat{A}$ and $A^*$ can be close enough only when $(\hat{\Theta}_h )_{22}$ is positive. If not, then we have $\inner{\Theta^*}{\diag(1,1,0,\cdots,0)} - \max_{A\in \cB} \inner{\Theta^*}{A} = \Omega(\epsilon)$ and this incurs $\epsilon T$ exploitation regret. To guarantee $\epsilon_h \leq \epsilon$, we need to spend $\tilde{O}(d^2 r \lambda_r^{-2}\epsilon^{-2})$ samples for exploration. Thus, with this analysis, the best regret upper bound we can hope for is $$ \min(\epsilon T, \frac{d^2 r}{\lambda_r^2 \epsilon^2}).$$
Choosing $\epsilon$ that maximizes this leads to $\tilde{O}((d^2 r T^2 \lambda_r^2)^{1/3} )$ regret upper bound. This is much worse than their previous bound $\tilde{O}(\sqrt{d^2 r T}{/\lambda_r})$.

\subsection{{Discussion of} \citet{kang2022efficient}} \label{appsubsec: kang et al}
The result of \citet{kang2022efficient} is directly associated with a sampling distribution constant called $M$, which was treated as a constant unrelated to dimensionality in the paper. 
However, we explain here that $M$, {has hidden dependence on the dimensionality}.

%Here is an explicit example how bad their algorithm can be. 
To see this, consider the reward model $y_t = \inner{\Theta^*}{X_t} + \eta_t$ where $\eta_t \sim N(0, \sigma^2)$. It lies in the 
(conditional) canonical exponential family:
\[
p_{\Theta^*}(y_t \mid X_t)
= 
\exp\del{ \frac{y_t \beta - b(\beta)}{\phi} + c (y_t, \phi) },
\]
where $\beta = \inner{\Theta^*}{X_t}$, 
$b(\beta) = \frac{1}{2} \beta^2$, $\phi = \sigma^2$, $c(y_t, \phi) = \ln\del{\frac{1}{\sqrt{2\pi \sigma^2}}} - \frac{y_t^2}{2\sigma^2}$. 
The inverse link function is $\mu( \beta ) = \nabla b( \beta ) = \beta$.

%(mean function)
%Suppose our arm set is
Consider arm set $\cA = \cbr{X \in \RR^{d_1 \times d_2}: \| X \|_F \leq 1}$.  We consider $\cD = N(0, \frac{c}{d_1 d_2} I_{d_1 d_2})$ (with $c = O(\frac{1}{\ln T})$), so that $T$ arms drawn iid from $\cD$ all lie in $\cX$ with probability $1-O(\frac1T)$.
With this distribution, 
\[
p(X) \propto \exp(-\frac{\| X \|_F^2}{\frac{2c}{d_1 d_2}}) 
\implies
\ln p(X)
= -\frac{d_1 d_2 \| X \|_F^2}{2c} + \mathrm{constant},
\]
Therefore, the associated score function 
$S(X) = \nabla \ln p(X) = -\frac{d_1 d_2}{c} X$.

Now, checking~\citep[Assumption 3.3]{kang2022efficient}, we have for all $i,j$,
\[
\EE\sbr{S(X)_{i,j}^2}
=
\frac{(d_1 d_2)^2}{c^2} \EE\sbr{ X_{i,j}^2 }
= \frac{d_1 d_2}{c} 
\]
As $M$ is chosen such that for all $i,j$, $\EE\sbr{S(X)_{i,j}^2} \leq M$, $M$ has to be at least $\frac{d_1 d_2}{c} \geq d_1 d_2$.

Plugging this into ~\citep[Theorem 4.1]{kang2022efficient}, and note that $\mu^* = \EE\sbr{\mu'(\inner{X}{\Theta^*})} = 1$, 
we have that given $T_1$ iid samples from $\cD$, the estimator $\hat{\Theta}$ has a Frobenius recovery error bound of: 
\[
\| \hat{\Theta} - \Theta^* \|_F^2
\leq 
\tilde{O}( (\sigma^2 + S_f^2) \frac{d_1 d_2 d r }{T_1} ).
\]
\edit{COMMENTED BELOW}{As one can see from the result above, and as they have revised later, their new bound is actually no better than the known low-rank bound $\tilde{O}(\sigma^2 \frac{d_1 d_2 d r}{T})$ of \cite{jun19bilinear} and \cite{lu2021low}, and shows the importance of correct description of the arm-set-dependent parameter. }

For \citet{kang2022efficient}, they also stated their result based on the Frobenius norm bounded arm set: $\cA \subset \{A\in \RR^{d_1 \times d_2}: \|A\|_F\leq 1 \}$. When we change the Frobenius norm bound to operator norm bound, their estimation bound (\citet[Theorem 4.1]{kang2022efficient}) does not change much, but their regret analysis on ESTS needs additional $d^{0.25}$ factor. This additional dimensional dependence also applies for all ESTR-based algorithms \citep{jun19bilinear, lu2021low} and it is because of the log-determinant term computation - check Lemma \ref{lem:jun logdet} and Lemma \ref{lem:valko logdet} to see details of why additional $d$ appears. 

\subsection{Justifying regret bound of \citet{lu2021low} in Table \ref{table:comparison}}\label{appsubsec:d14 modification}

In this section, we show that the regret bound of LowESTR~\cite{lu2021low} (originally proposed for the setting of $\cA \subset B_{\frob}(1)$), when applied to our setting ($\cA \subset B_{\op}(1))$, gives a regret bound of $\tilde{O}( d^{1/4}\sqrt{r\frac{\sigma^2}{\lambda_{\min}(Q(\pi))^2} T}\del{\frac{S_*}{\lambda_{r}}})$.
First, with the new assumption on the arm set $\cA$, it is necessary to set $\lambda_\perp = \frac{T}{r}$ instead of $\frac{T}{r d}$ in~\cite{jun19bilinear,lu2021low} to ensure that $\log\frac{|V_T|}{|\Lambda|} \leq \tilde{O}(r d)$. 

Therefore, the total regret bound of LowESTR is 
\[
\tilde{O}\del{ 
S_* n_0 
+  
\sigma k\sqrt{T} + B\sqrt{k\lambda T} + B_\perp \sqrt{k\lambda_{\perp}T}
}
\]

\begin{theorem}[Lemma 23 and Appendix E.2 of \cite{lu2021low}] For the nuclear norm regularized least square estimator $\hat{\Theta}_{nuc}$, we have
    $$\|\hat{\Theta}_{nuc}-\Theta^*\|_F^2 \leq 4.5 \frac{\lambda_n^2}{\kappa^2} r \approx \frac{\sigma^2}{n \lambda_{\min}(Q(\pi))^2} \cdot r$$
where $\kappa$ is the restricted strong convexity constant (in \cite{lu2021low} it is $\lambda_{\min} (Q(\pi))$), and $\lambda_n$ is a constant which satisfies $\|\frac{1}{n}\sum_{t=1}^n \eta_t X_t\|_{\op} \leq \frac{\lambda_n}{2}$ (it is $O(\sqrt{\frac{ \sigma}{n}})$; by \cite{tsybakov2011nuclear}, Propositioin 2). 
% \ks{Here's the place where I made a miscalculation maybe. It was $\lambda_n = \frac{d\lambda_{\max}(Q)}{n}$ for some reason (I couldn't find the reasoning now), which is kind of too lenient. I used the lemma of professor Chicheng, and made a recalculation. Please check. }
% \chicheng{
% Kyoungseok, that lemma was bounding 
% $\frac1n \sum_{i=1}^n A_i y_i - \EE_{A \sim \Pi} \sbr{\inner{\Theta^*}{A} A}$; the second term is a population average instead of $\frac{1}{n}\sum_{i=1}^n \inner{A_i}{ \Theta^*} A_i$.
% We can apply the same~\cite{tsybakov2011nuclear}'s Proposition 2 directly (it is some generalization of Matrix Bernstein's inequality) by defining $Z_i = A_i \eta_i$ though. I think $\lambda_n = O(\sqrt{\frac{\sigma^2}{n}})$ suffices. 
% }
\end{theorem}

% \edit{For $\textsf{Sig}(\pi)$, one could use $\lambda_{\max}(Q(\pi))$ instead.}{} \kj{I did not follow how $Sig(\pi)$ is being treated like a scalar. I thought the matrix sub-gaussian parameter is a matrix? See Definition 9.2 of \url{https://www.stat.cmu.edu/~siva/teaching/709/lec9.pdf}}

%\chicheng{I think it is better to make a claim that directly related to the main text. Perhaps just avoid mentioning $\textsf{Sig}(\pi)$ and state the bound in terms of $\lambda_{\max}(Q(\pi))$?}

%\ks{Main point: First, emphasize what they did - used the bound of Frobenius norm so they are 'forced' to use Frob-Frob-Frob product instead of op-Frob-op product as we did. The factor $r$ comes directly from this fact. }

%their \emph{bounds} on $\|U_\perp^\top U^*\|_{\op}$ and $\|V_\perp^\top V^*\|_{\op}$ are at least a factor $\sqrt{r}$ looser than ours, 
Under this result, they are forced to use the Frobenius version of Wedin's Theorem and trivially bound $\|U_\perp^\top U^*\|_{\op}$ by $\|U_\perp^\top U^*\|_{F}$ (marked as (opF) in Eq. \eqref{appeq: Bperp of Lu}.
%, plus, their recovery guarantee on $\Theta^*$ is stated in Frobenius norm.
%their $\|U_\perp^\top U^*\|_{F}$ and $\|V_\perp^\top V^*\|_{F}$ are at least $\sqrt{r}$ times larger than our $\|U_\perp^\top U^*\|_{\op}$ and $\|V_\perp^\top V^*\|_{\op}$. 
%\chicheng{I agree with Kwang -- To be precise, Kyoungseok, I think what you intend to say is something like ``Their \emph{bounds} on $\|U_\perp^\top U^*\|_{\op}$ and $\|V_\perp^\top V^*\|_{\op}$ are at least a factor $\sqrt{r}$ looser than ours, since they use the Frobenius version of Wedin's Theorem and trivially bound $\|U_\perp^\top U^*\|_{\op}$ by $\|U_\perp^\top U^*\|_{F}$, plus, their recovery guarantee on $\Theta^*$ is weaker.''}
This leads to the following looser estimation:

\begin{align}\label{appeq: Bperp of Lu}
    \| \hat{U}_\perp^\top \Theta \hat{V}_\perp \|_{F} 
\leq
\| \hat{U}_\perp^\top U\|_{\op} \cdot \| \Sigma \|_F \cdot \| V^\top \hat{V}_\perp \|_{\op} 
\overset{(opF)}{\leq}
\| \hat{U}_\perp^\top U\|_{F} \cdot \| \Sigma \|_F \cdot \| V^\top \hat{V}_\perp \|_{F} 
\leq 
\frac{\sigma^2 \| \Theta^* \|_F }{\lambda_{\min}(Q(\pi))^2 \cdot n_0 \lambda_r (\Theta^*)^2} \cdot r
\end{align}

Note that there's $r$ term on RHS now. Since $\frac{1}{\lambda_{\min}(Q(\pi))}\geq \frac{1}{C_{\min}} \geq \frac{B_{\min}(\cA)}{d} \geq d$ by Lemma \ref{lem: Upper bound Bmin}, \lpopart version bound is much tighter than Eq. \eqref{appeq: Bperp of Lu} in all manners.

Now from the construction, $B_\perp 
\leq 
\frac{\sigma^2 \| \Theta^* \|_F }{\lambda_{\min}(Q(\pi))^2 \cdot n_0 \lambda_r (\Theta^*)^2} \cdot r 
\leq 
\frac{\sigma^2 \sqrt{d} S_* }{\lambda_{\min}(Q(\pi))^2 \cdot n_0 \lambda_r (\Theta^*)^2} \cdot r 
$. 

Therefore, the total regret of LowESTR can be bounded by 
\begin{align*}
Reg(T)
\leq & 
\tilde{O}\del{ n_0 S_*
+
\sigma r d \sqrt{T}
+ 
\frac{T d^{0.5} \sigma^2 S_* }{n_0 \lambda_{\min}(Q(\pi))^2 \lambda_r (\Theta^*)^2}\cdot r} 
\\
\leq & 
\tilde{O}\del{\sigma r d \sqrt{T}
+ 
\sigma \sqrt{ S_*^2  \frac{d^{0.5} \sigma^2}{ {\lambda_{\min}(Q(\pi))^2 \lambda_r (\Theta^*)^2}} T \cdot r}}
\end{align*}
with the optimal tuning of $n_0$.

\begin{remark}
    As mentioned in Remark \ref{rem:etc-improve}, our LPA-ESTR also achieves an improved regret guarantee over LowESTR (\cite{lu2021low}) not only w.r.t. $d$ but also w.r.t. rank $r$ too. 

The main reason is that the \lpopart provides operator norm-based recovery bound as discussed in Theorem \ref{thm:lpopart}. This allows us to use the operator norm version of Wedin Theorem (See Section \ref{appsec: proof of estr bound}), which means we obtained the bound of $\|U_\perp^\top U^*\|_{\op}$ and $\|V_\perp^\top V^*\|_{\op}$. From this bound, we used the fact that $\|AB\|_F \leq \|A\|_{\op} \|B\|_F$ to derive the following relationship:

 \begin{align}\| \hat{U}_\perp^\top \Theta \hat{V}_\perp \|_{F} 
\leq
\| \hat{U}_\perp^\top U\|_{\op} \cdot \| \Sigma \|_F \cdot \| V^\top \hat{V}_\perp \|_{\op} 
\leq 
\frac{\ldyad \sigma^2 \| \Theta^* \|_F }{n_0 \lambda_r (\Theta^*)^2} \tag{This is \lpopart version.} \end{align}
% \kj{[ ] are we using Fenchel-Young inequality (or AM-GM inequality) above? Would be great to spell it out.}
% \chicheng{I think we are using $\|A B\|_F \leq \| A\|_{\op} \| B\|_F$}

Remember that there's no $r$ term on the RHS. On the other hand, \cite{lu2021low} used the Frobenius norm version of the Wedin Theorem, since they mainly used the Frobenius norm bound of the nuclear norm regularized least square.

\end{remark}

% \chicheng{Perhaps add some context and explain more explicitly. E.g.
% ``In this section, we show that the regret bound of LowESTR~\cite{lu2021low} (originally proposed for the setting of $\cA \subset B_{\frob}(1)$), when applied to our setting ($\cA \subset B_{\op}(1))$'', gives a regret bound of $\tilde{O}( d^{1/4}\sqrt{r\frac{d \cdot \lambda_{\max}(Q(\pi))}{\lambda_{\min}(Q(\pi))^2} T}\del{\frac{S_*}{\lambda_{r}}})$.
% First, with the new assumption on the arm set $\cA$, it is necessary to set $\lambda_\perp = \frac{T}{r}$ instead of $\frac{T}{r d}$ in~\cite{jun19bilinear,lu2021low} to ensure that $\log\frac{|V_T|}{|\Lambda|} \leq \tilde{O}(r d)$. 

% Therefore, the total regret bound of LowESTR is 
% \[
% \tilde{O}\del{ 
% S_* n_0 
% +  
% \sigma k\sqrt{T} + B\sqrt{k\lambda T} + B_\perp \sqrt{k\lambda_{\perp}T}
% }
% \]
% with $B_\perp = $(perhaps move the bound on $\| \hat{U}_\perp^\top \Theta \hat{V}_\perp \|_F \leq ..$ to here).
% }

% In Appendix \ref{appsec: proof of estr bound}, starting from Lemma \ref{lem:valko logdet}, the results are slightly different from the original proof of \cite{jun19bilinear} because we have different Frobenius norm bound (note that for all $a \in \cA$, $\|a\|_F \leq \sqrt{d}\|a\|_{\op} \leq \sqrt{d}$. Since \cite{lu2021low} heavily depends on the proof of \cite{jun19bilinear}, they should also suffer additional $d^{1/4}$ regret in the operator norm bounded arm set. 

\subsection{Comparison with \citet{jedra2024low}}
As mentioned in Appendix \ref{sec:relwork}, many studies on low-rank bandits focus on cases where the bandit instance has a special arm set. A notable example is the recent work by \citet{jedra2024low}, who studied contextual bandits with a low-rank structure where, in each round, if the (context, arm) pair $(i, j) \in [m] × [n]$ is selected, the learner observes a noisy sample of the $(i, j)$-th entry of an unknown low-rank reward matrix. This can be seen as a low-rank bandit with a canonical arm set, that is, $\cA=\{\text{reshape}(e_s): s=1, ..., d_1 d_2\}$, where they were able to obtain a regret bound of $O(r^{7/4} d^{3/4} \sqrt{T})$. Our paper can also be applied to this setting, and in that case, the regret bound is $O(d^{3/2}\sqrt{T})$. At first glance, the algorithm of \citet{jedra2024low} seems superior to ours, but this is because they focus only on a specific setting where the arm set consists of canonical vectors.

Such examples can be found in various instances in the bandit world. For instance, as mentioned in \citep[Section 23.3]{lattimore20bandit}, the $K$-armed bandit can be considered a linear bandit with exactly $K$ dimensions where all arms are canonical vectors. The UCB algorithm for K-armed bandits has a regret bound of $O(\sqrt{KT})$. On the other hand, OFUL \cite{ay11improved}, the algorithm known to be minimax optimal for linear bandits in general, has a regret bound of $O(d\sqrt{T})$ in a $d$-dimensional space, which becomes $O(K\sqrt{T})$ in this specific case, thus larger than the regret bound of UCB. However, one cannot claim that UCB is a superior algorithm to OFUL because the generality of instances each algorithm can handle is different. 

{In addition, the difference in the assumption about the unknown parameter $\Theta^*$ also affects the result. Their algorithms operate (and are optimized for) $\max_{i,j} |\Theta_{ij}|$-bounded setup, whereas our paper operates under $||\Theta||_{*}$-bounded setup.}

\ks{Regret bound mention? One may attempt to apply ours to Jedra et al setting, but it makes our work looks worse than theirs. (Regret bound 1 2). However, this work is not comparable with ours = conclusion. Even in linear bandit, bound does not match - K-armed mab vs linear bandit. }

\subsection{Justifying arm-set dependent constant of \citet{jun19bilinear} in Table \ref{table:comparison}}
Consider $\pi$ to be the uniform distribution of $\cX_0 \times \cZ_0$, where $\cX_0 = \cbr{X_1, \ldots, X_{d_1}}$ and $\cZ_0 = \cbr{Z_0, \ldots, Z_{d_2}}$ are sets of linearly independent vectors in $\cX$ and $\cZ$, respectively. {This is exactly how \cite{jun19bilinear} have sampled. They achieved the regret bound of on $\tilde{O}(\|X^{-1}\|_{op} \|Z^{-1}\|_{op} d^{3/2}\sqrt{rT})$ where $X:= [X_1; \cdots,; X_{d_1}]$ and $Z:=X:= [Z_1; \cdots,; Z_{d_2}]$. In this section, we show that this $\|X^{-1}\|_{op} \|Z^{-1}\|_{op}$ is actually $\sqrt{\frac{1}{d_1 d_2\lambda_{\min}(Q(\pi))}}$, and therefore must be larger than or equal to $\sqrt{\frac{1}{d_1 d_2 C_{\min}(\cA)}}$. }

\begin{align*}
    d_1 d_2 Q\edit{}{(\pi)} &= \sum_{i=1}^{d_1} \sum_{j=1}^{d_2} (X_i \otimes Z_j )(X_i \otimes Z_j )^\top\\
    &=\sum_{i=1}^{d_1} \sum_{j=1}^{d_2} (X_i \otimes Z_j )(X_i^\top \otimes Z_j^\top )\\
    &=\sum_{i=1}^{d_1} \sum_{j=1}^{d_2} (X_i X_i^\top) \otimes (Z_j Z_j^\top )\\
    &=\sum_{i=1}^{d_1} (X_i X_i^\top) \otimes (Z Z^\top)\\
    &= (XX^\top) \otimes (ZZ^\top)
\end{align*}
Therefore, 
\begin{align*}
    \frac{1}{d_1 d_2}||Q^{-1}||_{\op} &=||[(XX^\top) \otimes (ZZ^\top)]^{-1}||_{\op}\\
    &=||[(XX^\top)^{-1} \otimes (ZZ^\top)^{-1}]||_{\op}\\
    &=||[XX^{\top}]^{-1}||_{\op} ||[ZZ^{\top}]^{-1}||_{\op}\\
    &= ||X^{-1}||_{\op}^2 ||Z^{-1}||_{\op}^{2}
\end{align*}

\section{Comparison between our algorithm and~\citet{tsybakov2011nuclear}}\label{Appendix:comparison with Tsybakov}

%will be filled later. 

%$y_i = \inner{\Theta}{X_i} + \eta_i$ for $\eta_i$'s being iid zero-mean random variables,

Suppose we are given $(A_i, y_i)_{i=1}^n$ iid samples such that $A_i \sim \Pi$ and $\Pi$ is supported on $\cbr{A: \| A \|_{\op} \leq 1}${, and for every $i$, $y_i = \inner{\Theta^*}{A_i} + \eta_i$, where $\eta_i$'s are independent zero-mean $\sigma$-subgaussian noise.}
~\cite{tsybakov2011nuclear} considers a nuclear-norm penalized estimator,  
defined as follows: 
\begin{equation}
\hat{\Theta} = \argmin_{\Theta} \| \Theta \|_{L_2(\Pi)}^2 - \inner{\frac{2}{n}\sum_{i=1}^n y_i A_i}{\Theta} + \lambda \| \Theta \|_*, 
\label{eqn:nuclear-norm-penalized}
\end{equation}
where $\| B \|_{L_2(\Pi)} = \sqrt{ \EE_{A \sim \Pi} \inner{A}{B}^2 }$.

%\EE_{A \sim \Pi} \sbr{ \inner{A}{\Theta}^2 }
%\begin{definition}
%Define 
%\end{definition}

%Suppose we are given $(A_i, y_i)_{i=1}^n$ iid samples such that $A_i \sim \Pi$.
\begin{theorem}[Adapted from~\cite{tsybakov2011nuclear}, Corollary 1]\label{thm:tsybakov}
Given the setting above, and suppose additionally that: 
\begin{itemize}
\item there exists $C > 0$ such that for all $B$, $\| B \|_{L_2(\Pi)}^2 \geq C \| B \|_F^2$, 
\item rank-$r$ matrix $\Theta_0$ is such that $\| \frac1n \sum_{i=1}^n A_i y_i - \EE_{A \sim \Pi} \sbr{\inner{\Theta_0}{A} A} \|_{\op} \leq \frac{\lambda}{2}$. 
\end{itemize}
Then, there exists some absolute constant $c > 0$ such that  
\[
\| \hat{\Theta} - \Theta_0 \|_{F}
\leq c \frac{\sqrt{r} \lambda}{C},
\quad 
\| \hat{\Theta} - \Theta_0 \|_{F}
\leq c \frac{r \lambda}{C}.
\]
\end{theorem}

Now the Lemma \ref{lem:m-concentration} below states that $\Theta^*$ satisfies the condition of $\Theta_0$ in Theorem \ref{thm:tsybakov}.

\begin{lemma} 
\label{lem:m-concentration}
% \kj{I am guessing that you are assuming $\EE[y_i] = \EE[\la A_i, \Th_0\ra]$ here? my impression is that Theorem 8 does not necessarily assume realizability, but I could be wrong.}
% \chicheng{Yes, we were assuming realizability. I added assumptions on the data generation process.}
Suppose $n \geq O(\ln\frac{d}{\delta})$. Then 
with probability $1-\delta$, 
\[
\norm{ \frac1n \sum_{i=1}^n A_i y_i - \EE_{A \sim \Pi} \sbr{\inner{\Theta^*}{A} A} }_{\op} 
\leq
O\del{ (S_* + \sigma) \sqrt{\frac{\ln\frac{d}{\delta}}{n}} }
\]
\end{lemma}
% \del{ } I(|y_i| \leq S_* + \sigma \sqrt{\ln\frac{n}{\delta}})
\begin{proof}
Let $Z_i =A_i y_i - \EE_{A \sim \Pi} \sbr{\inner{\Theta^*}{A} A} $. We first upper bound $\| Z_i \|_{\op}$'s $\psi_2$-Orlicz norm; to this end, first note that 
\[
\norm{ \| A_i y_i \|_{\op} }_{\psi_2}
\leq 
\norm{ | y_i | }_{\psi_2}
\leq 
\norm{ | \inner{\Theta}{A_i} | }_{\psi_2}
+ 
\norm{ |\eta_i| }_{\psi_2}
\leq 
S_* + \sigma
\]
Therefore, $\EE_{A \sim \Pi} \sbr{\inner{\Theta^*}{A} A} = \EE[A_i y_i]$ also satisfies that 
\[ 
\norm{ \| \EE_{A \sim \Pi} \sbr{\inner{\Theta^*}{A} A} \|_{\op} }_{\psi_2}
\leq 
S_* + \sigma,
\]
hence $\| Z_i \|_{\psi_2} \leq \norm{ \| A_i y_i \|_{\op} }_{\psi_2} + \norm{ \| \EE_{A \sim \Pi} \sbr{\inner{\Theta^*}{A} A} \|_{\op} }_{\psi_2} \leq 2(S_* + \sigma)$.

Meanwhile, 
\[
\| \EE[Z_i Z_i^\top] \|_{\op}
\leq 
\| \EE[A_i A_i^\top y_i^2] \|_{\op}
= 
\| \EE[A_i A_i^\top (\inner{\Theta^*}{A_i}^2 + \sigma^2)] \|_{\op} 
\leq S_*^2 + \sigma^2,
\]
likewise, 
\[
\| \EE[Z_i^\top Z_i] \|_{\op}
\leq S_*^2 + \sigma^2.
\]
Therefore, applying Proposition 2 of~\cite{tsybakov2011nuclear}\footnote{The original proposition statement is stated for the setting of $\sigma_Z^2 = \max( \EE[Z_i Z_i^\top], \EE[Z_i^\top Z_i] )$ exactly; it can be checked that the proposition continues to hold when $\sigma_Z^2 \geq \max( \EE[Z_i Z_i^\top], \EE[Z_i^\top Z_i] )$.} on $Z_1, \ldots, Z_n$, with $\sigma_Z = S_* + \sigma$, $\alpha = 2$, and $U_Z^{(\alpha)} = 2(S_* + \sigma)$, $t = \ln \frac1\delta$ gives that with probability $1-\delta$,
\[
\norm{ \frac1n \sum_{i=1}^n Z_i}_{\op}
= 
O\del{ 
 \sigma_Z \sqrt{ \frac{\ln\frac{d}{\delta}}{n} } + U_Z^{(\alpha)} \sqrt{\ln\frac{U_Z^{(\alpha)}}{\sigma_Z}} \frac{\ln\frac{d}{\delta}}{n}}
\leq 
O\del{ (S_* + \sigma) \sqrt{\frac{\ln\frac{d}{\delta}}{n}} }.
\]

%c \del{
%for some absolute constant $c > 0$. 

%By union bound, with probability $1-\delta/2$, $Z_i =  A_i y_i - \EE_{A \sim \Pi} \sbr{\inner{\Theta_0}{A} A}$ for all $i$.

% Meanwhile, applying matrix Bernstein's inequality, we have 
% \[
% \| \frac1n \sum_{i=1}^n Z_i \|
% \]

\end{proof}

Applying the theorem to $(A_i, y_i)_{i=1}^n$ with $\Theta_0$ set to be $\Theta^*$, where $A_i \sim \pi^*$ as defined in~\eqref{eq: def of Cmin}, we can choose $C = C_{\min}(\cA)$.
On the other hand, Lemma~\ref{lem:m-concentration} below shows that choosing 
$\lambda = O\del{ (S_* + \sigma) \sqrt{\frac{\ln\frac{d}{\delta}}{n}} }$, 
with probability $1-\delta$, $\| \frac1n \sum_{i=1}^n A_i y_i - \EE_{A \sim \Pi} \sbr{\inner{\Theta_0}{A} A} \|_{\op} \leq \frac{\lambda}{2}$. 
Therefore, we conclude that with the above setting of $\lambda$ and $\Pi = \pi^*$,
the nuclear norm penalized estimator $\hat{\Theta}$ defined in Eq.~\eqref{eqn:nuclear-norm-penalized} with 
satisfies that 
\[
\| \hat{\Theta} - \Theta^* \|_{F}
\leq \tilde{O}\del{ \frac{S_* + \sigma}{C_{\min}(\cA)} \sqrt{\frac{r}{n}} },
\quad 
\| \hat{\Theta} - \Theta^* \|_{*}
\leq \tilde{O}\del{ \frac{S_* + \sigma}{C_{\min}(\cA)} \sqrt{\frac{r^2}{n}} }.
\]

%such that with probability $1-\delta$, .
%we have that the nuclear-norm penalized estimator~\eqref{eqn:nuclear-norm-penalized} 

\section{Experimental details settings}\label{appsec: experiment settings}

\subsection{Experiment settings}
Common settings
\begin{itemize}
    \item Computation resource: Apple M2 Pro, 16GB memory.
    \item Error bar: 1-standard deviation for the shadowed area. 
    \item We attached our code as supplementary material and will upload a public link when this paper is accepted. Please read \textbf{README.md} file before running.   
\end{itemize}

\subsubsection{Figure 2 left}
\begin{itemize}
    \item Dimension $d_1 = d_2 = 3$
    \item Time steps: from 1000 to 10000, increased by 1000
    \item $\Theta^* = u v^\top$, where $u$ and $v$ are drawn from $\mathbb{S}^{d_1-1}$ and $\mathbb{S}^{d_2-1}$, respectively ($\mathbb{S}^{d-1}$ is the $d$-dimensional unit sphere.)
    \item Action set $\cA$ is drawn uniformly at random from the $\cB_{Frob}(1)$. $|\cA|=150$. 
    \item Noise $\eta_t \sim N(0, 1)$, which means $\sigma^2 = 1$. 
    \item Repeated the experiment 60 times
\end{itemize}

\subsubsection{Figure 2 right}

\begin{itemize}
    \item Dimension $d_1 = d_2 = 3$
    \item Time steps: from 10000 to 100000, increased by 10000
    \item $\Theta^* = u v^\top$, where $u$ and $v$ are drawn from $\mathbb{S}^{d_1-1}$ and $\mathbb{S}^{d_2-1}$, respectively ($\mathbb{S}^{d-1}$ is the $d$-dimensional unit sphere.)
    \item Action set $\cA$ is $\cA_{hard}$, which is defined as follows:
    \begin{align*}
        a_i = \begin{cases}
            \mathsf{reshape}(\frac{1}{\sqrt{3}} e_1) & \text{if $i=1$}\\
            \mathsf{reshape}(e_1 + \frac{1}{\sqrt{3}}e_i) & \text{if $i=2,3, \cdots, d_1 d_2$}
        \end{cases}
    \end{align*}
    \item Noise $\eta_t \sim N(0, 1)$, which means $\sigma^2 = 1$. 
    \item Repeated the experiment 60 times
\end{itemize}

\subsubsection{Figure 3 left}

\begin{itemize}
    \item Dimension $d_1 = d_2 = 5$
    \item Time steps: 100000
    \item $\Theta^* = u v^\top$, where $u$ and $v$ are drawn from $\mathbb{S}^{d_1-1}$ and $\mathbb{S}^{d_2-1}$, respectively ($\mathbb{S}^{d-1}$ is the $d$-dimensional unit sphere.)
    \item Action set $\cA$ is drawn uniformly at random from the $\cB_{Frob}(1)$. $|\cA|=100$.    \item Noise $\eta_t \sim N(0, 1)$, which means $\sigma^2 = 1$. 
    \item Repeated the experiment 60 times
\end{itemize}

\subsubsection{Figure 3 right}

\begin{itemize}
    \item Dimension $d_1 = d_2 = 6$
    \item Time steps: 100000
    \item $\Theta^* = u v^\top$, where $u$ and $v$ are drawn from $\mathbb{S}^{d_1-1}$ and $\mathbb{S}^{d_2-1}$, respectively ($\mathbb{S}^{d-1}$ is the $d$-dimensional unit sphere.)
    \item Action set $\cA$ is in bilinear setting. Which means, $\cA = \{xz^{\top}: x \in \cX, z \in \cZ \}$ where $\cX$ and $\cZ$ are drawn uniformly at random from the $\mathbb{S}^{d_1 -1}$ and $\mathbb{S}^{d_2 -1}$, respectively. $|\cX|=4d_1 = 24$, $|\cZ|=4d_2 = 24$.
    \item Noise $\eta_t \sim N(0,1)$, which means $\sigma^2 = 1$. 
    \item Repeated the experiment 60 times
\end{itemize}

\subsection{Algorithm for Left figures of Figure 3}
\begin{algorithm}[h]
\caption{Nuc-ETC (Nuclear norm regularized least square based Explore then commit)}
\label{alg:etc-nuc}

\begin{algorithmic}[1]

\STATE {\bfseries Input:} time horizon $T$, arm set $\cA$, exploration lengths $n_0^{*}$, regularization parameter $\lambda$

\STATE Solve the optimization problem in Eq. \eqref{eq: def of Cmin} and denote the solution as $\pi^*$

\FOR{$t = 1, \ldots,n_0^{*}$} 
\STATE Independently pull the arm $A_t$ according to $\pi^*$ and receives the reward $Y_t$
\ENDFOR
\STATE $\hat\Theta_*:= \arg \min_{\Theta\in \mathbb{R}^{d_1 \times d_2}} \frac{1}{2}\sum_{t=1}^{n_0^*} \del{\inner{\Theta}{A_t}-Y_t}^2+\lambda \|\Theta\|_*$

\FOR{$t=n_0^*+1,\ldots,T$}
\STATE Pull the arm $X_t = \argmax_{A \in \cA} \inner{\hat{\Theta}_*}{A}$
\ENDFOR
\end{algorithmic}
\end{algorithm}

\subsubsection{Theoretical analysis of the exploration length $n_0^*$}
As discussed in Appendix \ref{Appendix:comparison with Tsybakov}, we have the following guarantee for the nuclear norm error bound of the nuclear norm regularized least square estimator:
$$ \|\hat{\Theta}-\Theta^*\|_*\leq \tilde{O}\del{\frac{S_*+\sigma}{C_{\min} (\cA)}}\sqrt{\frac{r^2}{n_0^*}}$$

Also, we have the following upper bound of the instantaneous regret after $n_0^*$:
    \begin{align*}
        \inner{\Theta^*}{A^* - A_t} &=\inner{\Theta^* - \hat{\Theta}}{A^*}+\inner{\hat\Theta}{A^*}-\inner{\Theta^*}{A_t}\\
        &\leq \inner{\Theta^* - \hat{\Theta}}{A^*} + \inner{\hat\Theta-\Theta^*}{A_t} \tag{Definition of $A_t$}\\
        &\leq \|\Theta^* - \hat{\Theta}\|_* (\|A^*\|_{\op} + \|A_t\|_{\op}) \tag{Holder's inequality}\\
        &\leq 2\|\Theta^* - \hat{\Theta}\|_* 
    \end{align*}
Overall,the regret is 
\begin{align*}
    \Reg_T &= \sum_{t=1}^{T}\inner{\Theta^*}{A^* -A_t} = \sum_{t=1}^{n_0^*}\inner{\Theta^*}{A^* -A_t} +\sum_{t=n_0^*+1}^{T}\inner{\Theta^*}{A^* -A_t}\\
    &\leq S_* n_0^* +  \tilde{O}\del{\frac{S_*+\sigma}{C_{\min} (\cA)}}\sqrt{\frac{r^2}{n_0^*}} \cdot (T-n_0^*)
\end{align*}
and the $n_0^*$ which optimizes above value is $n_0^* = \del{\sigma^2 r^2 T^2 C_{\min}(\cA)^{-2}S_*^{-2}}^{1/3}$
\subsection{Computational efficiency of Algorithm \ref{alg:lowPopart}}\label{appsubsec:comp efficiency}
For estimation only (Algorithm 1), we need $O(d_1^3 d_2^3)$ for matrix inversion (Eq. (2)), $O(n_0 (d_1 d_2)^2)$ for estimators in Line 2, and $O(d_1^2 d_2)$ for SVD in Line 3 and 4,
% , and this is strict (note: kwang removed this part)
and no more computation is needed. On the other hand, \cite{tsybakov2011nuclear} and other popular tools require optimizations that have several iterations dependent on the precision requirement of the optimization. For \cite{tsybakov2011nuclear}, it requires $O(n_0 d_1 d_2)$ for each iteration. In our experiment, both were very fast (ours: 0.3 sec, \cite{tsybakov2011nuclear}: 0.1 sec). For the experimental design part, no prior work explicitly studied on experimental design in the low-rank setting as far as we know. One natural approach is to optimize the conditions of the covariance matrix such as RIP, but there is no known computationally efficient way to directly compute these quantities (See the last part of the second contribution in Section 1). Other naive approaches are A/D/E/G/V-optimalities that are used in linear experimental design. They can be optimized by traditional optimization solvers like CVXPY or MOSEK. Our algorithm could also be done in the same way since our optimization problem is also convex. (in our experiment, ours: $0.046$ sec, E-optimality: $0.039$ sec).

\subsection{\edit{}{Additional Experiments}}

\subsubsection{Effect of the thresholding process}

\begin{figure}
    \centering
\includegraphics[width=0.5\linewidth]{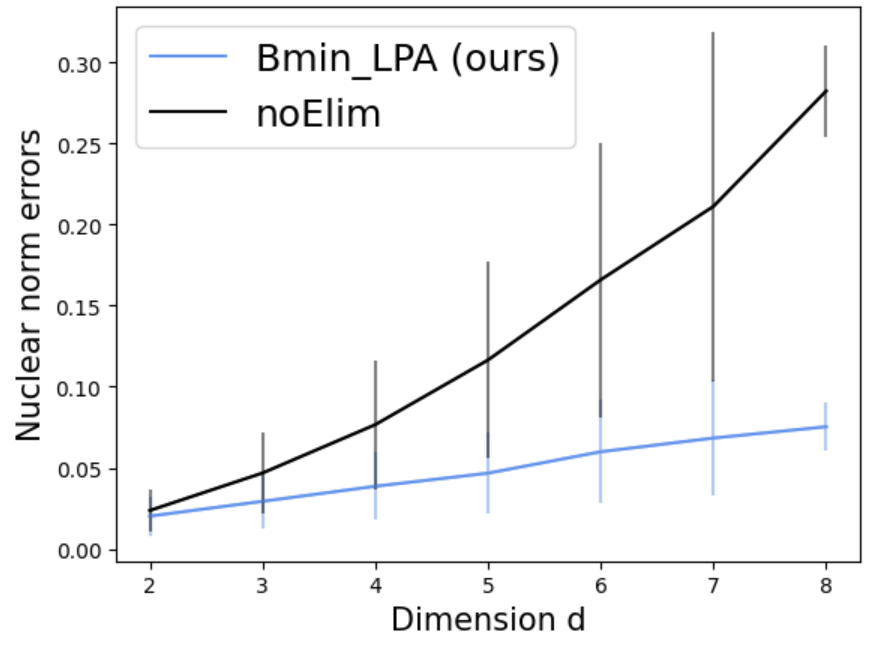}
    \caption{Experiment results on nuclear norm error}
    \label{appfig:threshold}
\end{figure}

We made an experiment to show the \edit{necessity of the HT}{utility of the hard thresholding} step. Bmin-LPA (blue) is our algorithm with hard thresholding, while noElim (black) is the algorithm without hard thresholding. As \edit{you}{we} can see in Figure \ref{appfig:threshold}, \edit{HT}{hard thresholding} step is necessary to remove noisy observations and to utilize rank information, especially when dimension $d$ gets larger.

This hard thresholding step is not that restrictive since it only eliminates singular values that are small enough that if the corresponding singular value is nonzero, \edit{}{when applied to bandit problems, we expect that} it will not harm the overall cumulative regret.

\textbf{Experiment setting}
\begin{itemize}
    \item $d_1 = d_2 =d$, $d=2,3, \cdots, 8$
\item $T=10000$,
\item $\Theta^*$ : random rank-1 matrix with $||\Theta^*||_F =1$.
\item $A \subset \mathbb{R}^{d \times d}$: uniformly drawn from the $\mathbb{S}_F^{d^2-1} (1)$ (Frobenius norm unit sphere), $|A|=4d$. Of course $A$ changes when $d$ changes.
\item Noise distribution : $N(0, 1)$.
\item Repeat the experiment 30 times for each $d$.
\end{itemize}

\subsubsection{Real-world dataset}
We used the Movielens dataset (movielens-old 100k) to try the algorithm on a real-world dataset. As one can check from Figure \ref{appfig:realdata} below, our algorithm shows superior performance compared to other traditional low-rank algorithms.
\begin{figure}
    \centering
    \includegraphics[width=0.5\linewidth]{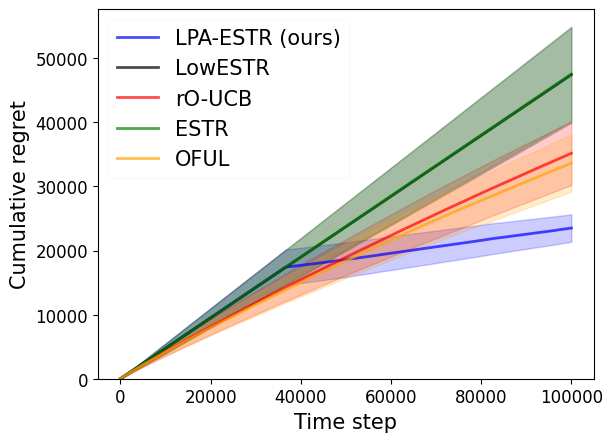}
    \caption{Experiment results on bandit using a real-world dataset}
    \label{appfig:realdata}
\end{figure}

\paragraph{Experiment setting}
\begin{itemize}
    \item $d_1 = d_2 = d=10$
    \item $X \subset \mathbb{R}^{d_1}$ and $Z \subset \mathbb{R}^{d_2}$ are the subset of the left (and right, respectively) singular vectors of the rank-d approximation from SVD result of the Movielens rating after matrix completion (KNNImputer in Scikit-learn\cite{scikit-learn}). We randomly select $|X|=50$ among 1000 users (and $|Z|=50$ among 1700 movies.)
    \item $\Theta^*$: random rank-1 matrix with $||\Theta^*||_F=1$, $r=1$.
    \item $T=10^5$
    \item Noise distribution: $N(0,1)$.
    \item Repeat each algorithm 30 times to measure the average cumulative regret.
\end{itemize}

\section{Proof of Lower Bound (Theorem~\ref{thm:lower-bound-lr})}

%Now, let's consider the low-rank setting. 

% \begin{theorem}
% For any $d$, $r$ such that $2r-1 \leq d-1$,
% $n \geq 1$, 
% $\kappa \in [ \frac{\sqrt{r}}{d} , \frac{1}{\sqrt{d}}]$, $\sigma > 0$, $R_{\max} \in [ \sigma \sqrt{ \frac{r}{n\kappa^2}}, \sigma \sqrt{ \frac{d^6 \kappa^4}{n r^2} } ]$,
% any bandit algorithm $\cA$, there exists an $(2r-1)$-rank $d \times d$-dimensional bandit environment $\Theta$ with noise $\sigma$, and arm set $\cA \subset \cbr{ a: \| a \|_{\op} \leq 1 }$, and $\max_{a \in \cA} |\inner{\Theta}{a}| \leq R_{\max}$, such that 
% \[
% \EE_{\Theta, \cA}[\Reg(\Theta, n)]
% \geq 
% \Omega(R_{\max}^{1/3} \sigma^{2/3} r^{1/3} n^{2/3} \kappa^{-2/3})
% \]
% \label{thm:lower-bound-lr}
% \end{theorem}

% \begin{proof}[Proof of Theorem~\ref{thm:lower-bound-lr}]
% \end{proof}
We first state a more precise version of Theorem~\ref{thm:lower-bound-lr}: 
\begin{theorem}[Theorem~\ref{thm:lower-bound-lr} restated]
\label{thm:lower-bound-lr-restated}
For any $d$, $r$ sufficiently large such that $2r-1 \leq d-1$, $d \geq 4000r$, 
$T \geq 1$, 
%$\kappa \in [ \frac{\sqrt{r}}{d} , \frac{1}{\sqrt{d}}]$, 
$C \in [\frac{400 r}{d^2} , \frac{1}{10 d}]$, 
$\sigma > 0$, 
%$R_{\max} \in [ \sigma \sqrt{ \frac{r}{n\kappa^2}}, \sigma \sqrt{ \frac{d^6 \kappa^4}{n r^2} } ]$,
$\rmax \in [ 125 \sigma \sqrt{ \frac{r}{T C}}, \frac{\sigma}{64} \sqrt{ \frac{d^6 C^2}{T r^2} } ]$,
any bandit algorithm $\cB$, there exists a $(2r-1)$-rank $d$-dimensional bandit environment with $\sigma$-subgaussian noise with an arm set $\cA \subset \cbr{ a: \| a \|_{\op} \leq 1 }$ such that $C_{\min}(\cA) \geq C$ and $\Theta^*$ which satisfies $\max_{A \in \cA}|\inner{\Theta^*}{A}|\leq \rmax$, such that 
\[
\EE_{\Theta, \cB}[\Reg(\Theta, T)]
\geq 
\frac{1}{64} \sigma^{2/3} \rmax^{1/3}  r^{1/3} T^{2/3} C^{-1/3}.
\]
\end{theorem}

Our lower bound instance construction and proof resembles those of sparse linear bandit lower bound argument of~\cite{hao2020high}. We make a few important modifications tailored to the low-rank bandit setting:
\begin{itemize}
\item Matrix-valued arms and hypotheses: in contrast to \citet{hao2020high} which considers vector-valued arms and hypotheses, here we consider arm sets and hypotheses in $\RR^{d \times d}$; specifically, the $(d,d)$-th entry of arm serves as penalizing the informative arms in $\cH$; the row and column subspace of the null hypothesis $\Theta$ is spanned by $\cbr{e_1, \ldots, e_{r-1}}$; similarly, the row  alternative hypothesis $\tilde{\Theta}$ is supported on some $r$-dimensional subspace of $\mathrm{span}\cbr{e_r, \ldots, e_{d-1}}$. 

\item Range of $C$, lower bound of $C_{\min}(\cA)$: \citet{hao2020high} considers the setting where all arms are $\ell_\infty$ bounded by 1, which induces a constraint that $C \leq 1$; in contrast, we consider the setting where all arms have operator norm bounded by 1, which induces a different constraint on $C$: $C \leq \frac{1}{10d}$.

\item A different averaging argument for the low-rank setting: to establish that hypotheses $\Theta$ and $\tilde{\Theta}$ have small divergence, we consider the average KL divergence between $\PP_{\Theta}$ and some $\PP_{\Theta + 2\epsilon \tilde{Z}}$, where $\tilde{Z}$ is chosen randomly from $\cS'$ (see Eq.~\eqref{eqn:def-s-prime}). This is a uncountably infinite set; we utilize symmetry of the Haar measure to bound the average KL divergence.  
\end{itemize}

\subsection{The construction}
\label{sec:construction}

\paragraph{Basic notations.} Below, for $X \in \RR^{d \times d}$, denote by $X^1 = X_{1:d-1, 1:d-1} \in \RR^{(d-1) \times (d-1)}$, and 
$X^2 = X_{r:d-1, r:d-1} \in \RR^{(d-r) \times (d-r)}$. 
Define $\inner{X}{Y}_1 = \inner{X^1}{Y^1}$, and $\inner{X}{Y}_2 = \inner{X^2}{Y^2}$. 

Consider low-rank bandit environment $r_t = \inner{\Theta}{A_t} + \eta_t$, where $\eta_t \sim N(0, \sigma^2)$ is additive Gaussian noise. As we will see in subsequent constructions, we ensure that the arm set $\cA \subset \cbr{a: \| a \|_{\op} \leq 1}$
and $C_{\min}(\cA) \geq C$.
We also ensure that for all instances $\Theta$ considered, $\| \Theta \|_* \leq \rmax$, \edit{}{so that $\max_{A \in \cA} \inner{\Theta}{A}\leq \rmax$}. 

\paragraph{Setting of parameters.} We choose $\epsilon 
=  
\del{ \frac{\rmax \sigma^2}{T r^2 C} }^{\frac{1}{3}}$. By the relationships of the parameters in the theorem statement, we have that the following happen simultaneously:
\begin{enumerate}
\item $d^2 \geq \frac{16 Tr^2\epsilon^2}{\sigma^2}$. This follows from the assumption that $\rmax \leq \frac{\sigma}{64} \sqrt{\frac{d^6 C^2}{T r^2}}$, and will be used when we upper bound the KL divergence of the two hard bandit instances we construct. 

\label{item:lower-bounding-kl}

\item $r \epsilon \leq \frac{\rmax}{24}$. This follows from the assumption that $\rmax \geq 125 \sigma \sqrt{\frac{r}{T C}}$, and 
wil be used to ensure that for the hard bandit instances constructed, $\| \Theta \|_* \leq \rmax$. 

\label{item:theta-bound}

\item $C \leq \frac{1}{10 d}$. {This requirement for $C$ is without loss of generality (up to a constant factor), as we know from Corollary \ref{appcor: Cmin is smaller than 1d} that any $\cA$ satisfying Assumption~\ref{assumption:op-bound} has $C_{\min}(\cA) \leq \frac{1}{d}$.}
\chicheng{Kyoungseok, can you add the reference after the lemma is written?}

\label{item:req-c}

\end{enumerate}

\paragraph{Action space.} Define arm set $\cA = \cH \cup \cS$, where:
\begin{itemize}
\item The ``informative and high regret'' arm set     
\[ 
    \cH =
    \cbr{
    X = 
    \begin{bmatrix}
    x_{11}, \ldots, x_{1d} \\
    \ldots \\
    x_{d1}, \ldots, \frac12
    \end{bmatrix}
    :
    \forall (i,j) \neq (d,d), 
    x_{i,j} \in \cbr{-\sqrt{2C}, \sqrt{2C}}
    \wedge
    \| X \|_{\op} \leq 1
    }
\]

%\frac12 + \sqrt{C d \ln d}
%\wedge
    % \| X_1 \|_{\op} \leq \kappa \sqrt{d \ln d}
%\red{To ensure that this has operator norm bounded by 1 we need $\kappa \leq \frac{1}{\sqrt{d}}$}

%Note that this ensures that $C_{\min}(\cA) \geq C_{\min}(\cH) \geq \kappa^2$. \chicheng{Need to prove this via rigorous concentration arguments -- the key observation is that when $X$ is uniformly randomly drawn, $\| X \|_{\op} \leq 1 + \kappa \sqrt{d \ln d}
%$ should happen with overwhelmingly high probability.}

%\red{Need to assume $r \kappa \sqrt{d} \epsilon \leq \rmax$}

%\wedge
    % \| X_1 \|_{\op} \leq \kappa \sqrt{d \ln d}

\item The ``low informative and low regret'' arm set 
\[
\cS = \cbr{ 
    \begin{bmatrix}
    U V^\top & 0 \\
    0 & 0 
    \end{bmatrix}: U, V \in \RR^{(d-1) \times (r-1)}, U^\top U = V^\top V = I_{r-1}}
\]
%\red{Need to assume $r \epsilon \leq \rmax$}
\end{itemize}

By construction, all arms in $\cH$ have operator norms at most 1; meanwhile, all arms in $\cS$ has a singular value decomposition 
$
    \begin{bmatrix}
    U \\
    0 
    \end{bmatrix}
    \begin{bmatrix}
    \diag(\one_{r-1}) & 0 \\
    0 & 0
    \end{bmatrix}
    \begin{bmatrix}
    V^\top & 0 
    \end{bmatrix}
$,
which also have operator norms at most 1. 
We also have the following claim which shows that the arm set is well-conditioned; its proof is deferred to Section~\ref{sec:proof-aux}.

\begin{claim}
\label{claim:c-min}
$C_{\min}(\cA) \geq C$.
\end{claim}

\paragraph{Bandit environments (hypotheses).} Define a ``null hypothesis'', a rank-$r$ matrix bandit environment 
\[
\Theta = \begin{bmatrix}
    \diag(\epsilon \one_{r-1}) & 0 & 0 \\
    0 & \diag(0_{d-r}) & 0 \\
    0 & 0 &  -\frac{\rmax}{2}
    \end{bmatrix},
\]
and define its ``support matrix''
\[
Z = 
\begin{bmatrix}
    \diag(\one_{r-1}) & 0 & 0 \\
    0 & \diag(0_{d-r}) & 0 \\
    0 & 0 &  0
    \end{bmatrix},
\]

Define an ``alternative hypothesis'' of bandit environment $\tilde{\Theta} = \Theta_0 + 2\epsilon \tilde{Z}$, where 
\[
\tilde{Z} = \argmin_{Z \in \cS'} \EE_\Theta \sbr{ \sum_{t=1}^n (\inner{A_t}{Z})^2 },
\]
here, 
\begin{equation}
\cS' = \cbr{ 
\begin{bmatrix}
    \diag(0_{r-1}) & 0 & 0 \\
    0 & U V^\top & 0 \\
    0 & 0 &  0
    \end{bmatrix}
    :\quad 
 U, V \in \RR^{(d-r) \times (r-1)}, U^\top U = V^\top V = I_{r-1}
} \subseteq \cS.
\label{eqn:def-s-prime}
\end{equation}
Note that by item~\ref{item:theta-bound}, $r\epsilon \leq \frac{\rmax}{24}$, therefore, $\| \Theta \|_* = \frac{\rmax}{2} + (r-1) \epsilon \leq \rmax$, 
and 
$\| \tilde{\Theta} \|_* = \frac{\rmax}{2} + (r-1) \epsilon + 2(r-1) \epsilon \leq \rmax$.

Our goal below is to show that one of $\EE_\Theta[\Reg(\Theta, n)]$ and $\EE_{\tilde{\Theta}}[\Reg(\tilde{\Theta}, n)]$  must be large.

\begin{remark} 
If we define $T = \cbr{\Theta + 2\epsilon Z: Z \in \cS'}$, then 
\[
\tilde{\Theta}
= 
\argmin_{{\Theta'} \in T} \KL(\PP_\Theta, \PP_{\Theta'}).
\]
Intuitively, $\tilde{\Theta}$ is the environment in $T$ that is ``most indistinguishable'' from $\Theta$.
\end{remark}

We make the following observation on the optimal arm and optimal reward of these two environments, whose proof can be found in Section~\ref{sec:proof-aux}:
\begin{claim}
\label{claim:optimal-rewards}
\begin{enumerate}
\item $\max_{A \in \cA} \inner{A}{\Theta} = (r-1)\epsilon$, $\argmax_{A \in \cA} \inner{A}{\Theta} = Z$;
\item $\max_{A \in \cA} \inner{A}{\tilde{\Theta}} = 2(r-1)\epsilon$, $\argmax_{A \in \cA} \inner{A}{\tilde{\Theta}} = \tilde{Z}$.
\end{enumerate}
\end{claim}

\subsection{Proof of Theorem~\ref{thm:lower-bound-lr}}

\paragraph{Step 1: Not enough exploration leads to high regret.} 
Denote by $T(\cH) := \sum_{t=1}^T I(A_t \in \cH)$ the number of times the learner chooses arms in the informative arm set $\cH$. 
We show the following claim, which formalizes the intuition that if $\EE_\Theta [T(\cH)]$, the number of times the informative arms are chosen, is small, at least one of the environments $\Theta, \tilde{\Theta}$ must induce a large regret.
\begin{claim}
\label{claim:ne-explore}
\[
\max( \EE_{\Theta}[\Reg(\Theta, T)], \EE_{\Theta'}[\Reg(\Theta', T)] ) 
\geq 
T r \epsilon \exp\del{- \frac{\EE_\Theta [T(\cH)] r C \epsilon^2}{\sigma^2} - \frac{T r^2 \epsilon^2}{d^2 \sigma^2} }
\]
\end{claim}
%Intuitively, this claim 

\begin{proof}[Proof of Claim~\ref{claim:ne-explore}]
Define event 
\[
D = \cbr{ \sum_{t=1}^T I(A_t \in \cS) \inner{A_t}{Z}  \leq \frac{T(r-1)}{2} }
\]

We show via the following lemma that to ensure low-regret in $\Theta$ and $\tilde{\Theta}$, it is necessary to control the probabilities of $D$ and $D^c$ to be small under the respective environment; its proof is deferred to Section~\ref{sec:proof-aux}:
\begin{lemma}
\label{lem:reg-d-lb}
$\EE_\Theta\sbr{ \Reg(\Theta, T)} \geq \frac{T(r-1)\epsilon}{2} \PP_\theta(D)$ 
and 
$\EE_{\tilde{\Theta}}\sbr{ \Reg(\tilde{\Theta}, n)} \geq \frac{T(r-1)\epsilon}{2} \PP_{\tilde{\theta}}(D^c)$
\end{lemma}

As an important consequence, by Bretagnolle-Huber inequality and divergence decomposition, 
\begin{align*}
& \max( \EE_{\Theta}[\Reg(\Theta, T)], \EE_{\Theta'}[\Reg(\Theta', T)] )  
\\
\geq &
\frac{T (r-1) \epsilon}{4} (P_\Theta(D) + P_{\tilde{\Theta}}(D^c)) \\
\geq &
\frac{T (r-1) \epsilon}{4} \exp\del{ - \KL(\PP_\Theta, \PP_{\tilde{\Theta}} ) } \\
\geq & 
\frac{T (r-1) \epsilon}{8} \exp\del{ - \EE_\Theta \sbr{ \sum_{t=1}^T \frac{1}{2\sigma^2}\inner{\Theta - \tilde{\Theta}}{A_t}^2 } } \\
\geq & 
\frac{T r \epsilon}{16} \exp\del{ - \frac{2\epsilon^2}{\sigma^2} \EE_\Theta \sbr{ \sum_{t=1}^T \inner{\tilde{Z}}{A_t}^2 } }
\end{align*}
The following key lemma (proof in Section~\ref{sec:proof-aux})
upper bounds the $\EE_\Theta \sbr{ \sum_{t=1}^T \inner{\tilde{Z}}{A_t}^2 }$ term in the exponent:
\begin{lemma}
\label{lem:tilde-z-divergence}
$\EE_\Theta \sbr{ \sum_{t=1}^T \inner{\tilde{Z}}{A_t}^2 } \leq 4 \EE_\Theta [T(\cH)] r C + \frac{8 T r^2}{d^2}$.
\end{lemma}
The claim follows by plugging this lemma into the above inequality.
\end{proof}

\paragraph{Step 2: Concluding the lower bound.} We now use Claim~\ref{claim:ne-explore} to conclude the minimax regret lower bound. Observe that by our setting of parameters, $d^2 \geq \frac{16 T r^2 \epsilon^2}{\sigma^2}$, therefore, Claim~\ref{claim:ne-explore} simplifies to 
\[
\max( \EE_{\Theta}[\Reg(\Theta, T)], \EE_{\tilde{\Theta}}[\Reg(\tilde{\Theta}, T)] ) 
\geq 
\frac{T r \epsilon}{32} \exp\del{- \frac{\EE_\Theta [T(\cH)] r C \epsilon^2}{\sigma^2}}
\]

Before proceeding, we make another important observation that under $\Theta$, arms in $\cH$ indeed incur large regret:
%\begin{observation}
\begin{claim}
\label{claim:reg-t-h}
\[
\EE_{\Theta}[\Reg(\Theta, n)]
\geq 
\frac{\rmax}{8} \EE_{\Theta}[T(\cH)]
\]
\end{claim}

%We now consider two cases:
%\item We will do the tuning so that 
%the exponential term in the first bound is a constant. To this end, we first set $d$ so that \red{} 

We now consider two cases: 
\begin{itemize}
\item If $\EE_\Theta [T(\cH)] \leq \frac{\sigma^2}{2 r C \epsilon^2}$, the exponent in the first inequality becomes a constant, so that we have 
$\max( \EE_{\Theta}[\Reg(\Theta, T)], \EE_{\tilde{\Theta}}[\Reg(\tilde{\Theta}, T)] ) \geq \frac{T r \epsilon}{64}$.

\item Otherwise, $\EE_\Theta [T(\cH)] > \frac{\sigma^2}{2 r C \epsilon^2}$. In this case, 
$
\EE_{\Theta}[\Reg(\Theta, T)] \geq \rmax \frac{\sigma^2}{16 r C \epsilon^2}
$
\end{itemize}
In summary, for any bandit algorithm,
\[
\max_{\Theta' \in \cbr{\Theta, \tilde{\Theta}}} \EE_{\Theta'}[\Reg(\Theta', n)]
\geq 
\frac1{64} \min\del{ T r \epsilon, \frac{\rmax \sigma^2}{ r C \epsilon^2 } }.
\]
%, there exists $\Theta' \in \cbr{\Theta, \tilde{\Theta}}$ such that 
Note that our choice of $\epsilon =  
\del{ \frac{\rmax \sigma^2}{T r^2 C} }^{\frac{1}{3}}$ balances these two terms and approximately maximizes the above; plugging its value, we get,
\[
\max_{\Theta' \in \cbr{\Theta, \tilde{\Theta}}} \EE_{\Theta'}[\Reg(\Theta', T)]
\geq
\frac1{64}
\rmax^{1/3} \sigma^{2/3} r^{1/3} T^{2/3} C^{-1/3}.
\]
This concludes the proof of Theorem~\ref{thm:lower-bound-lr}.
\qed

\subsection{Proofs of auxiliary lemmas}
\label{sec:proof-aux}

\subsubsection{Proofs related to the properties of the lower bound construction}

\begin{proof}[Proof of Claim~\ref{claim:c-min}]
It suffices to show that the uniform  distribution $\pi$ over $\cH \subseteq \cA$ satisfies $\EE_{A \sim \pi}\sbr{ \vec(A) \vec(A)^\top } \succeq C I$. 
%Indeed, choose $\pi$ to be the uniform distribution over $\cH$.

Define $\tilde{\pi}$ to be the uniform distribution over 
\[ 
    \tilde{\cH} =
    \cbr{
    X = 
    \begin{bmatrix}
    x_{11}, \ldots, x_{1d} \\
    \ldots \\
    x_{d1}, \ldots, \frac12
    \end{bmatrix}
    :
    \forall (i,j) \neq (d,d), 
    x_{i,j} \in \cbr{-\sqrt{2C}, \sqrt{2C}}
    };
\]
it can be seen that $\pi$ is distribution $\tilde{\pi}$ restricted to the set $\cbr{A: \| A \|_{\op} \leq 1}$. 
It therefore suffices to show that 
$\EE_{A \sim \tilde{\pi}}\sbr{ \vec(A) \vec(A)^\top I( \| A \|_{\op} \leq 1) } \succeq C I$, as 
\[
\EE_{A \sim \pi}\sbr{ \vec(A) \vec(A)^\top } = \frac{1}{ \PP_{A \sim \tilde{\pi}}( \| A \|_{\op} \leq 1 ) } \EE_{A \sim \tilde{\pi}}\sbr{ \vec(A) \vec(A)^\top I( \| A \|_{\op} \leq 1) }.
\]

Note that $\EE_{A \sim \tilde{\pi}}\sbr{ \vec(A) \vec(A)^\top} \succeq 2 C I$; hence, it reduces to show that 
\[
\norm{ \EE_{A \sim \tilde{\pi}}\sbr{ \vec(A) \vec(A)^\top I(\| A \|_{\op} > 1) } }_{\op} \leq C.
\]
Note that for all $A \in \tilde{\cH}$, $\| \vec(A) \|_2 = \| A \|_F \leq d$, which implies that $\| \vec(A) \vec(A)^\top \|_{\op} \leq d^2$; therefore, it suffices to show that 
\[
\PP_{A \sim \tilde{\pi}} ( \| A \|_{\op} \geq 1 )
\leq 
\frac{C}{d^2}.
\]

%A \sim 
Indeed, denote by a random sample from $\tilde{\pi}$ by 
$A
= 
\begin{bmatrix}
    x_{11}, \ldots, x_{1d} \\
    \ldots \\
    x_{d1}, \ldots, \frac12
\end{bmatrix}
$
where all $x_{i,j}$'s are drawn uniformly from $\cbr{-\sqrt{2C}, \sqrt{2C}}$. 

From Lemma~\ref{lem:mat-pm1}, with probability $1-\frac{C}{d^2}$, random matrix 
$
X = 
\begin{bmatrix}
    x_{11}, \ldots, x_{1d} \\
    \ldots \\
    x_{d1}, \ldots, x_{dd}
\end{bmatrix}
$  
(where $x_{dd}$ is also drawn uniformly from $\cbr{-\sqrt{2C}, \sqrt{2C}}$) satisfies that
\[
\| X \|_{\op} \leq c \sqrt{2C} \del{ \sqrt{d} + \sqrt{ \ln\frac{d^2}{2C} } }
\]

Observe that by the assumption that $C \leq \frac{1}{10 d}$, $c \sqrt{2C} \del{ \sqrt{d} + \sqrt{ \ln\frac{d^2}{2C} } } + \sqrt{2C} \leq \frac12$, therefore, 
\[
\norm{A}
\leq 
\norm{X}_{\op} 
+ 
\norm{\begin{bmatrix}
    0, \ldots, 0 \\
    \ldots \\
    0, \ldots, x_{dd}
\end{bmatrix}  }_{\op}
+
\norm{\begin{bmatrix}
    0, \ldots, 0 \\
    \ldots \\
    0, \ldots, \frac12
\end{bmatrix}  }_{\op}
\leq \frac12 + \frac12 \leq 1.
\]

%= 
% \norm{ 
% \begin{bmatrix}
%     x_{11}, \ldots, x_{1d} \\
%     \ldots \\
%     x_{d1}, \ldots, \frac12
% \end{bmatrix}
% }_{\op}

%\chicheng{Some details need to be added here.}
\end{proof}

\begin{proof}[Proof of Claim~\ref{claim:optimal-rewards}]
We prove the two items respectively.
\begin{enumerate}
\item Observe that 
\[
\max_{A \in \cH} \inner{A}{\Theta} \leq (r-1)\epsilon - \frac14 \rmax \leq -\frac18 \rmax,
\]
\[
\max_{A \in \cS} \inner{A}{\Theta} = \max_{A \in \cS} \inner{A^1}{\Theta^1}
\leq \max_{A \in \cS} \| A^1 \|_{*} \|{\Theta^1} \|_{\op} \leq (r-1) \epsilon 
\]
Furthermore, note that $Z \in \cS$ and 
\[
\inner{Z}{\Theta} = (r-1)\epsilon. 
\]
This shows that $Z$ maximizes $\inner{A}{\Theta}$ over all $A \in \cA$ and achieves objective value $(r-1)\epsilon$.

\item Observe that 
\[
\max_{A \in \cH} \inner{A}{\tilde{\Theta}} \leq 3(r-1)\epsilon - \frac14 \rmax \leq -\frac18 \rmax,
\]
\[
\max_{A \in \cS} \inner{A}{\tilde{\Theta}} = \max_{A \in \cS} \inner{A^1}{\tilde{\Theta}^1}
\leq \max_{A \in \cS} \| A^1 \|_{*} \|{\tilde{\Theta}^1} \|_{\op} \leq 2(r-1) \epsilon 
\]
Furthermore, note that $\tilde{Z} \in \cS$ and 
\[
\inner{\tilde{Z}}{\tilde{\Theta}} = 2(r-1)\epsilon. 
\]
This shows that $\tilde{Z}$ maximizes $\inner{A}{\tilde{\Theta}}$ over all $A \in \cA$ and achieves objective value $2(r-1)\epsilon$.
\end{enumerate}
\end{proof}

In the proof of Claim~\ref{claim:c-min}, we need the following lemma on $\pm 1$ random matrices:
\begin{lemma}
\label{lem:mat-pm1}
Suppose $A$ is a random matrix whose entries are drawn iid and uniformly at random from $\cbr{-1, +1}$. Then there exists some constant $c$, such that with probability $1-\delta$,
\[
\| A \|_{\op} \leq c \del{ \sqrt{d} + \sqrt{ \ln\frac{1}{\delta} } }.
\]
\end{lemma}
\begin{proof}
This follows from~\citet[][Theorem 5.39]{vershynin2010introduction} applied to matrix $A$, and the observation that every row of $A$ is a 1-subgaussian random vector.
\end{proof}

\subsubsection{Proofs related to the lower bound proof}

\begin{proof}[Proof of Lemma~\ref{lem:reg-d-lb}]
For the first inequality, it suffices to show that when event $D$ happens, $\Reg(\Theta, T) \geq \frac{T(r-1)\epsilon}{2}$. 
%\EE_{\Theta}[
%\EE_{\Theta}\sbr{} 
Indeed,  
\begin{align*}
\Reg(\Theta, T)
&= 
T (r-1)\epsilon - \sum_{t=1}^T \inner{A_t}{\Theta} I(A_t \in \cS)  - 
\sum_{t=1}^T \inner{A_t}{\Theta} I(A_t \in \cH) \\
&\geq  
T(r-1)\epsilon - \sum_{t=1}^T \inner{A_t}{\Theta} I(A_t \in \cS) \\
&=  
T(r-1)\epsilon - \epsilon \sum_{t=1}^T \inner{A_t}{Z} I(A_t \in \cS) \\
& \geq 
\frac{T(r-1)\epsilon}{2}
\end{align*}
where the first inequality uses the observation that when $A_t \in \cH$,
$\inner{A_t}{\Theta} = \inner{A_t}{\Theta}_1 - \frac14 \rmax \leq \| A_t^1 \|_{\op} \| \Theta^1 \|_* \leq (r-1)\epsilon \| A_t \|_{\op}  - \frac14 \rmax \leq (r-1)\epsilon - \frac14 \rmax \leq 0$; the second equality is due to that for $A_t \in \cS$, $\inner{A_t}{\Theta} = \inner{A_t}{\epsilon Z}$; the second inequality is due to the definition of event $D$.

For the second inequality, we first claim that 
\begin{equation}
D^c \subset \cbr{ \sum_{t=1}^T I(A_t \in \cS) \inner{A_t}{\tilde{Z}}  \leq \frac{n(r-1)}{2} }
\label{eqn:dc-tildez}
\end{equation}

To see this, note that 
\[
\sum_{t-1}^T I(A_t \in \cS) \inner{A_t}{Z + \tilde{Z}}
\leq 
\sum_{t-1}^T I(A_t \in \cS) \|A_t\|_* \| Z + \tilde{Z} \|_{\op}
\leq 
T(r-1),
\]
where the second inequality uses the observations that for $A_t \in \cS$, $\| A_t \|_* = r-1$, and that 
$\| Z + \tilde{Z} \|_{\op} \leq 1$. 
Also, recall from its definition that when $D^c$ happens, 
\[
\sum_{t=1}^T I(A_t \in \cS) \inner{A_t}{\tilde{Z}} \geq \frac{T(r-1)}{2},
\]
Subtracting the above two inequalities, we have that $
\sum_{t=1}^T I(A_t \in \cS) \inner{A_t}{\tilde{Z}}  \leq \frac{T(r-1)}{2}
$ holds. 

We now claim that when event $D^c$ happens, $\Reg(\tilde{\Theta}, T) \geq \frac{T(r-1)\epsilon}{2}$. 
\begin{align*}
\Reg(\tilde{\Theta}, T)
&= 
2T(r-1)\epsilon - \sum_{t=1}^T \inner{A_t}{\tilde{\Theta}} I(A_t \in \cS)  - 
\sum_{t=1}^T \inner{A_t}{\tilde{\Theta}} I(A_t \in \cH) \\
&\geq  
2T(r-1)\epsilon - \sum_{t=1}^T \inner{A_t}{\tilde{\Theta}} I(A_t \in \cS) \\
&=  
2T(r-1)\epsilon - \epsilon \sum_{t=1}^T \del{ \inner{A_t}{Z} + 2 \inner{A_t}{\tilde{Z}} } I(A_t \in \cS) \\
&\geq 
T(r-1)\epsilon - \epsilon \sum_{t=1}^T \inner{A_t}{\tilde{Z}} I(A_t \in \cS) \\
& \geq 
\frac{T(r-1)\epsilon}{2}
\end{align*}
where the first inequality uses the observation that when $A_t \in \cH$,
$\inner{A_t}{\tilde{\Theta}} = \inner{A_t}{\tilde{\Theta}}_1 - \frac14 \rmax \leq \| A_t^1 \|_{\op} \| \tilde{\Theta}^1 \|_* \leq 3 (r-1)\epsilon \| A_t \|_{\op}  - \frac14 \rmax \leq 3 (r-1)\epsilon - \frac14 \rmax \leq 0$; the second equality is due to that $\inner{A_t}{\tilde{\Theta}} = \inner{A_t}{\epsilon Z + 2\epsilon\tilde{Z}}$; the second inequality uses the fact that $\sum_{t=1}^T \del{ \inner{A_t}{Z} + \inner{A_t}{\tilde{Z}} } I(A_t \in \cS) \leq \sum_{t=1}^T \| A_t \|_* \| Z + \tilde{Z} \| \leq n(r-1)$; the 
third inequality uses our claim Eq.~\eqref{eqn:dc-tildez} above.
\end{proof}

\begin{proof}[Proof of Lemma~\ref{lem:tilde-z-divergence}] 
Note that 
$
\EE_\theta \sbr{ \sum_{t=1}^T \inner{\tilde{Z}}{A_t}^2 }
\leq 
\EE_\theta  \EE_{Z \sim D}
\sbr{
\sum_{t=1}^T \inner{Z}{A_t}^2
}
$ for any distribution $D$ over $\cS'$. We choose $D$ in the following manner.
Randomly draw $Z$ using this procedure and call the $D$ the resultant distribution of $Z$: 
\begin{enumerate}
\item Draw $U = (u_1, \ldots, u_{r-1}), V = (v_1, \ldots, v_{r-1}) \in \RR^{(d-r) \times (r-1)}$ from Haar measure over matrices with orthonormal columns.
\item Draw $\sigma \sim \Uniform(\cbr{-1,+1}^{d-r})$, a Rademacher random vector;
\item Define 
\[
Z = 
\begin{bmatrix}
\diag(0_{r-1}) & 0 & 0 \\
0 & U \diag(\sigma) V^\top & 0 \\
0 & 0  & 0
\end{bmatrix}.
\]
\end{enumerate}

% For the rest of the proof, 
% it suffices to show that for any arm $A \in \cA$, 
% \[
%  \EE_{Z \sim D}
% \sbr{
% \sum_{t=1}^n \inner{x}{A}^2
% }
% \leq 
% I() r C 
% \]

%\chicheng{TBD: Need to ensure $U, V$ have ranges within $\mathrm{span}(e_{r}, \ldots, e_{d-1})$}
%\ks{What about $U',V'\in \RR^{(d-r-2) \times (r-1)}$ and $U=[O_{r-1\times r-1} U']$}
%\chicheng{Right - I just changed the above.}

% \item draw $T \sim \Uniform( {\cbr{s,s+1,\ldots,d-1} \choose s-1} )$ (this would constitute the support of $x$)
% \item draw $\sigma \sim \Uniform(\cbr{-1,+1}^d)$ which is a Rademacher random vector
% \item let $x = \one_T \odot \sigma$, where $\one_T$ is the vector with $1$ on $T$;  

%\ks{What is $A_1$?}
%\chicheng{This notation is defined in the third item of this section. Actually I switched to using $A_2$ below.}
Now, fix a $A \in \RR^{d \times d}$; we seek to bound $\EE_{Z \sim D}[\inner{Z}{A}^2]$. 
Observe that $\inner{Z}{A} = \inner{Z}{A}_2$ as $Z$ is only nonzero in rows and columns $r$ through $d-1$. 
Therefore, (recalling the notation $A^2$ in Section~\ref{sec:construction})
\begin{align*}
\EE_{Z \sim D}
\sbr{
\sum_{t=1}^n \inner{Z}{A}^2
}
= & 
\EE\sbr{ 
\EE
\sbr{
\del{\sum_{i=1}^{r-1} \sigma_i (u_i^\top A^2 v_i) }^2
\mid u_1, \ldots, u_{r-1}, v_1, \ldots, v_{r-1}}
} \\
= & 
\EE
\sbr{
\sum_{i=1}^{r-1} (u_i^\top A^2 v_i)^2
}
\\
= & 
(r-1)
\EE_{u, v \sim \Uniform(S^{d-r-1})}
\sbr{
(u^\top A^2 v)^2
}
\\
= &
\frac{r-1}{(d-r)^2} \| A^2 \|_F^2
\leq 
\frac{4r}{d^2} \| A^2 \|_F^2
\end{align*}
where the second equality use the observation that $\EE\sbr{\sigma_i \sigma_j} = I(i=j)$; the third equality uses the linearity of expectation and that $u_i, v_i$'s marginal distributions uniform over $d-r$-dimensional unit sphere. 
The last equality uses the observation that
\[
\EE_{u, v \sim \Uniform(S^{d-r-1})}
\sbr{
(u^\top A^2 v)^2
}
= 
\frac{1}{d-r} \EE_{v \sim \Uniform(S^{d-r-1})} \| A^2 v \|_2^2
= 
\frac{1}{(d-r)^2} \| A^2 \|_F^2
\]

%\approx 
% \frac{1}{d^2} \EE_{u, v \sim N(0, I_{d-1})}
% \sbr{
% (u^\top A_2 v)^2
% }
%\chicheng{The last step is pending verification..}

The above calculation implies that:
\begin{itemize}
\item For $A \in \cH$, as $\| A^2 \|_F = \sqrt{C} (d-r)$, we have $\EE_{Z \sim D}
\sbr{
\inner{Z}{A}^2
} \leq 8 r C$,

%\chicheng{I think $d \kappa$ is the best possible upper bound for $\| A_2 \|_F$ for $A_2 \in \cH$ -- note that in our construction, we need to ensure that there exists some $\pi$ such that $\lambda_{\min}( \sum_{a \in \cH} \pi_a \ve(a) \ve(a)^\top ) \geq \kappa^2$
% This implies 
% \[
% \tr(\sum_{a \in \cH} \pi_a \ve(a) \ve(a)^\top )) \geq d^2 \kappa^2
% \implies
% \max_{a \in \cH} \| a \|_{F}^2
% =
% \max_{a \in \cH} \| \vec(a) \|_2^2
% \geq d^2 \kappa^2.
% \]
% }

\item For $A \in \cS$, as $\| A^2 \|_F \leq \sqrt{r} \| A^2 \|_{\op} \leq \sqrt{r}$, we have $\EE_{Z \sim D}
\sbr{
\inner{Z}{A}^2
} \leq \frac{4r^2}{d^2}$.

%\chicheng{(5/12/23) Made a pass until here. The remaining needs to be revised.}
\end{itemize}

Therefore, 
\[
\EE_\Theta \sbr{ \sum_{t=1}^n \inner{\tilde{x}}{A_t}^2 }
\leq 
\EE_\Theta \sbr{ \sum_{t=1}^T I(A_t \in \cH) r \kappa^2 + \sum_{t=1}^T I(A_t \in \cS) \frac{r^2}{d}} 
\leq 
8 \EE_\Theta [T(\cH)] r C + \frac{4 T r^2}{d^2}.
\]
\end{proof}

\begin{proof}[Proof of Claim~\ref{claim:reg-t-h}]
First, by the definition of regret, 
\begin{align*}
\Reg(\Theta, T)
= &
\sum_{t=1}^T \del{ \max_{A \in \cA} \inner{A}{\Theta} -  \inner{A_t}{\Theta} } I(A_t \in \cS) + \del{ \max_{A \in \cA} \inner{A}{\Theta} - \inner{A_t}{\Theta} } I(A_t \in \cH) \\
\geq & 
\sum_{t=1}^T \del{ \max_{A \in \cA} \inner{A}{\Theta} - \inner{A_t}{\Theta} } I(A_t \in \cH)
\end{align*}

Observe that: first, $\max_{A \in \cA} \inner{A}{\Theta} = (r-1)\epsilon$; 
second, 
when $A_t \in \cH$,
$\inner{A_t}{\tilde{\Theta}} = \inner{A_t}{\tilde{\Theta}}_1 - \frac14 \rmax \leq \| A_t^1 \|_{\op} \| \tilde{\Theta}^1 \|_* \leq 3 (r-1)\epsilon \| A_t \|_{\op}  - \frac14 \rmax \leq 3 (r-1)\epsilon - \frac14 \rmax \leq -\frac18 \rmax$. Therefore, $\Reg(\Theta, n) \geq \frac18 \rmax \cdot T(\cH)$. The claim follows by taking expectation over both sides with respect to $\PP_{\Theta}$.
\end{proof}

% \putbib[sections/bibliography]
% \end{bibunit}

% \bibliography{sections/bibliography} % just so that my latex tool 
%%%%%%%%%%%%%%%%%%%%%%%%%%%%%%%%%%%%%%%%%%%%%%%%%%%%%%%%%%%%%%%%%%%%%%%%%%%%%%%
%%%%%%%%%%%%%%%%%%%%%%%%%%%%%%%%%%%%%%%%%%%%%%%%%%%%%%%%%%%%%%%%%%%%%%%%%%%%%%%
\end{document}